\pdfoutput=1
\documentclass{article}

\usepackage[main, final]{neurips_2026}

\usepackage[utf8]{inputenc}
\usepackage[T1]{fontenc}

\usepackage{xcolor}
\definecolor{citecol}{HTML}{1F6FEB}
\usepackage[colorlinks=true,citecolor=citecol,linkcolor=black,urlcolor=citecol,hyperfootnotes=false]{hyperref}
\setcitestyle{numbers,square,comma}
\usepackage{url}
\usepackage{booktabs}
\usepackage{amsfonts}
\usepackage{amsmath}
\usepackage{amssymb}
\usepackage{amsthm}
\newtheorem{lemma}{Lemma}
\newtheorem{theorem}{Theorem}
\newtheorem{proposition}{Proposition}

\theoremstyle{definition}

\theoremstyle{remark}

\theoremstyle{plain}
\usepackage{nicefrac}
\usepackage{microtype}
\usepackage{graphicx}
\usepackage{subcaption}

\usepackage{algorithm}
\usepackage{algorithmic}
\usepackage{adjustbox}
\usepackage{lscape}
\usepackage{multirow}
\usepackage{colortbl}
\definecolor{countbg}{RGB}{253,230,230}
\usepackage{placeins}
\usepackage{float}
\usepackage{etoc}
\usepackage{enumitem}

\newcommand{\best}[1]{\textcolor{red}{\textbf{#1}}}
\newcommand{\second}[1]{\textcolor{blue}{\underline{#1}}}

\title{Koopman Generative Operators for Efficient Probabilistic Time-Series Forecasting}

\author{%
  \textbf{Raz Marshanski}$^{1}$ \qquad \textbf{Liran Nochumsohn}$^{1}$ \qquad \textbf{Mayank Jauhari}$^{2*}$ \\[0.8em]
  \textbf{Boris N. Oreshkin}$^{2*}$ \qquad \textbf{Omri Azencot}$^{1}$ \\[1em]
  $^{1}$Ben-Gurion University of the Negev \qquad
  $^{2}$Amazon Science
}

\begin{document}

\maketitle
{\renewcommand{\thefootnote}{}\footnotetext{Correspondence to: \texttt{razmar@post.bgu.ac.il}. \quad $^{*}$This work is not related to the authors' positions at Amazon.}}

\addtocontents{toc}{\protect\setcounter{tocdepth}{-2}}

\begin{abstract}

Probabilistic time-series forecasting requires models that simultaneously capture structured temporal dynamics, expressive uncertainty, and efficient inference, yet existing approaches fall short of this goal: latent dynamical models impose structure but rely on restrictive generation mechanisms, while modern generative methods such as diffusion and flow matching achieve flexibility at the cost of iterative and computationally expensive sampling. We introduce the Koopman Generative Operator (KGO), a probabilistic forecasting method that conceptualizes prediction as the evolution of structured uncertainty. KGO integrates three core components: (1) Koopman Patch Embedding (KoPE) for temporally consistent latent trajectory extrapolation; (2) Koopman Flow Matching (KoFM), which enables fast, single-step generation through a closed-form matrix exponential in a Koopman latent space, bypassing iterative sampling; and (3) an Adaptive Uncertainty Gate (AUG), which provides probabilistic forecasts by adapting uncertainty per-variable and per-horizon, corresponding to a learned decomposition of aleatoric uncertainty without manual tuning. By unifying these components, KGO achieves state-of-the-art accuracy on 13/17 ProbTS benchmarks in CRPS and 16/17 in NMAE. Further, by eliminating iterative sampling, KGO delivers at least a 25$\times$ reduction in inference time compared to iterative generative models. These results position KGO as a practical and principled framework that advances the accuracy--efficiency trade-off in probabilistic time-series forecasting, combining leading accuracy on the evaluated benchmarks with fast, single-step generation.
\par\vspace{0.5em}
{\centering \raisebox{-0.15em}{\includegraphics[height=1em]{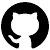}}\ \url{https://github.com/azencot-group/KGO}\par}

\end{abstract}

\vspace{-6mm}
\section{Introduction}
\label{sec:intro}

\vspace{-2mm}
\emph{Probabilistic time-series forecasting} is essential for decision-making under uncertainty, requiring models to predict not only expected outcomes but full predictive distributions over trajectories, rather than single-point estimates~\cite{rangapuram2018deepstate, salinas2020deepar}. In applications such as energy systems, finance, and healthcare, miscalibrated uncertainty, i.e., inaccurate estimates of the range and likelihood of possible outcomes, can have severe consequences: underestimation leads to critical failures, while overestimation results in overly conservative and inefficient decisions~\cite{gneiting2014probabilistic, hong2016probabilistic}. Recent advances in deep learning have significantly improved the modeling of complex temporal dependencies and multimodal predictive distributions, with approaches based on flows, diffusion, and VAEs demonstrating strong empirical performance~\cite{rasul2021tmaf, rasul2021timegrad, wang2024k2vae, naiman2024generative, naiman2024utilizing, gonen2026time, liu2025sundial}. However, practical deployment remains challenging, as modern probabilistic forecasting systems must achieve high predictive accuracy, reliable uncertainty estimates, and efficient inference in real-world settings~\cite{salinas2020deepar, rasul2021timegrad}. Bridging this gap between modeling capability and deployment requirements remains a central challenge in probabilistic forecasting.

Despite these advances, existing probabilistic forecasting approaches address different aspects of the problem in a fragmented manner, and no single framework simultaneously captures expressive uncertainty, structured temporal dynamics, and efficient inference. First, modern generative approaches based on diffusion and flow matching provide flexible modeling of complex, multimodal predictive distributions, but rely on \emph{iterative sampling} procedures whose computational cost scales with the number of steps, limiting their applicability in large-scale or real-time settings~\cite{rasul2021timegrad, tashiro2021csdi, lipman2023flow, berman2026one}. Second, many methods compress the historical context into a single representation, either as an explicit latent state, as in latent-variable and Koopman-based models, or through local conditioning mechanisms such as last-patch representations. While latent-variable approaches enable structured temporal evolution, they typically tie uncertainty to the initial state, inducing variability through its propagation rather than modeling how uncertainty evolves over time. This results in a tight coupling between temporal dynamics and uncertainty modeling, limiting flexibility in how predictive distributions are formed, while local conditioning strategies can obscure long-range dependencies and \emph{limit extrapolation} across extended horizons~\cite{rangapuram2018deepstate, wang2024k2vae, liu2025sundial}. Finally, existing approaches often adopt fixed or heuristic strategies for uncertainty modeling, treating uncertainty in a \emph{global or implicit manner}, despite the fact that predictability, and thus the appropriate level of uncertainty, \emph{varies across variables and forecast horizons}~\cite{li2025d3u, li2024tmdm}. Together, these limitations highlight the need for a unified approach that jointly supports efficient generation, models temporal dynamics, and adapts uncertainty to the underlying predictability of the data.

To address these limitations, we argue that probabilistic forecasting should be viewed as modeling the \emph{evolution of structured uncertainty over time}, rather than predicting distributions in a static manner or generating samples through iterative simulation. Future outcomes arise from underlying temporal dynamics, implying that effective forecasting requires representations that preserve and extrapolate this structure across the prediction horizon. At the same time, uncertainty is inherently heterogeneous: the predictability of future observations varies across variables and time, suggesting that uncertainty should be adapted to the underlying signal rather than treated uniformly. This perspective reframes forecasting as a transformation of uncertainty governed by temporal structure, rather than a static prediction or an iterative generative process. It suggests decoupling temporal evolution from uncertainty modeling, allowing each to be handled in a principled and efficient manner.

To instantiate this perspective, we propose \emph{KGO}, a unified framework for probabilistic forecasting. In contrast to prior Koopman-based approaches that evolve latent states and rely on sampling-based decoding, KGO directly models the transport from noise to future trajectories. At its core is \emph{Koopman Flow Matching (KoFM)}, a formulation that replaces iterative generative processes with a closed-form transformation in a learned latent space~\cite{engel2000one, azencot2013operator, azencot2020forecasting}. Specifically, rather than simulating trajectories through multiple steps as in diffusion or flow-based models, KoFM learns a structured transport that maps noise to future samples via linear evolution, enabling expressive yet tractable dynamics and efficient single-step generation. To support this mechanism, we introduce \emph{Koopman Patch Embedding (KoPE)}, which constructs latent representations that preserve temporal structure and admit extrapolation across the forecast horizon. By lifting observed time series into a space where evolution can be modeled in a structured manner, KoPE provides the state representation on which KoFM operates. Finally, to account for heterogeneous uncertainty, we propose an \emph{Adaptive Uncertainty Gate (AUG)} that modulates the contribution of deterministic and stochastic components based on the predictability of each variable and horizon, allowing the model to allocate uncertainty dynamically. Together, these components realize forecasting as a transformation of structured uncertainty, yielding accurate probabilistic forecasts with efficient inference.

We evaluate KGO on the ProbTS benchmark~\cite{zhang2024probts} and observe consistent improvements across a wide range of probabilistic forecasting tasks. KGO achieves the lowest CRPS on \textbf{13/17} evaluated settings and the lowest NMAE on \textbf{16/17}, with strong results in both regimes: it ranks first on \textbf{7/9} long-term datasets in CRPS and \textbf{9/9} in NMAE, and on \textbf{6/8} short-term datasets in CRPS and \textbf{7/8} in NMAE. Notably, KGO enables single-step inference, yielding over \textbf{25$\times$} faster inference than iterative methods. Ablation studies show that replacing KoPE with a last-patch representation, the sLSTM encoder with a Transformer, or KoFM with a normalizing-flow head increases average CRPS by 16.9\%, 4.7\%, and 9.4\%, respectively. These results demonstrate that modeling forecasting as the evolution of structured uncertainty yields both strong predictive accuracy and substantial gains in inference efficiency.

\noindent\textbf{Contributions.}
We introduce the \emph{Koopman Generative Operator (KGO)}, a framework for probabilistic time-series forecasting that models prediction as the \emph{evolution of structured uncertainty} rather than iterative generative sampling.
\vspace{-3mm}
\begin{enumerate}
\item \textbf{A unifying perspective.}
We formulate probabilistic forecasting as the evolution of structured uncertainty, providing an alternative to iterative diffusion and flow-based generation and motivating a transport-based view of prediction.
\vspace{-1mm}
\item \textbf{A unified methodology.}
We develop KGO, which integrates (i)~\emph{Koopman Patch Embedding (KoPE)} for structured latent evolution, (ii)~\emph{Koopman Flow Matching (KoFM)} for closed-form, single-step probabilistic generation, and (iii)~an \emph{Adaptive Uncertainty Gate (AUG)} for per-variable, per-horizon uncertainty modeling. Together, these components bridge the gap between temporal structure, expressive uncertainty, and efficient inference.
\vspace{-1mm}
\item \textbf{Strong empirical performance.}
On the ProbTS benchmark, KGO achieves state-of-the-art results on \textbf{29/34} configurations, improving CRPS by up to \textbf{27\%}, and reducing iterative inference cost by at least \textbf{25$\times$}.
\end{enumerate}

\section{Related Work}
\label{sec:related}

\vspace{-2mm}
\subsection{Deterministic Time-Series Forecasting}
\vspace{-2mm}

Classical time-series forecasting has long relied on statistical models such as ARIMA~\cite{box2015time} and vector autoregression~\cite{sims1980macro}, which provide strong theoretical foundations but struggle with complex non-linearities and high-dimensional dependencies. The advent of deep learning introduced more expressive models, including recurrent architectures such as LSTMs and GRUs~\cite{elman1990rnn,hochreiter1997lstm}, which improved flexibility but remain limited in capturing long-range temporal dependencies. Recent approaches have explored a wide range of architectures, including convolutional, Transformer-based, and MLP-based models~\cite{wu2021autoformer,nie2023patchtst,zeng2023dlinear,wang2024timemixer, nochumsohn2025multi, nochumsohn2025super, liu2025freqmoe}. Among these, Transformer and MLP-based methods have emerged as dominant paradigms due to their ability to model long-range dependencies and scale to high-dimensional data. More recently, there has been renewed interest in recurrent architectures through the development of xLSTM and its variants~\cite{beck2024xlstm,kraus2025xlstmmixer,kong2025plstm,alharthi2024xlstmtime}, which improve memory capacity and training stability, making them competitive with modern sequence models. Despite these advances, these methods primarily produce point forecasts and do not model predictive uncertainty, limiting their applicability in decision-making settings where uncertainty is critical.

\vspace{-2mm}
\subsection{Probabilistic Time-Series Forecasting}
\vspace{-2mm}

Probabilistic time-series forecasting has been addressed through a range of approaches that differ in how predictive distributions are modeled and generated. Early methods such as DeepAR~\cite{salinas2020deepar} and DeepState~\cite{rangapuram2018deepstate} parameterize simple distributions conditioned on historical context, offering efficient inference but limited expressivity. To capture more complex, multimodal distributions, subsequent works have introduced generative models based on normalizing flows~\cite{rasul2021tmaf}, variational autoencoders~\cite{wang2024k2vae}, and diffusion and flow matching-based approaches~\cite{rasul2021timegrad,tashiro2021csdi,kollovieh2024tsdiff,naiman2024utilizing,liu2025sundial}. From the perspective of generation, diffusion and flow-based methods excel in predicting distributions via flexible generative transport processes, but rely on \emph{iterative sampling} procedures whose computational cost scales with the number of steps, limiting their use in real-time settings~\cite{liu2025sundial}. This has led to the perception of a trade-off between expressivity and efficiency, which we address by enabling flexible probabilistic modeling with single-step inference.

In terms of \emph{temporal modeling}, Koopman-based approaches impose structured, approximately linear dynamics in a learned latent space for sequence prediction and forecasting~\cite{azencot2020forecasting,liu2024koopa,wang2023koopman,naiman2023operator}. This principle has also been extended to probabilistic and generative sequence modeling~\cite{naiman2024generative,wang2024k2vae,zhang2025skolr}, where latent uncertainty is propagated through structured dynamics. In contrast, many forecasting architectures capture temporal dependencies implicitly through learned conditioning mechanisms, without explicitly
constraining the evolution of the latent state~\cite{nie2023patchtst}. $K^2$VAE~\cite{wang2024k2vae} models uncertainty through a latent initial state sampled from a learned posterior, whose deterministic evolution induces a distribution over future trajectories. In contrast, KGO decouples temporal evolution from generative modeling: Koopman dynamics structure latent trajectories, while a generative operator directly maps source noise to future observations, allowing uncertainty to be modeled at the level of the forecasted sequence. Beyond modeling predictive distributions, KGO explicitly learns how to balance deterministic and stochastic predictions across variables and forecast horizons.

\section{Method}
\label{sec:method}

\begin{figure}[!h]
    \centering
    \includegraphics[width=\linewidth]{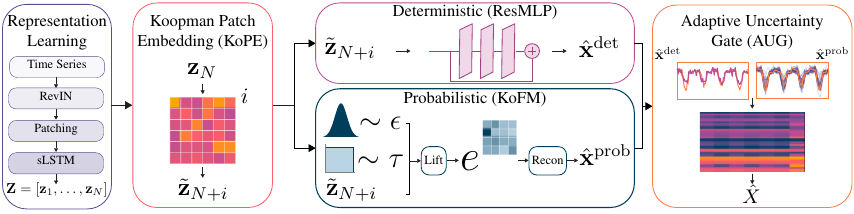}
    \caption{KGO consists of four main parts. (Left) A representation learning stage transforms the input time series via RevIN normalization, patching, and an sLSTM encoder into a latent state.
    (Middle-left) Koopman Patch Embedding (KoPE) extrapolates latent states by applying powers of a learned linear operator in latent space.
    (Middle-right, top) A residual MLP (ResMLP) decodes the KoPE latents into a deterministic point forecast. (Middle-right, bottom) Koopman Flow Matching (KoFM) generates probabilistic forecasts by interpolating between noise and data and transporting samples through a matrix exponential conditioned on the KoPE latents. (Right) An Adaptive Uncertainty Gate (AUG) combines deterministic and probabilistic predictions for the final forecast. Extended details in App.~\ref{app:arch-details}.}
    \label{fig:architecture}
\end{figure}

\subsection{Modeling the Evolution of Structured Uncertainty}
\label{subsec:principle}

We denote by $\mathbf{x}_{1:L}$ the observed historical time series of length $L$, and by $\mathbf{x}_{L+1:L+H}$ the future horizon of length $H$. We formalize probabilistic time-series forecasting as the problem of modeling the predictive distribution of future observations given historical context:
\begin{equation}
    p(\mathbf{x}_{L+1:L+H} \mid \mathbf{x}_{1:L}) \ .
\end{equation}
This requires capturing not only the expected trajectory, but also the variability and structure of possible future outcomes. We propose to view probabilistic forecasting as modeling the \emph{evolution of structured uncertainty over time}. Under this perspective, future trajectories arise from underlying temporal dynamics that govern both the signal and its uncertainty, rather than being generated independently or through repeated sampling procedures. This view imposes three key requirements on an effective forecasting model: (i) \emph{efficient generation}, enabling direct sampling of future trajectories without iterative simulation; (ii) \emph{structured temporal evolution}, allowing the model to encode and extrapolate dynamics over long horizons; and (iii) \emph{adaptive uncertainty}, capturing the fact that predictability varies across variables and forecast horizons. These considerations motivate a unified approach that explicitly models forecasting as the evolution of structured uncertainty, jointly satisfying the above requirements within a single framework.

\subsection{Overview of KGO}
\label{subsec:overview}

We realize this perspective through \emph{KGO}, a unified framework for probabilistic forecasting that models prediction as the evolution of structured uncertainty. KGO consists of three main components: (i) a \emph{Koopman Flow Matching (KoFM)} module that enables efficient, single-step generation of future trajectories, (ii) a \emph{Koopman Patch Embedding (KoPE)} module that encodes and extrapolates temporal dynamics in a structured latent space, and (iii) an \emph{Adaptive Uncertainty Gate (AUG)} that dynamically balances deterministic and stochastic predictions according to the predictability of the data.

Although these components are conceptually introduced in terms of their roles, their implementation follows a different dependency structure. In particular, KGO first constructs a latent representation of the historical context, which is then extrapolated into the future by KoPE. The resulting sequence of latent states conditions KoFM, which generates probabilistic forecasts in a single step. Finally, AUG fuses the outputs of deterministic and probabilistic decoding streams to produce the final prediction. Formally, given an input time series $\mathbf{x}_{1:L}$, KGO first encodes it into a sequence of latent states $\mathbf{Z} = [\mathbf{z}_1, \dots, \mathbf{z}_N]$. KoPE then extrapolates the final latent $\mathbf{z}_N$ to produce future latent states $\tilde{\mathbf{Z}}_{\text{KoPE}} = [\tilde{\mathbf{z}}_{N+1}, \dots, \tilde{\mathbf{z}}_{N+M}]$. These latents serve as conditioning variables for KoFM, which maps noise samples to future trajectories, $\hat{\mathbf{x}}^{\text{prob}}_{L+1:L+H}$. In parallel, a deterministic decoder produces a point forecast, $\hat{\mathbf{x}}^{\text{det}}_{L+1:L+H}$, from the same latent trajectory via a residual MLP. The two outputs are combined by AUG to yield the final prediction. An overview of the architecture is illustrated in Fig.~\ref{fig:architecture}.

\subsection{Representation Learning}
\label{subsec:representation}

KGO models the evolution of structured uncertainty by first constructing a latent representation of the historical context. We apply reversible instance normalization (RevIN)~\citep{kim2021reversible} to mitigate non-stationarity, and partition the input $\mathbf{x}_{1:L}$ into $N=\lfloor L/P \rfloor$ non-overlapping patches of length $P$, which are projected into $D$-dimensional embeddings $\mathbf{X}^{P} \in \mathbb{R}^{N \times D}$~\citep{nie2023patchtst}. These embeddings are processed by an sLSTM encoder~\citep{beck2024xlstm}, which captures temporal dependencies while maintaining efficient recurrent computation. Compared to Transformer-based encoders, sLSTM provides lightweight state tracking with linear complexity, making it suitable for long-context forecasting. The encoder outputs latent states $\mathbf{Z} = [\mathbf{z}_1, \ldots, \mathbf{z}_N]$, where the final state $\mathbf{z}_N$ serves as a compact representation of the system state and initializes subsequent Koopman-based temporal evolution.

\subsection{Koopman Patch Embedding (KoPE)}
\label{subsec:kope}

A key requirement for modeling the evolution of structured uncertainty is the ability to represent and extrapolate temporal dynamics in a form that supports long-range forecasting. Standard conditioning strategies often rely on compressed representations of the past~\cite{nie2023patchtst, liu2025sundial}, which can limit the preservation and propagation of temporal structure across the prediction horizon. To address this, we introduce \emph{Koopman Patch Embedding (KoPE)}, which constructs a structured latent trajectory by modeling temporal evolution via a learned linear operator in latent space. This formulation is motivated by the Koopman perspective, where nonlinear dynamics can be represented through linear evolution in a lifted space, an approach that has been successfully exploited for sequence forecasting and generative modeling~\cite{azencot2020forecasting,naiman2024generative}, yielding future states by repeatedly applying the operator.

Given the final latent state $\mathbf{z}_N \in \mathbb{R}^D$ produced by the encoder (see Sec.~\ref{subsec:representation}), KoPE generates a sequence of future latent states by modeling temporal evolution via a learned linear operator. Specifically, we learn an operator $K \in \mathbb{R}^{D \times D}$ that approximates Koopman evolution, and obtain future states through repeated application:
\begin{equation}
\tilde{\mathbf{z}}_{N+i} = K^{i-1} \mathbf{z}_N \ , \qquad i = 1, \dots, M \ ,
\end{equation}
where $M = \lceil H / P \rceil$ is the number of patches covering the forecast horizon. To keep the rollout stable, $K$ is normalized by its spectral norm whenever this norm exceeds $1$, so that $\lVert K\rVert_2 \le 1$. This rollout propagates the latent state forward in time, preserving temporal structure across the horizon and yielding a structured latent trajectory $\tilde{\mathbf{Z}}_{\text{KoPE}} = [\tilde{\mathbf{z}}_{N+1}, \ldots, \tilde{\mathbf{z}}_{N+M}]$.

Importantly, unlike prior Koopman-based approaches where latent evolution directly defines the generative process~\cite{wang2024k2vae}, KoPE is not used to generate forecasts. Instead, it constructs a latent trajectory $\tilde{\mathbf{Z}}_{\text{KoPE}}$ over the forecast horizon that encodes temporal dynamics and serves as conditioning for the subsequent generative module. This decoupling separates temporal evolution from uncertainty modeling, allowing each to be handled independently and more flexibly.

\subsection{Koopman Flow Matching (KoFM)}
\label{subsec:kofm}

While KoPE provides a structured latent representation of future dynamics, a second requirement is to efficiently model the predictive distribution over future observations. Existing generative approaches, such as diffusion and flow matching~\cite{sohl2015deep, ho2020denoising, karras2022elucidating, lipman2023flow}, construct flexible distributions but rely on \emph{iterative sampling}, requiring multiple function evaluations at inference time.  To address this limitation, we propose \emph{Koopman Flow Matching (KoFM)}, a generative framework that replaces iterative transport with a \emph{closed-form transformation}. Building on the use of Koopman modeling for efficient generative transport~\cite{berman2026one}, KoFM directly parameterizes the transport using a linear dynamical system in a learned latent space, rather than learning a time-dependent velocity field that incrementally transports noise to data, enabling efficient one-step inference.

The flow matching formulation~\cite{lipman2023flow} defines a linear interpolation between noise and data:
\begin{equation} \label{eqn:flow_matching}
    \mathbf{x}^{(\tau)}_{N+i} = (1-\tau)\boldsymbol{\epsilon} + \tau \mathbf{x}_{N+i} \ ,
    \qquad \tau \sim \mathcal{U}[0,1] \ , \;\; \boldsymbol{\epsilon} \sim \mathcal{N}(\mathbf{0}, \mathbf{I}) \ ,
\end{equation}
where $\tau$ is the transport time. Sampling is performed by drawing noise $\boldsymbol{\epsilon}$ and transporting it along this path; multiple trajectories are obtained by sampling $\boldsymbol{\epsilon}$ independently. In practice, flow matching learns a time-dependent velocity field $\mathbf{v}_\theta(\mathbf{x}^{(\tau)}, \tau)$ that transports samples along this path from noise ($\tau=0$) to data ($\tau=1$) by matching the conditional displacement between endpoints. This induces a nonlinear flow defined by the ordinary differential equation $d\mathbf{x} / d\tau = \mathbf{v}_\theta(\mathbf{x}^{(\tau)}, \tau)$. In contrast, we restrict this transport to a \emph{time-independent} vector field in a learned latent space, corresponding to a finite-dimensional approximation of the Koopman generator~\cite{engel2000one}. This replaces the general nonlinear flow with a linear dynamical system whose evolution admits a closed-form solution.

KoFM generates future patches through lifting, latent transport, and decoding. Here, $i$ indexes patches along the forecast horizon, whereas $\tau\in[0,1]$ denotes transport time from noise to data. The interpolated input $\mathbf{x}^{(\tau)}_{N+i}$ is first lifted through a learned transformation conditioned on the KoPE latent $\tilde{\mathbf{z}}_{N+i}$, producing $\tilde{\mathbf{z}}^{(\tau)}_{N+i}$. KoFM then applies a closed-form affine transport:
\begin{equation} \label{eqn:latent_dynamics}
    \tilde{\mathbf{z}}^{(1)}_{N+i} = e^{(1-\tau)\mathbf{A}} \tilde{\mathbf{z}}^{(\tau)}_{N+i} + (1-\tau)\mathbf{b} \ .
\end{equation}
Here, $\mathbf{A}$ is a learned generator and $\mathbf{b}$ is a learned offset. The matrix exponential represents the exact homogeneous linear evolution, while the offset provides a time-scaled translation. A reconstruction head decodes the transported latents into forecast patches, which are concatenated to produce $\hat{\mathbf{x}}^{\text{prob}}_{L+1:L+H}$. At inference, $\tau=0$, so the cached matrix $e^{\mathbf{A}}$ enables single-step transport through a matrix product and bias addition. During training, $\tau\sim\mathcal{U}[0,1]$, and to improve optimization we mask $\tau$ to $0$ for each sample with probability $0.5$. App.~\ref{app:exact_koopman_integration} details the affine correction and training-time computation.

\subsection{Adaptive Uncertainty Gate (AUG)}
\label{subsec:aug}

While KoFM enables efficient generation of probabilistic forecasts, a final requirement is to account for the heterogeneous nature of uncertainty. In time-series forecasting, predictability varies across variables and forecast horizons, implying that the appropriate balance between deterministic and stochastic components should be adaptive rather than fixed~\cite{li2025d3u, li2024tmdm}. To this end, we combine two complementary prediction streams. A deterministic decoder maps the KoPE latent trajectory $\tilde{\mathbf{Z}}_{\text{KoPE}}$ to a point forecast $\hat{\mathbf{x}}^{\text{det}}_{L+1:L+H}$, capturing the predictable component of the signal, while KoFM produces a stochastic prediction $\hat{\mathbf{x}}^{\text{prob}}_{L+1:L+H}$ that models the full predictive distribution.

We introduce an \emph{Adaptive Uncertainty Gate (AUG)} that dynamically combines these two components. Specifically, for each horizon step $h$, we compute a gating coefficient $g^h \in [0,1]$ based on the historical context, and define the final prediction as:
\begin{equation}
    \label{eq:aug_prediction}
    \hat{x}_{L+h} = (1-g^h)\,\hat{x}^{\text{det}}_{L+h} +  g^h\,\hat{x}^{\text{prob}}_{L+h} \ .
\end{equation}
The gating coefficients are produced by a lightweight encoder--router network that maps the input context to per-(variate, horizon) weights, allowing the model to allocate uncertainty dynamically. This enables the model to rely more on deterministic predictions when the signal is predictable, and increase the contribution of the probabilistic component in uncertain regimes. AUG learns to modulate predictive dispersion across variables and forecast horizons, supporting adaptive modeling of aleatoric uncertainty. Appendix~\ref{subsec:theory-gate} characterizes this effect under the prediction rule in Eq.~\eqref{eq:aug_prediction}.

\subsection{Training Objective}

KGO is trained end-to-end on the gate-fused output. For each input, we draw \(J\) samples
\(\{\hat{\mathbf{x}}^{(j)}_{L+1:L+H}\}_{j=1}^J\) and optimize a CRPS-based objective composed of two complementary terms:
\begin{equation}
\mathcal{L}
=
\frac{1}{2}
\underbrace{
\Big(
\mathbb{E}\|\hat{X} - X^{\star}\|_1
-
\frac{1}{2}\mathbb{E}\|\hat{X} - \hat{X}'\|_1
\Big)
}_{\text{Energy Score}}
+
\frac{1}{2|Q|}
\sum_{q \in Q}
\rho_q\!\big(X^{\star} - \hat{X}_q\big).
\end{equation}
The first term corresponds to the \emph{Energy Score} \citep{gneiting2007strictly, chen2024generative} formulation of CRPS, which combines an \(L_1\) accuracy term with a repulsion term that discourages sample collapse. Here, \(\hat{X}\) and \(\hat{X}'\) denote independent samples from the model; in practice, \(\hat{X}'\) is created by randomly shuffling the samples in \(\hat{X}\). Optimizing this objective for a conditional generative model enables learning forecasts that avoid systematic bias and dispersion errors. An ablation study demonstrating the effectiveness of this component is provided in App.~\ref{app:loss_function}. The second term is the \emph{CRPS quantile loss}~\cite{salinas2020deepar, rasul2021timegrad}, computed over a set of quantile levels \(Q\), where \(\hat{X}_q\) denotes the empirical \(q\)-quantile and \(\rho_q(u) = u(q - \mathbb{I}[u < 0])\) is the pinball loss. This term directly supervises predictive quantiles across probability levels, complementing the sample-based term in learning the location and spread of the predictive distribution.

\section{Experiments}
\label{sec:experiments}

\vspace{-2mm}
\subsection{Experimental Setup}
\label{subsec:exp-setup}
\vspace{-2mm}

We evaluate KGO on the ProbTS benchmark~\citep{zhang2024probts} across short- and long-horizon probabilistic forecasting. Experiments cover $17$ data configurations from standard benchmarks (ETT, Electricity, Exchange, Traffic, Weather, Solar, ILI) under the official protocol. In the short-term regime, context length equals the horizon ($24$--$30$), while in the long-term regime, the context length is fixed at 96 and performance is averaged over horizons $\{96,192,336,720\}$; for ILI, the context length is 36 and the horizons are $\{24,36,48,60\}$. We compare against eight baselines spanning major probabilistic forecasting paradigms: flow-based models (GRU~MAF, Trans~MAF, GRU~NVP)~\citep{rasul2021tmaf}, diffusion models (TimeGrad, TSDiff, CSDI)~\citep{rasul2021timegrad, kollovieh2024tsdiff, tashiro2021csdi}, and variational autoencoder-based models ($K^2$VAE, $D^3$VAE)~\citep{wang2024k2vae, li2022d3vae}. These cover autoregressive, flow, diffusion, and latent-dynamics-based designs, providing a comprehensive comparison across model families. We follow the ProbTS evaluation pipeline using CRPS for probabilistic accuracy and NMAE for point forecasting (lower is better), reporting averages over 5 independent runs. Additional details on datasets, metrics, and implementation are provided in App.~\ref{app:training-details}--\ref{app:metrics}. Extended results, ablation studies, and interpretability analyses are provided in Appendices~\ref{app:extended_results}--\ref{app:analysis}.

\begingroup
\setlength{\intextsep}{6pt}
\par\vspace{\intextsep}\noindent
\begin{minipage}{\linewidth}
\centering
\captionsetup{type=table}
\caption{\textbf{Long-term forecasting.} Results across nine datasets, averaged over horizons and five independent runs (Sec.~\ref{subsec:exp-setup}). Lower CRPS/NMAE is better. \best{Red bold}: best; \second{blue underline}: second best; ``--'': unavailable. Per-horizon results and standard deviations appear in Tables~\ref{tab:full:ltsf:crps}--\ref{tab:full:ltsf:nmae}.}
\label{tab:ltsf}
\begingroup
\setlength{\tabcolsep}{2pt}
\begin{adjustbox}{width=1.0\textwidth, center}
\begin{tabular}{@{}l *{14}{c} *{2}{>{\columncolor{black!5}}c}@{}}
\toprule
\multicolumn{1}{l}{Dataset} & \multicolumn{2}{c}{TSDiff} & \multicolumn{2}{c}{CSDI} & \multicolumn{2}{c}{TimeGrad} & \multicolumn{2}{c}{GRU MAF} & \multicolumn{2}{c}{Trans MAF} & \multicolumn{2}{c}{GRU NVP} & \multicolumn{2}{c}{$K^2$VAE} & \multicolumn{2}{>{\columncolor{black!5}}c@{}}{\textbf{KGO (Ours)}} \\
\cmidrule(lr){2-3}\cmidrule(lr){4-5}\cmidrule(lr){6-7}\cmidrule(lr){8-9}\cmidrule(lr){10-11}\cmidrule(lr){12-13}\cmidrule(lr){14-15}\cmidrule(lr){16-17}
\multicolumn{1}{l}{Metric}  & CRPS & NMAE & CRPS & NMAE & CRPS & NMAE & CRPS & NMAE & CRPS & NMAE & CRPS & NMAE & CRPS & NMAE & CRPS & NMAE \\
\midrule
ETTm1-L        & 0.407 & 0.519 & 0.324 & 0.420 & 0.587 & 0.736 & 0.412 & 0.528 & 0.476 & 0.605 & 0.453 & 0.585 & \second{0.262} & \second{0.328} & \best{0.244} & \best{0.308} \\
ETTm2-L        & 0.275 & 0.338 & 0.173 & 0.222 & 0.447 & 0.545 & 0.309 & 0.377 & 0.272 & 0.339 & 0.414 & 0.542 & \second{0.165} & \second{0.194} & \best{0.138} & \best{0.173} \\
ETTh1-L        & 0.457 & 0.586 & 0.479 & 0.603 & 0.502 & 0.651 & 0.353 & 0.440 & 0.354 & 0.453 & 0.441 & 0.559 & \second{0.294} & \second{0.374} & \best{0.270} & \best{0.347} \\
ETTh2-L        & 0.339 & 0.417 & 0.241 & 0.311 & 0.435 & 0.545 & 0.480 & 0.553 & 0.282 & 0.358 & 0.597 & 0.736 & \second{0.221} & \second{0.236} & \best{0.160} & \best{0.204} \\
Electricity-L  & 0.407 & 0.519 & -- & -- & 0.101 & 0.126 & 0.094 & 0.121 & -- & -- & 0.101 & 0.127 & \best{0.066} & \second{0.105} & \second{0.080} & \best{0.095} \\
Exchange-L     & 0.084 & 0.111 & 0.069 & 0.086 & 0.082 & 0.095 & 0.070 & 0.083 & 0.067 & 0.086 & 0.073 & 0.093 & \second{0.045} & \second{0.053} & \best{0.039} & \best{0.048} \\
Traffic-L      & 0.327 & 0.392 & -- & -- & 0.211 & 0.245 & -- & -- & -- & -- & 0.198 & 0.245 & \best{0.142} & \second{0.238} & \second{0.193} & \best{0.223} \\
Weather-L      & 0.132 & 0.134 & \second{0.077} & 0.093 & 0.125 & 0.155 & 0.133 & 0.165 & 0.123 & 0.154 & 0.119 & 0.147 & 0.081 & \second{0.090} & \best{0.060} & \best{0.074} \\
ILI-L          & 0.248 & 0.259 & 0.276 & 0.290 & 0.284 & 0.310 & 0.129 & 0.161 & 0.124 & 0.151 & 0.283 & 0.309 & \second{0.116} & \second{0.144} & \best{0.097} & \best{0.118} \\
\hline
\rowcolor{black!3} $1^{st}$ \textbf{Count} & 0 & 0 & 0 & 0 & 0 & 0 & 0 & 0 & 0 & 0 & 0 & 0 & 2 & 0 & \textbf{7} & \textbf{9} \\
\rowcolor{black!3} $2^{nd}$ \textbf{Count} & 0 & 0 & 1 & 0 & 0 & 0 & 0 & 0 & 0 & 0 & 0 & 0 & 6 & 9 & 2 & 0 \\
\hline
\end{tabular}
\end{adjustbox}
\endgroup
\end{minipage}\par\vspace{\intextsep}

\par\vspace{\intextsep}\noindent
\begin{minipage}{\linewidth}
\centering
\captionsetup{type=table}
\caption{\textbf{Short-term forecasting.} Results across eight datasets, averaged over five independent runs. Lower CRPS/NMAE is better. \best{Red bold}: best; \second{blue underline}: second best. Full results and standard deviations appear in Tables~\ref{tab:full:stsf_CRPS}--\ref{tab:full:stsf_NMAE}.}
\label{tab:stsf}
\begingroup
\setlength{\tabcolsep}{2pt}
\begin{adjustbox}{width=1.0\textwidth, center}
\begin{tabular}{@{}l *{14}{c} *{2}{>{\columncolor{black!5}}c}@{}}
\toprule
\multicolumn{1}{l}{Dataset} & \multicolumn{2}{c}{TSDiff} & \multicolumn{2}{c}{$D^3$VAE} & \multicolumn{2}{c}{CSDI} & \multicolumn{2}{c}{TimeGrad} & \multicolumn{2}{c}{GRU MAF} & \multicolumn{2}{c}{Trans MAF} & \multicolumn{2}{c}{$K^2$VAE} & \multicolumn{2}{>{\columncolor{black!5}}c@{}}{\textbf{KGO (Ours)}} \\
\cmidrule(lr){2-3}\cmidrule(lr){4-5}\cmidrule(lr){6-7}\cmidrule(lr){8-9}\cmidrule(lr){10-11}\cmidrule(lr){12-13}\cmidrule(lr){14-15}\cmidrule(lr){16-17}
\multicolumn{1}{l}{Metric} & CRPS & NMAE & CRPS & NMAE & CRPS & NMAE & CRPS & NMAE & CRPS & NMAE & CRPS & NMAE & CRPS & NMAE & CRPS & NMAE \\
\midrule
ETTm1-S & 0.209 & 0.276 & 0.198 & 0.250 & 0.140 & 0.169 & 0.186 & 0.246 & 0.151 & 0.198 & 0.139 & 0.162 & \second{0.135} & \second{0.152} & \best{0.104} & \best{0.133} \\
ETTm2-S & 0.124 & 0.162 & 0.303 & 0.378 & 0.144 & 0.181 & 0.148 & 0.189 & 0.146 & 0.193 & 0.180 & 0.224 & \second{0.122} & \second{0.146} & \best{0.092} & \best{0.118} \\
ETTh1-S & 0.304 & 0.400 & 0.324 & 0.410 & 0.262 & 0.339 & 0.273 & 0.356 & 0.258 & 0.326 & 0.309 & 0.400 & \second{0.256} & \second{0.312} & \best{0.228} & \best{0.280} \\
ETTh2-S & 0.204 & 0.272 & 0.216 & 0.267 & 0.133 & 0.161 & 0.184 & 0.224 & 0.160 & 0.208 & 0.200 & 0.256 & \second{0.128} & \second{0.140} & \best{0.104} & \best{0.130} \\
Electricity-S & 0.111 & 0.115 & 0.071 & 0.092 & \best{0.051} & \second{0.066} & 0.057 & 0.073 & 0.056 & 0.073 & 0.054 & 0.071 & 0.053 & 0.068 & \second{0.052} & \best{0.064} \\
Exchange-S & 0.077 & 0.096 & 0.011 & \second{0.012} & \second{0.009} & 0.013 & \second{0.009} & \second{0.012} & 0.012 & 0.016 & 0.012 & 0.016 & \second{0.009} & \best{0.009} & \best{0.007} & \best{0.009} \\
Traffic-S & 0.189 & 0.206 & 0.143 & 0.178 & 0.147 & 0.175 & 0.130 & \second{0.155} & 0.144 & 0.182 & 0.133 & 0.160 & \second{0.129} & 0.157 & \best{0.127} & \best{0.152} \\
Solar-S & 0.568 & 0.635 & 0.769 & 0.998 & \second{0.392} & \second{0.533} & 0.465 & 0.609 & 0.486 & 0.603 & 0.442 & 0.577 & \best{0.367} & \best{0.480} & 0.456 & 0.569 \\
\midrule
\rowcolor{black!3} $1^{st}$ \textbf{Count} & 0 & 0 & 0 & 0 & 1 & 0 & 0 & 0 & 0 & 0 & 0 & 0 & 1 & 2 & \textbf{6} & \textbf{7} \\
\rowcolor{black!3} $2^{nd}$ \textbf{Count} & 0 & 0 & 0 & 1 & 2 & 2 & 1 & 2 & 0 & 0 & 0 & 0 & 6 & 4 & 1 & 0 \\
\bottomrule
\end{tabular}
\end{adjustbox}
\endgroup
\end{minipage}\par\vspace{\intextsep}

\endgroup

\vspace{-2mm}
\subsection{Main Results}
\label{subsec:exp-main}
\vspace{-2mm}

Tables~\ref{tab:ltsf} and~\ref{tab:stsf} demonstrate that KGO achieves the lowest CRPS on 13/17 evaluated settings and the lowest NMAE on 16/17, highlighting its ability to produce accurate predictive distributions (CRPS) while maintaining superior point-forecasting accuracy (NMAE). In the long-term forecasting regime, KGO is particularly dominant, ranking first on 7/9 datasets in CRPS and 9/9 in NMAE. By comparison, the closest competitor, $K^2$VAE, attains only 2/9 first-place results in CRPS and none in NMAE. Notable gains are observed on ETTh2-L, where KGO reduces CRPS by more than 27\% and NMAE by more than 13\% relative to $K^2$VAE. Similarly, on Weather-L, KGO reduces CRPS by approximately 26\% and NMAE by approximately 18\% relative to $K^2$VAE. In the short-term forecasting regime, KGO also delivers the best overall performance, obtaining 13 of the 16 best scores across CRPS and NMAE, compared with 3 for $K^2$VAE, the next strongest baseline. The largest gains are on the ETT datasets, where KGO reduces CRPS relative to $K^2$VAE by approximately 23\% on ETTm1-S, 25\% on ETTm2-S, and 19\% on ETTh2-S. Overall, these findings position KGO as a leading method for probabilistic time-series forecasting across the evaluated benchmarks. In particular, KGO substantially outperforms prior flow-based approaches such as Trans~MAF, while also surpassing stronger generative baselines such as $K^2$VAE.

\vspace{-2mm}
\subsection{Efficiency}
\label{subsec:ablation_efficiency}
\vspace{-2mm}

\begin{figure}[!h]
    \centering
    \includegraphics[width=1.0\linewidth]{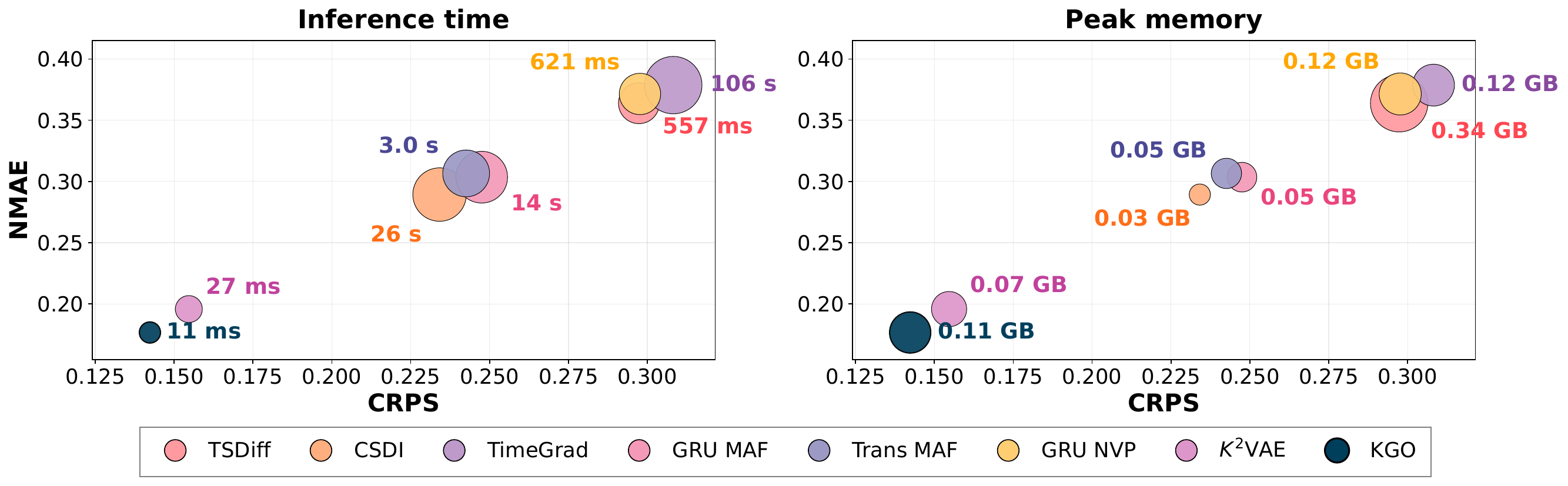}
    \caption{Accuracy--efficiency trade-off on the (CRPS, NMAE) plane. \emph{Left:} bubble size encodes inference latency per window (log scale), generating 100 forecast trajectories. \emph{Right:} bubble size encodes peak GPU memory. Each method is averaged over the datasets for which its accuracy is available in Table~\ref{tab:ltsf}.}
    \label{fig:ablation-efficiency}
\end{figure}

To assess the trade-off between accuracy and efficiency, we benchmark KGO against the same ProbTS baselines on forecast quality (CRPS, NMAE), inference latency, and peak GPU memory. All models are evaluated with matched context/horizon lengths, batch size~1, FP32 precision, and 100 forecast trajectories per window on the same hardware. Figure~\ref{fig:ablation-efficiency} shows the (CRPS, NMAE) plane with bubble size encoding inference latency (left) and peak memory (right); full numbers and the measurement protocol are in App.~\ref{app:ablation:efficiency}.
KGO occupies the lower-left corner of the (CRPS, NMAE) plane and has the lowest average latency in the panel: 11 ms/window versus 27 ms/window for $K^{2}$VAE, a $2.5\times$ speedup, averaged over 36 dataset--horizon pairs. Its average peak GPU memory is 0.11 GB, above $K^{2}$VAE (0.07 GB) and the CSDI and MAF baselines (0.03--0.05 GB), but below TimeGrad and GRU NVP (0.12 GB) and TSDiff (0.34 GB).

KGO is channel-independent, processing variates in parallel, while future patches and generated samples are batched within the decoder. For fixed patch size and latent dimensions, inference computation and decoder activation memory scale linearly in each of these axes when all samples are processed together. The sLSTM encoder likewise has linear complexity in the number of context patches, while KoFM generates samples in one shot through a matrix-exponential transport, avoiding iterative sampling. Batching exploits available GPU parallelism and can limit latency growth until hardware saturation, after which runtime increases with the computational workload. This parallelism comes with a memory trade-off: processing more trajectories simultaneously increases peak activation memory.

\vspace{-2mm}
\subsection{Foundation Models}
\label{subsec:ablation_foundation}
\vspace{-2mm}

We compare full-shot KGO against three off-the-shelf zero-shot foundation models, TiRex~\citep{auerschmid2025tirex}, Chronos-2~\citep{ansari2025chronos}, and Sundial~\citep{liu2025sundial}, under the same ProbTS evaluator; full protocol and the extended table are in App.~\ref{app:foundation_models}.
The results in Tab.~\ref{tab:ablation-foundation} show a clear and consistent advantage for KGO over the foundation-model baselines. Across most datasets, KGO achieves the lowest CRPS and NMAE, ranking first on 5/6 datasets for both metrics, while the baselines rarely take the top spot. The gains are particularly pronounced on Weather and the ETT variants, reflecting improvements in both probabilistic and point-forecasting accuracy. A minor exception is Exchange, where Chronos-2 and TiRex slightly outperform KGO. Overall, despite the growing popularity of these foundation models for zero-shot forecasting, the results demonstrate that a full-shot approach, training directly on the target data, remains highly effective, especially in probabilistic forecasting where precise distributional modeling is critical.

\begin{table}[!htbp]
\centering
\caption{\textbf{Foundation-model comparison.} Per-dataset CRPS/NMAE for KGO (trained from scratch) versus three zero-shot time-series foundation models, averaged across horizons $\{96,192,336,720\}$ at context $L{=}96$, seed $0$. Lower is better. \best{Red bold}: best; \second{blue underline}: second best; bottom rows count per-dataset rankings.}
\label{tab:ablation-foundation}
\small
\setlength{\tabcolsep}{4pt}
\begin{adjustbox}{width=.70\linewidth}
\begin{tabular}{@{}l *{6}{c} *{2}{>{\columncolor{black!5}}c}@{}}
\toprule
Dataset & \multicolumn{2}{c}{TiRex} & \multicolumn{2}{c}{Chronos-2} & \multicolumn{2}{c}{Sundial} & \multicolumn{2}{>{\columncolor{black!5}}c@{}}{\textbf{KGO (Ours)}} \\
\cmidrule(lr){2-3} \cmidrule(lr){4-5} \cmidrule(lr){6-7} \cmidrule(lr){8-9}
Metric & CRPS & NMAE & CRPS & NMAE & CRPS & NMAE & CRPS & NMAE \\
\midrule
ETTh1 & 0.291 & 0.375 & \second{0.274} & 0.356 & 0.296 & \second{0.347} & \best{0.269} & \best{0.345} \\
ETTh2 & 0.169 & \second{0.216} & \second{0.168} & \second{0.216} & 0.186 & \second{0.216} & \best{0.159} & \best{0.203} \\
ETTm1 & 0.313 & 0.381 & 0.291 & 0.366 & \second{0.285} & \second{0.344} & \best{0.244} & \best{0.307} \\
ETTm2 & 0.151 & 0.187 & \second{0.147} & \second{0.181} & 0.158 & 0.186 & \best{0.138} & \best{0.174} \\
Weather & 0.109 & 0.131 & 0.095 & 0.115 & \second{0.087} & \second{0.102} & \best{0.060} & \best{0.075} \\
Exchange & \second{0.039} & \second{0.046} & \best{0.037} & \best{0.045} & 0.045 & 0.049 & 0.040 & 0.048 \\
\hline
\rowcolor{black!3} $1^{st}$ \textbf{Count} & 0 & 0 & 1 & 1 & 0 & 0 & \textbf{5} & \textbf{5} \\
\rowcolor{black!3} $2^{nd}$ \textbf{Count} & 1 & 2 & 3 & 2 & 2 & 4 & 0 & 0 \\
\hline
\end{tabular}
\end{adjustbox}
\end{table}

\FloatBarrier

\section{Ablation Study}
\label{sec:ablation}

\vspace{-2mm}
\subsection{Representation and Architecture}
\label{subsec:ablation_architecture}
\vspace{-2mm}

\begin{figure}[!h]
    \centering
    \includegraphics[width=\linewidth]{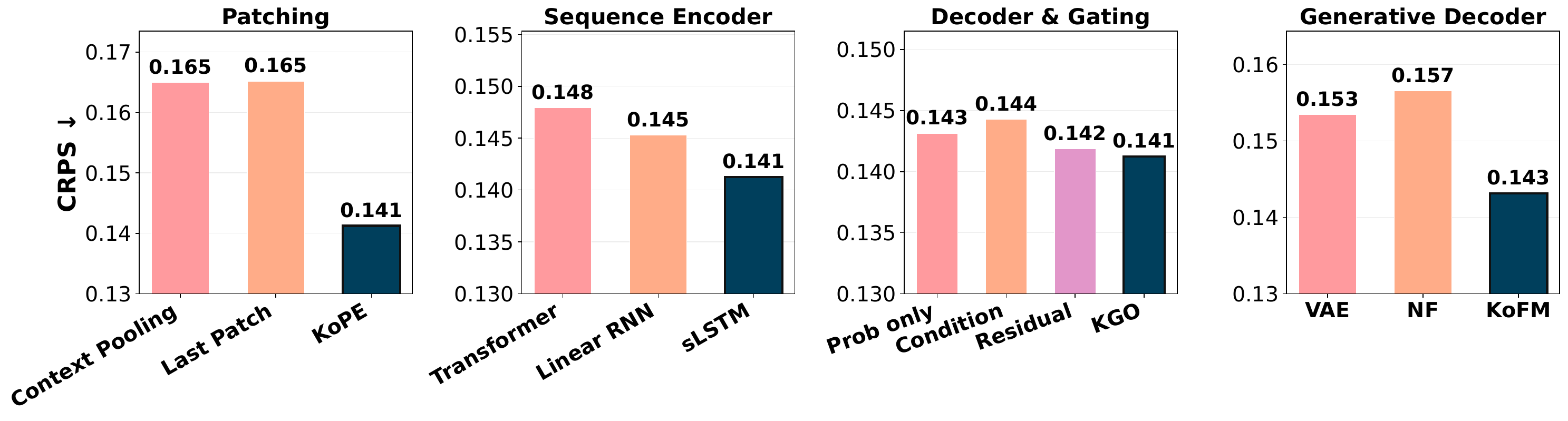}
    \vspace{-5mm}
    \caption{Architecture ablations averaged across datasets and forecasting horizons (lower CRPS is better). Panels evaluate the four core design choices underlying KGO: latent trajectory representation (KoPE), sequence encoder, deterministic--probabilistic coupling (AUG), and generative decoder (KoFM; each head used alone, without the gate or deterministic branch). Koopman-based latent rollout, sLSTM encoding, adaptive gating, and closed-form Koopman transport yield the lowest average CRPS. Full results appear in Appendices~\ref{app:patching_ablation}--\ref{app:generative_decoder}.}
    \label{fig:ablation-arch}
\end{figure}

In this section, we ablate the three main components, KoPE, KoFM, and AUG, as well as the backbone choice (sLSTM), reporting CRPS averaged over datasets and forecasting horizons (Fig.~\ref{fig:ablation-arch}). Across 36 dataset--horizon pairs, KoPE achieves lower average CRPS (0.141) than both the \textit{Last Patch} baseline (0.165) and the \textit{Context Pooling} head (0.165). The two non-Koopman alternatives perform nearly identically, indicating that the gain arises from the Koopman-based representation propagated to the decoder. Consistent with this, the dominant eigenvalues of the learned KoPE operator recover the daily and weekly cycles of the data (App.~\ref{app:spectral_analysis}). On the encoder side, holding depth and width fixed, replacing the \textit{sLSTM} backbone with either a \textit{Transformer} or a \textit{Linear RNN} degrades average CRPS (0.141 vs. 0.148 and 0.145, respectively), suggesting that recurrent backbones are better suited than attention-based alternatives in this setting. Combining the deterministic and probabilistic streams through AUG gives the lowest average CRPS among the coupling variants (0.141, vs. 0.144 for conditioning KoFM on the deterministic prediction and 0.142 for fixed residual addition), although the margins are small. AUG thus matches fixed coupling rules in accuracy while adaptively modeling aleatoric uncertainty, learning the contribution of the stochastic branch per variable and forecast horizon (App.~\ref{app:uncertainty_gate}). Finally, with each generative head used alone (no gate or deterministic branch) and trained with its native loss, replacing KoFM with either a single-step \textit{VAE} or a \textit{normalizing flow} head increases average CRPS by 7.2\% and 9.4\%, respectively, demonstrating that \textit{KoFM}, the closed-form Koopman push-forward, produces a stronger predictive distribution than standard one-shot generative heads.

\vspace{-2mm}
\subsection{Flow Matching}
\label{subsec:ablation_flow_matching}
\vspace{-2mm}

\begin{figure}[!htbp]
    \centering
    \includegraphics[width=\linewidth]{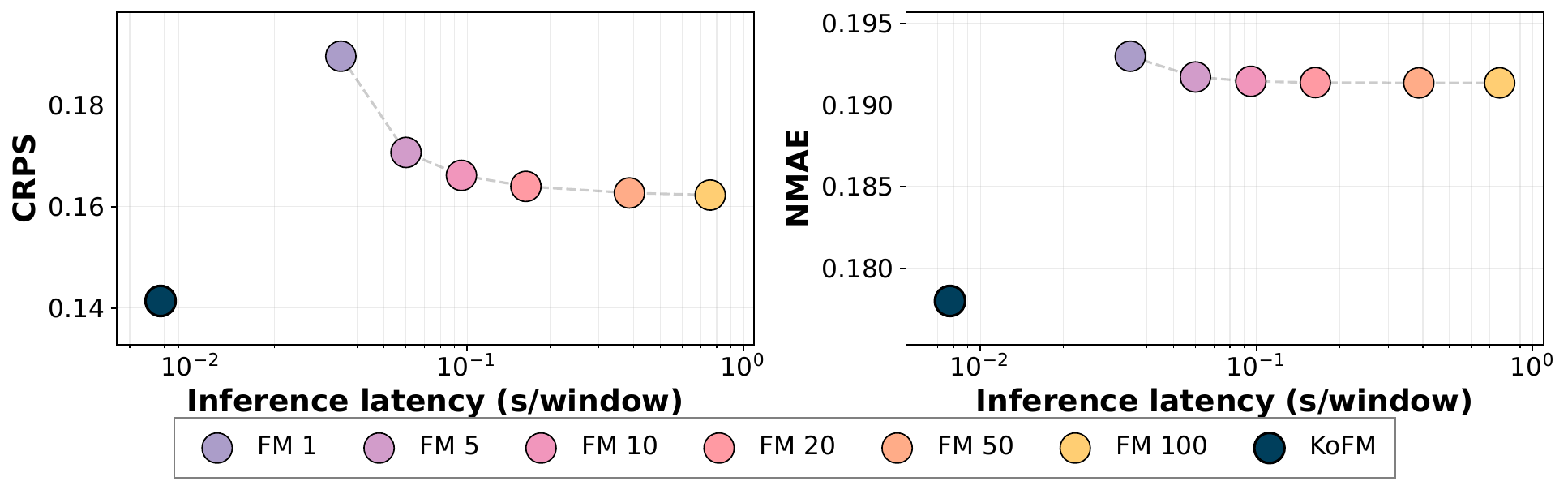}
    \caption{KoFM vs Sundial-base-128m ($N$-step Euler ODE): aggregate CRPS (left) and NMAE (right) vs inference latency (s/window, logarithmic scale). Each forecast call generates 100 trajectories per multivariate window.}
    \label{fig:ablation-flow-matching}
\end{figure}

To investigate KoFM's ability to deliver accurate predictions while keeping inference efficient, we compare KGO against Sundial-base-128m~\cite{liu2025sundial}, a strong pretrained flow-matching foundation model, by sweeping its Euler-ODE step budget $N$ and tracing the resulting accuracy--efficiency trade-off in CRPS, NMAE, and inference latency per window (Fig.~\ref{fig:ablation-flow-matching}). KGO occupies the lower-left of the trade-off plane: at the budget $N{=}50$ adopted in the original Sundial paper, KGO achieves $\sim$13\% lower CRPS and $\sim$7\% lower NMAE, with average latency of 7.8 ms/window versus 387 ms/window for Sundial, a $\sim$50$\times$ speedup. These averages cover 28 dataset--horizon pairs, with each call generating 100 forecast trajectories (per-dataset results in Appendix~\ref{app:flow_matching}).

\FloatBarrier
\section{Conclusion}
\label{sec:conclusion}

We presented the Koopman Generative Operator (KGO), a probabilistic forecasting framework that reinterprets prediction as the evolution of structured uncertainty in a Koopman latent space. By combining Koopman Patch Embedding (KoPE) for consistent latent trajectory extrapolation, Koopman Flow Matching (KoFM) for closed-form, single-step generation, and an Adaptive Uncertainty Gate (AUG) for adaptive, horizon- and variable-aware uncertainty modeling, KGO unifies structure, expressivity, and efficiency within a single model. Empirically, this design yields state-of-the-art accuracy on 29/34 ProbTS configurations, in both short- and long-term forecasting, while reducing inference cost by over an order of magnitude compared to iterative generative methods. These results highlight the advantage of embedding generative modeling within linearizable latent dynamics, enabling both principled uncertainty propagation and scalable inference. More broadly, KGO suggests a new perspective on probabilistic forecasting: rather than treating generation and dynamics as separate processes, they can be jointly governed through operator-theoretic evolution. An important direction for future work is to develop a unified Koopman framework that simultaneously models evolution in physical time and transport time, further bridging dynamical systems and generative modeling.

\section*{Acknowledgements}

This research was partially supported by the Lynn and William Frankel Center of The Stein Faculty of Computer and Information Science, Ben-Gurion University of the Negev, and by an ISF grant 1299/25.

\newpage
\bibliography{references}

@article{catt2026forecastability,
  title={Forecastability as an Information-Theoretic Limit on Prediction},
  author={Catt, Peter Maurice},
  journal={arXiv preprint arXiv:2603.27074},
  year={2026}
}

@inproceedings{kim2021reversible,
  author       = {Taesung Kim and
                  Jinhee Kim and
                  Yunwon Tae and
                  Cheonbok Park and
                  Jang{-}Ho Choi and
                  Jaegul Choo},
  title        = {Reversible Instance Normalization for Accurate Time-Series Forecasting
                  against Distribution Shift},
  booktitle    = {The Tenth International Conference on Learning Representations, {ICLR}},
  year={2022}
}

@inproceedings{beck2024xlstm,
  title        = {{xLSTM}: Extended Long Short-Term Memory},
  author       = {Maximilian Beck and
                  Korbinian P{\"{o}}ppel and
                  Markus Spanring and
                  Andreas Auer and
                  Oleksandra Prudnikova and
                  Michael Kopp and
                  G{\"{u}}nter Klambauer and
                  Johannes Brandstetter and
                  Sepp Hochreiter},
  booktitle    = {Advances in Neural Information Processing Systems 38},
  year = {2024}
}

@inproceedings{liu2024koopa,
  title        = {Koopa: Learning Non-stationary Time Series Dynamics with {Koopman} Predictors},
  author       = {Yong Liu and
                  Chenyu Li and
                  Jianmin Wang and
                  Mingsheng Long},
  booktitle    = {Advances in Neural Information Processing Systems 36},
  year={2023}
}

@inproceedings{wang2024k2vae,
  author       = {Xingjian Wu and
                  Xiangfei Qiu and
                  Hongfan Gao and
                  Jilin Hu and
                  Bin Yang and
                  Chenjuan Guo},
  title        = {{K2VAE:} {A} Koopman-Kalman Enhanced Variational AutoEncoder for Probabilistic Time Series Forecasting},
  booktitle    = {Forty-second International Conference on Machine Learning, {ICML}},
  year={2025}
}

@inproceedings{zhang2024probts,
  title        = {{ProbTS}: Benchmarking Point and Distributional Forecasting across Diverse Prediction Horizons},
  author       = {Jiawen Zhang and
                  Xumeng Wen and
                  Zhenwei Zhang and
                  Shun Zheng and
                  Jia Li and
                  Jiang Bian},
  booktitle    = {Advances in Neural Information Processing Systems 38},
  year={2024}
}

@inproceedings{rasul2021timegrad,
  title        = {Autoregressive Denoising Diffusion Models for Multivariate Probabilistic Time Series Forecasting},
  author       = {Kashif Rasul and
                  Calvin Seward and
                  Ingmar Schuster and
                  Roland Vollgraf},
  booktitle    = {Proceedings of the 38th International Conference on Machine Learning, {ICML}},
  year={2021}
}

@inproceedings{rasul2021tmaf,
  title        = {Multivariate Probabilistic Time Series Forecasting via Conditioned Normalizing Flows},
  author       = {Kashif Rasul and
                  Abdul{-}Saboor Sheikh and
                  Ingmar Schuster and
                  Urs M. Bergmann and
                  Roland Vollgraf},
  booktitle    = {9th International Conference on Learning Representations, {ICLR}},
  year={2021}
}

@inproceedings{tashiro2021csdi,
  title        = {{CSDI:} Conditional Score-based Diffusion Models for Probabilistic Time Series Imputation},
  author       = {Yusuke Tashiro and
                  Jiaming Song and
                  Yang Song and
                  Stefano Ermon},
  booktitle    = {Advances in Neural Information Processing Systems 34},
  year={2021}
}

@inproceedings{kollovieh2024tsdiff,
  title        = {Predict, Refine, Synthesize: Self-Guiding Diffusion Models for Probabilistic Time Series Forecasting},
  author       = {Marcel Kollovieh and
                  Abdul Fatir Ansari and
                  Michael Bohlke{-}Schneider and
                  Jasper Zschiegner and
                  Hao Wang and
                  Yuyang Wang},
  booktitle    = {Advances in Neural Information Processing Systems 36},
  year={2023}
}

@inproceedings{li2025d3u,
  title        = {Diffusion-based Decoupled Deterministic and Uncertain Framework for Probabilistic Multivariate Time Series Forecasting},
  author       = {Qi Li and
                  Zhenyu Zhang and
                  Lei Yao and
                  Zhaoxia Li and
                  Tianyi Zhong and
                  Yong Zhang},
  booktitle    = {The Thirteenth International Conference on Learning Representations, {ICLR}},
  year={2025}
}

@inproceedings{li2024tmdm,
  title        = {Transformer-Modulated Diffusion Models for Probabilistic Multivariate Time Series Forecasting},
  author       = {Yuxin Li and
                  Wenchao Chen and
                  Xinyue Hu and
                  Bo Chen and
                  Baolin Sun and
                  Mingyuan Zhou},
  booktitle    = {The Twelfth International Conference on Learning Representations, {ICLR}},
  year={2024}
}

@inproceedings{nie2023patchtst,
  title        = {A Time Series is Worth 64 Words: Long-term Forecasting with Transformers},
  author       = {Yuqi Nie and
                  Nam H. Nguyen and
                  Phanwadee Sinthong and
                  Jayant Kalagnanam},
  booktitle    = {The Eleventh International Conference on Learning Representations, {ICLR}},
  year={2023}
}

@inproceedings{zeng2023dlinear,
  title        = {Are Transformers Effective for Time Series Forecasting?},
  author       = {Ailing Zeng and
                  Muxi Chen and
                  Lei Zhang and
                  Qiang Xu},
  booktitle    = {Thirty-Seventh {AAAI} Conference on Artificial Intelligence, {AAAI}},
  year={2023}
}

@inproceedings{wu2021autoformer,
  title        = {Autoformer: Decomposition Transformers with Auto-Correlation for Long-Term
                  Series Forecasting},
  author       = {Haixu Wu and
                  Jiehui Xu and
                  Jianmin Wang and
                  Mingsheng Long},
  booktitle    = {Advances in Neural Information Processing Systems 34},
  year={2021}
}

@inproceedings{liu2025sundial,
  author       = {Yong Liu and
                  Guo Qin and
                  Zhiyuan Shi and
                  Zhi Chen and
                  Caiyin Yang and
                  Xiangdong Huang and
                  Jianmin Wang and
                  Mingsheng Long},
  title        = {Sundial: {A} Family of Highly Capable Time Series Foundation Models},
  booktitle    = {Forty-second International Conference on Machine Learning, {ICML}},
  series       = {Proceedings of Machine Learning Research},
  year         = {2025},
}

@article{auerschmid2025tirex,
  title={{TiRex}: Zero-shot forecasting across long and short horizons with enhanced in-context learning},
  author={Auer, Andreas and Podest, Patrick and Klotz, Daniel and B{\"o}ck, Sebastian and Klambauer, G{\"u}nter and Hochreiter, Sepp},
  journal={arXiv preprint arXiv:2505.23719},
  year={2025}
}

@inproceedings{lipman2023flow,
  title     = {Flow Matching for Generative Modeling},
  author    = {Lipman, Yaron and Chen, Ricky TQ and Ben-Hamu, Heli and Nickel, Maximilian and Le, Matt},
  booktitle = {The Eleventh International Conference on Learning Representations (ICLR)},
  year      = {2023},
}

@book{perko2013differential,
  title={Differential equations and dynamical systems},
  author={Perko, Lawrence},
  volume={7},
  year={2001},
  publisher={Springer Science \& Business Media}
}

@inproceedings{rangapuram2018deepstate,
  title        = {Deep State Space Models for Time Series Forecasting},
  author       = {Syama Sundar Rangapuram and
                  Matthias W. Seeger and
                  Jan Gasthaus and
                  Lorenzo Stella and
                  Yuyang Wang and
                  Tim Januschowski},
  booktitle    = {Advances in Neural Information Processing Systems 31},
  year         = {2018},
}

@inproceedings{chen2018neuralode,
  title        = {Neural Ordinary Differential Equations},
  author       = {Tian Qi Chen and
                  Yulia Rubanova and
                  Jesse Bettencourt and
                  David Duvenaud},
  booktitle    = {Advances in Neural Information Processing Systems 31},
  year={2018}
}

@article{salinas2020deepar,
  title={{DeepAR}: Probabilistic forecasting with autoregressive recurrent networks},
  author={Salinas, David and Flunkert, Valentin and Gasthaus, Jan and Januschowski, Tim},
  journal={International journal of forecasting},
  volume={36},
  number={3},
  pages={1181--1191},
  year={2020},
  publisher={Elsevier}
}

@article{hochreiter1997lstm,
  title={Long Short-Term Memory},
  author={Hochreiter, Sepp and Schmidhuber, J{\"u}rgen},
  journal={Neural Computation},
  volume={9},
  number={8},
  pages={1735--1780},
  year={1997}
}

@article{elman1990rnn,
  title={Finding Structure in Time},
  author={Elman, Jeffrey L},
  journal={Cognitive Science},
  volume={14},
  number={2},
  pages={179--211},
  year={1990}
}

@article{sims1980macro,
  title={Macroeconomics and Reality},
  author={Sims, Christopher A.},
  journal={Econometrica},
  volume={48},
  number={1},
  pages={1--48},
  year={1980}
}

@book{box2015time,
  title={Time Series Analysis: Forecasting and Control},
  author={Box, George E. P. and Jenkins, Gwilym M. and Reinsel, Gregory C. and Ljung, Greta M.},
  edition={5},
  publisher={Wiley},
  year={2015}
}

@inproceedings{wang2024timemixer,
  title        = {{TimeMixer}: Decomposable Multiscale Mixing for Time Series Forecasting},
  author       = {Shiyu Wang and
                  Haixu Wu and
                  Xiaoming Shi and
                  Tengge Hu and
                  Huakun Luo and
                  Lintao Ma and
                  James Y. Zhang and
                  Jun Zhou},
  booktitle    = {The Twelfth International Conference on Learning Representations, {ICLR}},
  year={2024}
}

@article{kraus2025xlstmmixer,
  title={{xLSTM}-mixer: Multivariate time series forecasting by mixing via scalar memories},
  author={Kraus, Maurice and Divo, Felix and Dhami, Devendra Singh and Kersting, Kristian},
  journal={arXiv preprint arXiv:2410.16928},
  year={2024}
}

@inproceedings{kong2025plstm,
  title        = {Unlocking the Power of {LSTM} for Long Term Time Series Forecasting},
  author       = {Yaxuan Kong and
                  Zepu Wang and
                  Yuqi Nie and
                  Tian Zhou and
                  Stefan Zohren and
                  Yuxuan Liang and
                  Peng Sun and
                  Qingsong Wen},
  booktitle    = {Thirty-Ninth {AAAI} Conference on Artificial Intelligence},
  year={2025}
}

@article{alharthi2024xlstmtime,
  title={{xLSTMTime}: Long-term time series forecasting with {xLSTM}},
  author={Alharthi, Musleh and Mahmood, Ausif},
  journal={Ai},
  volume={5},
  number={3},
  pages={1482--1495},
  year={2024},
  publisher={MDPI}
}

@article{gneiting2014probabilistic,
  title={Probabilistic forecasting},
  author={Gneiting, Tilmann and Katzfuss, Matthias},
  journal={Annual Review of Statistics and Its Application},
  volume={1},
  number={1},
  pages={125--151},
  year={2014},
  publisher={Annual Reviews}
}

@article{hong2016probabilistic,
  title={Probabilistic electric load forecasting: A tutorial review},
  author={Hong, Tao and Fan, Shu},
  journal={International Journal of Forecasting},
  volume={32},
  number={3},
  pages={914--938},
  year={2016},
  publisher={Elsevier}
}

@inproceedings{zhang2025skolr,
  author       = {Yitian Zhang and
                  Liheng Ma and
                  Antonios Valkanas and
                  Boris N. Oreshkin and
                  Mark Coates},
  title        = {{SKOLR:} Structured {Koopman} Operator Linear {RNN} for Time-Series Forecasting},
  booktitle    = {Forty-second International Conference on Machine Learning, {ICML}},
  series       = {Proceedings of Machine Learning Research},
  year         = {2025},
}

@inproceedings{wang2023koopman,
  author       = {Rui Wang and Yihe Dong and Sercan {\"{O}}. Arik and Rose Yu},
  title        = {Koopman Neural Operator Forecaster for Time-series with Temporal Distributional Shifts},
  booktitle    = {The Eleventh International Conference on Learning Representations, {ICLR}},
  year         = {2023},
}

@inproceedings{li2022d3vae,
  author       = {Yan Li and Xinjiang Lu and Yaqing Wang and Dejing Dou},
  title        = {Generative Time Series Forecasting with Diffusion, Denoise, and Disentanglement},
  booktitle    = {Advances in Neural Information Processing Systems (NeurIPS)},
  year         = {2022},
}

@inproceedings{sohl2015deep,
  title={Deep unsupervised learning using nonequilibrium thermodynamics},
  author={Sohl-Dickstein, Jascha and Weiss, Eric and Maheswaranathan, Niru and Ganguli, Surya},
  booktitle={International conference on machine learning},
  pages={2256--2265},
  year={2015},
  organization={pmlr}
}

@article{ho2020denoising,
  title={Denoising diffusion probabilistic models},
  author={Ho, Jonathan and Jain, Ajay and Abbeel, Pieter},
  journal={Advances in neural information processing systems},
  volume={33},
  pages={6840--6851},
  year={2020}
}

@article{karras2022elucidating,
  title={Elucidating the design space of diffusion-based generative models},
  author={Karras, Tero and Aittala, Miika and Aila, Timo and Laine, Samuli},
  journal={Advances in neural information processing systems},
  volume={35},
  pages={26565--26577},
  year={2022}
}

@article{ansari2025chronos,
  title={Chronos-2: From univariate to universal forecasting},
  author={Ansari, Abdul Fatir and Shchur, Oleksandr and K{\"u}ken, Jaris and Auer, Andreas and Han, Boran and Mercado, Pedro and Rangapuram, Syama Sundar and Shen, Huibin and Stella, Lorenzo and Zhang, Xiyuan and others},
  journal={arXiv preprint arXiv:2510.15821},
  year={2025}
}

@article{matheson1976scoring,
  title={Scoring rules for continuous probability distributions},
  author={Matheson, James E. and Winkler, Robert L.},
  journal={Management Science},
  volume={22},
  number={10},
  pages={1087--1096},
  year={1976}
}

@book{engel2000one,
  title={One-parameter semigroups for linear evolution equations},
  author={Engel, Klaus-Jochen and Nagel, Rainer},
  year={2000},
  publisher={Springer}
}

@article{gneiting2007strictly,
  title={Strictly proper scoring rules, prediction, and estimation},
  author={Gneiting, Tilmann and Raftery, Adrian E},
  journal={Journal of the American statistical Association},
  volume={102},
  number={477},
  pages={359--378},
  year={2007},
  publisher={Taylor \& Francis}
}

@article{chen2024generative,
  title={Generative machine learning methods for multivariate ensemble postprocessing},
  author={Chen, Jieyu and Janke, Tim and Steinke, Florian and Lerch, Sebastian},
  journal={The Annals of Applied Statistics},
  volume={18},
  number={1},
  pages={159--183},
  year={2024},
  publisher={Institute of Mathematical Statistics}
}

@article{nochumsohn2025multi,
  title={A multi-task learning approach to linear multivariate forecasting},
  author={Nochumsohn, Liran and Zisling, Hedi and Azencot, Omri},
  journal={arXiv preprint arXiv:2502.03571},
  year={2025}
}

@article{nochumsohn2025super,
  title={Super-Linear: A Lightweight Pretrained Mixture of Linear Experts for Time Series Forecasting},
  author={Nochumsohn, Liran and Marshanski, Raz and Zisling, Hedi and Azencot, Omri},
  journal={arXiv preprint arXiv:2509.15105},
  year={2025}
}

@article{liu2025freqmoe,
  title={{FreqMoE}: Enhancing time series forecasting through frequency decomposition mixture of experts},
  author={Liu, Ziqi},
  journal={arXiv preprint arXiv:2501.15125},
  year={2025}
}

@inproceedings{naiman2024generative,
  title={Generative modeling of regular and irregular time series data via {K}oopman {VAEs}},
  author={Naiman, Ilan and Erichson, N Benjamin and Ren, Pu and Mahoney, Michael W and Azencot, Omri},
  booktitle={International conference on learning representations},
  volume={2024},
  pages={50547--50573},
  year={2024}
}

@article{naiman2024utilizing,
  title={Utilizing image transforms and diffusion models for generative modeling of short and long time series},
  author={Naiman, Ilan and Berman, Nimrod and Pemper, Itai and Arbiv, Idan and Fadlon, Gal and Azencot, Omri},
  journal={Advances in Neural Information Processing Systems},
  volume={37},
  pages={121699--121730},
  year={2024}
}

@article{gonen2026time,
  title={Time series generation under data scarcity: A unified generative modeling approach},
  author={Gonen, Tal and Pemper, Itai and Naiman, Ilan and Berman, Nimrod and Azencot, Omri},
  journal={Advances in Neural Information Processing Systems},
  volume={38},
  pages={172751--172790},
  year={2026}
}

@article{berman2026one,
  title={One-step offline distillation of diffusion-based models via {K}oopman modeling},
  author={Berman, Nimrod and Naiman, Ilan and Eliasof, Moshe and Zisling, Hedi and Azencot, Omri},
  journal={Advances in Neural Information Processing Systems},
  volume={38},
  pages={14776--14811},
  year={2026}
}

@inproceedings{azencot2013operator,
  title={An operator approach to tangent vector field processing},
  author={Azencot, Omri and Ben-Chen, Mirela and Chazal, Fr{\'e}d{\'e}ric and Ovsjanikov, Maks},
  booktitle={Computer Graphics Forum},
  volume={32},  
  pages={73--82},
  year={2013},
  organization={Wiley Online Library}
}

@inproceedings{azencot2020forecasting,
  title={Forecasting sequential data using consistent {K}oopman autoencoders},
  author={Azencot, Omri and Erichson, N Benjamin and Lin, Vanessa and Mahoney, Michael},
  booktitle={International conference on machine learning},
  pages={475--485},
  year={2020},
  organization={PMLR}
}

@inproceedings{naiman2023operator,
  title={An operator theoretic approach for analyzing sequence neural networks},
  author={Naiman, Ilan and Azencot, Omri},
  booktitle={Proceedings of the AAAI conference on artificial intelligence},
  volume={37},
  pages={9268--9276},
  year={2023}
}
\bibliographystyle{abbrv}

\newpage
\addtocontents{toc}{\protect\setcounter{tocdepth}{2}}
\appendix
\setcounter{tocdepth}{2}

{\noindent\Huge\bfseries Appendix}\par
\vspace{1.5em}
\noindent\rule{\linewidth}{0.6pt}\par
\vspace{0.4em}
{\noindent\Large\bfseries Table of Contents}\par
\vspace{0.2em}
\noindent\rule{\linewidth}{0.4pt}\par
\vspace{0.8em}

\begingroup
\hypersetup{linkcolor=black}
\etocsettocstyle{}{}
\tableofcontents
\endgroup
\clearpage

\section{Architecture}
\label{app:arch-details}
The main paper focuses on the conceptual design and motivation underlying KGO. In this appendix, we provide additional implementation details for each module, following the structure of Sec.~\ref{sec:method} and Fig.~\ref{fig:architecture}: representation learning, Koopman Patch Embedding (KoPE), Koopman Flow Matching (KoFM), the deterministic decoder (ResMLP), and the Adaptive Uncertainty Gate (AUG).

\subsection{Representation Learning}
\label{app:arch-representation}
Given an input sequence $\mathbf{x}_{1:L} \in \mathbb{R}^L$, we first apply RevIN normalization independently per variable. The normalized sequence is partitioned into non-overlapping patches of length $P$, producing $N=\lfloor L/P \rfloor$ patches. Each patch is flattened and projected through a learned linear embedding layer into a $D$-dimensional latent representation.

The sequence encoder is implemented using stacked sLSTM blocks~\citep{beck2024xlstm}. Each block combines recurrent state updates with multiplicative gating and normalization, allowing efficient long-context processing while maintaining stable hidden-state dynamics. Compared to Transformer-based alternatives, the recurrent formulation reduces memory complexity from quadratic to linear in sequence length and empirically improves forecasting performance in our setting (Sec.~\ref{sec:ablation}). The encoder outputs a latent sequence $\mathbf{Z} = [\mathbf{z}_1,\dots,\mathbf{z}_N]$, from which the final latent state $\mathbf{z}_N$ initializes the KoPE rollout.

\subsection{Koopman Patch Embedding (KoPE)}
\label{app:arch-kope}
The approximate Koopman operator $K \in \mathbb{R}^{D \times D}$ is initialized orthogonally and normalized by its spectral norm whenever this norm exceeds $1$, so that $\lVert K\rVert_2 \le 1$. LayerNorm is applied before the operator rollout and to each resulting conditioning state. Future latent states are generated recursively:
\[
\tilde{\mathbf{z}}_{N+i} = K \tilde{\mathbf{z}}_{N+i-1} \ , \quad i = 2, \dots, M \ ,
\qquad
\tilde{\mathbf{z}}_{N+1} = \mathbf{z}_N \ .
\]
In practice, this recursive rollout is fully differentiable and efficiently computed in parallel over the horizon dimension. The resulting latent trajectory $\tilde{\mathbf{Z}}_{\text{KoPE}} = \left[\tilde{\mathbf{z}}_{N+1}, \tilde{\mathbf{z}}_{N+2}, \ldots, \tilde{\mathbf{z}}_{N+M} \right]$, $M = \lceil H/P \rceil$, serves as a conditioning signal shared by both deterministic and probabilistic decoding streams.

\subsection{Koopman Flow Matching (KoFM)}
\label{app:arch-kofm}
\paragraph{Lift.}
Within KoFM, the interpolated inputs $\mathbf{x}^{(\tau)}_{N+i}$ given by
\[
\mathbf{x}^{(\tau)}_{N+i} = (1-\tau)\boldsymbol{\epsilon} + \tau \mathbf{x}_{N+i} \ , \qquad \tau \in [0,1], \quad \boldsymbol{\epsilon} \sim \mathcal{N}(\mathbf{0},\mathbf{I}) \ ,
\]
are first mapped into the latent space through a learned linear projection. Conditioning on the KoPE trajectory is introduced through adaptive Layer Normalization (AdaLN): each KoPE latent $\tilde{\mathbf{z}}_{N+i}$ determines scale, shift, and residual modulation parameters. The normalized and modulated activations pass through an MLP, whose output is scaled and added to the projected input. The resulting conditioned latent $\tilde{\mathbf{z}}^{(\tau)}_{N+i}$ is then transported toward $\tau=1$.

\paragraph{Transport.}
The conditioned latent representation $\tilde{\mathbf{z}}^{(\tau)}_{N+i}$ is transported by a learned matrix $\mathbf{A} \in \mathbb{R}^{D \times D}$ and offset $\mathbf{b} \in \mathbb{R}^D$ through the closed-form affine map
\[
\tilde{\mathbf{z}}^{(1)}_{N+i} = e^{(1-\tau)\mathbf{A}} \tilde{\mathbf{z}}^{(\tau)}_{N+i} + (1-\tau)\mathbf{b} \ ,
\]
where the matrix exponential is computed using the differentiable operator provided by PyTorch. The exponential represents the exact homogeneous linear evolution, while the additive offset retains the first-order term of the corresponding affine ODE's offset integral (Appendix~\ref{app:exact_koopman_integration}). During training, $\tau\sim\mathcal{U}[0,1]$ is masked to $0$ for each sample with probability $0.5$ (Sec.~\ref{subsec:kofm}), and the remaining transport time $1-\tau$ is rounded to the nearest of 32 uniformly spaced values in $[0,1]$. At inference, $\tau=0$, and the cached matrix exponential and learned bias enable single-step latent transport. Multiple trajectories are generated by sampling independent Gaussian noise vectors $\boldsymbol{\epsilon}$ in parallel.

\paragraph{Reconstruction (Recon).}
A reconstruction head decodes the transported latents $\tilde{\mathbf{z}}^{(1)}_{N+1:N+M}$ into forecast patches. Each transported latent $\tilde{\mathbf{z}}^{(1)}_{N+i}$ is layer-normalized, modulated by AdaLN scale and shift parameters computed from the same KoPE latent $\tilde{\mathbf{z}}_{N+i}$ that conditions the Lift, and linearly projected to a patch of length $P$. The decoded patches are concatenated to produce $\hat{\mathbf{x}}^{\mathrm{prob}}_{L+1:L+H}$.

\subsection{Deterministic Decoder (ResMLP)}
\label{app:arch-resmlp}
In parallel to KoFM, KGO includes a deterministic decoder designed to capture the predictable component of the signal. Each KoPE latent $\tilde{\mathbf{z}}_{N+i}$ is processed by a residual MLP consisting of linear layers with GELU activations, Layer Normalization, and dropout regularization. A linear skip connection projects the latent directly into the output space:
\[
\hat{\mathbf{x}}^{\mathrm{det}}_{N+i}
=
\mathrm{MLP}(\tilde{\mathbf{z}}_{N+i})
+
W_{\mathrm{res}}\tilde{\mathbf{z}}_{N+i}.
\]
Predicted patches are concatenated to form the deterministic forecast $\hat{\mathbf{x}}^{\mathrm{det}}_{L+1:L+H}$.

\subsection{Adaptive Uncertainty Gate (AUG)}
\label{app:arch-aug}
AUG applies a lightweight encoder--router network independently to each variable's historical context. The router produces two logits per forecast step, corresponding to the stochastic and deterministic branches. During training, differentiable Gumbel-Softmax with a learned temperature produces the routing weights; at inference, ordinary softmax produces deterministic weights conditional on the context. The stochastic weight is denoted by $g^h$, with complementary deterministic weight $1-g^h$, following Eq.~\eqref{eq:aug_prediction}. This fine-grained routing mechanism enables KGO to interpolate adaptively between deterministic and probabilistic predictions depending on local predictability. The learned gates exhibit horizon- and variate-dependent weighting, with increasing stochastic contributions at later forecast steps on several datasets (Appendix~\ref{app:uncertainty_gate}), consistent with AUG's motivation to accommodate heterogeneous aleatoric uncertainty.

\clearpage

\section{Theory}
\subsection{Theoretical Analysis of the Uncertainty Gate}
\label{subsec:theory-gate}

Let $y$ denote a future value and $\mathbf{x}_{1:L}$ the historical context. Following Catt~\citep{catt2026forecastability}, the maximum achievable reduction in expected log loss relative to the marginal baseline equals the mutual information between the future and the context:
\begin{equation}
    \max_{\theta}\Big(H(y) + \mathbb{E}_{p(\mathbf{x}_{1:L},y)}\big[\log p_\theta(y\mid\mathbf{x}_{1:L})\big]\Big) = I(y;\mathbf{x}_{1:L}) .
\end{equation}
This forecastability differs across variables and forecast steps, and the uncertainty that remains given the context is aleatoric. AUG is motivated by this heterogeneity: the gate is trained end-to-end, so the gradient of the CRPS objective determines, per variable and forecast step, how much the forecast relies on the stochastic branch. Moreover, by the data processing inequality, any compression $f$ of the context cannot increase the available forecastability, $I(y;f(\mathbf{x}_{1:L}))\le I(y;\mathbf{x}_{1:L})$; AUG therefore conditions the gate on the raw context rather than on an encoded representation.

AUG adapts the contribution of stochastic predictions to account for differences in aleatoric uncertainty across variables and forecast horizons. We characterize its effect using the prediction rule in Eq.~\eqref{eq:aug_prediction}. Let $\mathbf{x}=\mathbf{x}_{1:L}$ denote the historical context. We analyze the predictive distribution at inference. Conditional on $\mathbf{x}$, the deterministic prediction $\hat{x}^{\mathrm{det}}_{L+h}$ and gate $g^h\in[0,1]$ are fixed, while $\hat{x}^{\mathrm{prob}}_{L+h}$ varies across generated samples.

\begin{proposition}[Gate-Controlled Predictive Dispersion]
\label{prop:gate-routing}
Assuming the stochastic prediction has finite conditional variance, the gated forecast
\[
    \hat{x}_{L+h}
    = (1-g^h)\hat{x}^{\mathrm{det}}_{L+h}
    + g^h\hat{x}^{\mathrm{prob}}_{L+h}
\]
satisfies
\begin{align}
    \mathbb{E}[\hat{x}_{L+h}\mid\mathbf{x}]
    &= (1-g^h)\hat{x}^{\mathrm{det}}_{L+h}
    + g^h\mathbb{E}[\hat{x}^{\mathrm{prob}}_{L+h}\mid\mathbf{x}],
    \\[0.3em]
    \operatorname{Var}(\hat{x}_{L+h}\mid\mathbf{x})
    &= (g^h)^2\operatorname{Var}(\hat{x}^{\mathrm{prob}}_{L+h}\mid\mathbf{x}).
\end{align}
\end{proposition}

\begin{proof}
Conditional on $\mathbf{x}$, the deterministic prediction and gate are constants. The identities follow from linearity of expectation and the variance of an affine transformation.
\end{proof}

Thus, AUG jointly controls the location and dispersion of the predictive distribution. For a fixed stochastic branch, smaller gate values contract generated samples toward the deterministic prediction, while larger values retain more of the stochastic branch's variability. KGO learns this balance through its CRPS-based training objective, allowing the predictive spread to adapt across variables and horizons without a manually specified uncertainty schedule.

This provides a direct interpretation of AUG as a mechanism for modeling aleatoric uncertainty. The gate itself is not an identifiable estimate of that uncertainty, since predictive dispersion also depends on the learned stochastic branch. Appendix~\ref{app:uncertainty_gate} presents empirical analyses of the resulting gate behavior.

\subsection{Matrix-Exponential Transport and Affine Correction}
\label{app:exact_koopman_integration}
In standard Flow Matching \citep{lipman2023flow} and Continuous Normalizing Flows (CNFs) \citep{chen2018neuralode}, generating a sample requires integrating a learned, time-dependent neural vector field $v_\theta(z, \tau)$ over the continuous flow time $\tau \in [0,1]$. Standard implementations rely on numerical ODE solvers (e.g., Runge--Kutta), which introduce truncation errors and require multiple sequential function evaluations.

In Koopman Flow Matching (KoFM), we project the data into a latent space. By Koopman semigroup theory \citep{engel2000one}, lifted observables of any nonlinear dynamical system evolve linearly under the Koopman generator $\mathcal{L}$ (i.e., $\dot{\psi} = \mathcal{L}\psi$). We approximate this infinite-dimensional operator with a finite-dimensional, linear, time-invariant (LTI) generator matrix $A$. Below, we establish the exact flow map for the homogeneous latent dynamics and its density transformation, assuming the latent distribution admits a density. We then describe the affine correction and time discretization used in the implementation.

\subsubsection{The Linear Vector Field and the Exact Flow Map}
Let $z_\tau \in \mathbb{R}^d$ be the latent state at flow-matching time $\tau \in [0,1]$. We parameterize the vector field as a linear autonomous ordinary differential equation:
\begin{equation}
    \frac{dz}{d\tau} = v(z) = A z
\end{equation}
where $A \in \mathbb{R}^{d \times d}$ is the learned generator matrix.

\begin{theorem}[Exact Flow Map]
\label{thm:exact_flow}
The unique solution to the initial value problem $\frac{dz}{d\tau} = Az$ is given by the matrix exponential flow map $\Phi_s: \mathbb{R}^d \to \mathbb{R}^d$:
\begin{equation}
    \Phi_s(z) = e^{As} z
\end{equation}
where $e^{As} = \sum_{k=0}^\infty \frac{(As)^k}{k!}$.
\end{theorem}

\begin{proof}
Using the Taylor series expansion of the matrix exponential, we differentiate term-by-term (which is valid due to uniform convergence):
\begin{equation}
    \frac{d}{ds} e^{As} = \sum_{k=1}^\infty \frac{A^k s^{k-1}}{(k-1)!} = A \sum_{k=0}^\infty \frac{A^k s^k}{k!} = A e^{As}
\end{equation}
Applying this to a state $z$, we obtain $\frac{d}{ds} (e^{As}z) = A(e^{As}z)$, verifying that the mapping satisfies the differential equation. Furthermore, because the vector field $v(z) = Az$ is globally Lipschitz continuous, the Picard--Lindel\"of theorem guarantees that this is the unique solution of the initial value problem \citep{perko2013differential}.
\end{proof}

For the homogeneous dynamics, a latent state $z_\tau$ at transport time $\tau$ is advanced to transport time $1$. The remaining integration time is $\Delta \tau = 1 - \tau$, giving:
\begin{equation}
    z_1 = \Phi_{1-\tau}(z_\tau) = e^{A(1-\tau)} z_\tau
\end{equation}
This mapping exactly parameterizes the homogeneous latent trajectory and motivates the matrix-exponential component of KoFM.

\subsubsection{Exact Log-Density Evaluation}
To demonstrate how tracking the single trajectory $\Phi_{1-\tau}$ correctly evolves the probability density, we analyze the density pointwise along the continuous flow.

\begin{lemma}
\label{lem:divergence}
For a linear vector field $v(z) = Az$, the divergence is the trace of the generator matrix: $\nabla \cdot v(z) = \text{Tr}(A)$.
\end{lemma}

\begin{theorem}[Log-Density Evaluation]
Under the linear vector field $v(z) = Az$, the log-density evaluates pointwise along the trajectory as:
\begin{equation}
    \log p_1(z_1) = \log p_\tau(\Phi_{-(1-\tau)}(z_1)) - \text{Tr}(A)(1-\tau)
\end{equation}
\end{theorem}

\begin{proof}
From the Continuous Normalizing Flow framework \citep{chen2018neuralode}, the instantaneous change of variables formula tracks the log-density $p_\tau(z(\tau))$ along a deterministic trajectory:
\begin{equation}
    \frac{d}{d\tau} \log p_\tau(z(\tau)) = -\nabla \cdot v(z(\tau))
\end{equation}
Substituting Lemma \ref{lem:divergence}, the divergence is the constant scalar $\text{Tr}(A)$. Integrating this from current time $\tau$ to target time $1$ yields:
\begin{equation}
    \log p_1(z(1)) - \log p_\tau(z(\tau)) = -\int_\tau^1 \text{Tr}(A) \, ds = -\text{Tr}(A)(1-\tau)
\end{equation}
Let $z_1 = z(1)$ and $z_\tau = z(\tau)$. Because the flow is invertible, the pre-image of the state is $z_\tau = \Phi_{-(1-\tau)}(z_1) = e^{-A(1-\tau)}z_1$. Substituting this into the integrated equation yields the final log-density identity.
\end{proof}
This identity describes the density transformation under the homogeneous latent flow. It does not establish an exact likelihood for the complete forecasting model, which also includes learned lifting and decoding maps.

\subsubsection{Efficient Computation and Affine Extension}
Evaluating \texttt{torch.matrix\_exp} directly on a batch of samples with unique, uniformly distributed flow times $\tau_i \sim \mathcal{U}(0,1)$ requires computing $B$ distinct matrix exponentials, which scales poorly during training. Furthermore, dynamically computing the matrix exponential at inference time is suboptimal.

\textbf{Training via Time-Binning:} During training, we discretize the time domain $[0,1]$ into $n_b$ uniformly spaced bins. Rather than computing $B$ distinct matrix exponentials, we compute exactly $n_b$ exponentials:
\begin{equation}
    \mathcal{E} = \left\{ e^{A \cdot \Delta \tau_k} \right\}_{k=1}^{n_b}
\end{equation}
We then assign each sample's integration time $\Delta \tau_i = (1-\tau_i)$ to its nearest bin. This reduces the forward pass complexity to $n_b$ exponentiations shared across the batch, heavily accelerating training while maintaining the continuous-time parameterization.

\textbf{Cached Single-Step Inference:} At inference, standard Gaussian noise is first lifted into the latent space at $\tau=0$. The remaining transport time is fixed at $\Delta \tau = 1$, so the implemented affine transport becomes:
\begin{equation}
    z_1 = e^{A} z_0 + b
\end{equation}
Because $A$ is static after training, we precompute and cache the single operator matrix $C = e^A$. Latent transport then reduces to a single $\mathcal{O}(d^2)$ matrix multiplication and bias addition per sample.

\textbf{Affine Implementation Note:} For the affine dynamics $\frac{d\mathbf{z}}{d\tau}=\mathbf{A}\mathbf{z}+\mathbf{b}$, the exact solution over the remaining transport time $\Delta\tau=1-\tau$ is
\[
    \mathbf{z}_1 = e^{\Delta\tau\mathbf{A}}\mathbf{z}_\tau
    + \int_0^{\Delta\tau} e^{s\mathbf{A}}\mathbf{b}\,ds.
\]
Equation~\eqref{eqn:latent_dynamics} uses $\Delta\tau\mathbf{b}$ in place of this integral, retaining its first-order term. Thus, the matrix exponential exactly represents the homogeneous linear evolution, while the additive offset generally approximates the corresponding affine ODE solution.

\clearpage

\section{Experiment Details}
\subsection{Training Details}
\label{app:training-details}

KGO is implemented in PyTorch and trained with PyTorch Lightning under the ProbTS pipeline~\citep{zhang2024probts}, using a single GPU. Larger datasets (Electricity, Traffic) are trained on an NVIDIA RTX 6000 Ada, while the remaining smaller datasets are trained on an RTX 4090. We optimize with AdamW ($\beta = (0.9, 0.95)$, weight decay $0.1$) under a cosine schedule with $10\%$ linear warmup, placing the Koopman generator parameters in a separate group at $0.1\times$ the base learning rate.

\subsection{Datasets}
\label{app:datasets}

Table~\ref{tab:datasets} summarizes the public benchmarks used throughout the paper. Long-term datasets (\textit{-L}) and short-term datasets (\textit{-S}) follow the ProbTS protocol of~\citep{zhang2024probts}; splits, normalization, and context/horizon settings are inherited without modification.

\begin{table}[htbp]
\centering
\caption{Statistical information of the datasets.}
\label{tab:datasets}
\footnotesize
\begin{adjustbox}{max width=\linewidth}
\begin{tabular}{@{}ll c c c r l@{}}
\toprule
\textbf{Horizon} & \textbf{Dataset} & \textbf{\#var.} & \textbf{range} & \textbf{freq.} & \textbf{timesteps} & \textbf{Description} \\
\midrule
\multirow{7}{*}{\textbf{Long-term}}
 & ETTh1/h2-L    & 7   & $\mathbb{R}^{+}$ & H         & 17{,}420 & Electricity transformer temperature per hour \\
 & ETTm1/m2-L    & 7   & $\mathbb{R}^{+}$ & 15min     & 69{,}680 & Electricity transformer temperature every 15 min \\
 & Electricity-L & 321 & $\mathbb{R}^{+}$ & H         & 26{,}304 & Electricity consumption (kWh) \\
 & Traffic-L     & 862 & $(0,1)$          & H         & 17{,}544 & Road occupancy rates \\
 & Exchange-L    & 8   & $\mathbb{R}^{+}$ & Busi.\ Day & 7{,}588 & Daily exchange rates of 8 countries \\
 & ILI-L         & 7   & $(0,1)$          & W         & 966      & Ratio of patients seen with influenza-like illness \\
 & Weather-L     & 21  & $\mathbb{R}^{+}$ & 10min     & 52{,}696 & Local climatological data \\
\midrule
\multirow{6}{*}{\textbf{Short-term}}
 & ETTh1/h2-S    & 7   & $\mathbb{R}^{+}$ & H         & 17{,}420 & Electricity transformer temperature per hour \\
 & ETTm1/m2-S    & 7   & $\mathbb{R}^{+}$ & 15min     & 69{,}680 & Electricity transformer temperature every 15 min \\
 & Exchange-S    & 8   & $\mathbb{R}^{+}$ & Busi.\ Day & 6{,}071 & Daily exchange rates of 8 countries \\
 & Solar-S       & 137 & $\mathbb{R}^{+}$ & H         & 7{,}009  & Solar power production records \\
 & Electricity-S & 370 & $\mathbb{R}^{+}$ & H         & 5{,}833  & Electricity consumption \\
 & Traffic-S     & 963 & $(0,1)$          & H         & 4{,}001  & Road occupancy rates \\
\bottomrule
\end{tabular}
\end{adjustbox}
\end{table}
\FloatBarrier

\subsection{Metrics}
\label{app:metrics}

\paragraph{Normalized Mean Absolute Error (NMAE).}
The Normalized Mean Absolute Error is a normalized version of the MAE, dimensionless and therefore comparable across datasets of different magnitude. It is defined as
\begin{equation}
\label{eq:nmae}
\mathrm{NMAE} \;=\; \frac{\sum_{k=1}^{K} \sum_{t=1}^{T} \bigl| x_t^{k} - \hat{x}_t^{k} \bigr|}{\sum_{k=1}^{K} \sum_{t=1}^{T} \bigl| x_t^{k} \bigr|}.
\end{equation}

\paragraph{Continuous Ranked Probability Score (CRPS).}
The CRPS~\citep{matheson1976scoring} quantifies the agreement between a cumulative distribution function (CDF) $F$ and an observation $x$:
\begin{equation}
\label{eq:crps}
\mathrm{CRPS} \;=\; \int_{\mathbb{R}} \bigl(F(z) - \mathbb{I}\{x \le z\}\bigr)^{2}\, dz,
\end{equation}
where $\mathbb{I}\{x \le z\}$ denotes the indicator function, equal to one if $x \le z$ and zero otherwise. As a strictly proper scoring rule, CRPS is minimized in expectation when the predictive distribution matches the true conditional distribution. When $F$ is replaced by the empirical CDF $\hat{F}(z) = \tfrac{1}{n}\sum_{i=1}^{n} \mathbb{I}\{X_i \le z\}$ with samples $X_i \sim F$, the CRPS can be computed directly from samples of the conditional distribution $p_\theta(\mathbf{x}_t \mid \mathbf{h}_t)$. We use $n = 100$ samples to estimate the empirical CDF.

\clearpage

\section{Extended Results}
\label{app:extended_results}
\subsection{Full Results Tables}
\label{app:results}

\begin{table}[!htbp]
\centering
\begingroup
\setlength{\tabcolsep}{2pt}
\renewcommand{\arraystretch}{1.1}
\caption{Full long-term probabilistic forecasting results across all baselines and datasets; protocol as in Table~\ref{tab:ltsf}. Lower CRPS values indicate better predictions. Each cell reports the mean $\pm$ standard deviation of the CRPS across 5 independent runs of retraining and evaluation. \best{Red bold}: best; \second{blue underline}: second best. Due to the excessive time and memory consumption, some results are unavailable and denoted as ``--''.}
\label{tab:full:ltsf:crps}

\begin{adjustbox}{max width=\linewidth, max totalheight=0.92\textheight}
\begin{tabular}{ll cccccc cc c}
\toprule
Dataset & Horizon &
TSDiff & CSDI & TimeGrad &
GRU MAF & Trans MAF & GRU NVP &
$K^2$VAE &
\textbf{KGO (Ours)} \\
\midrule

\multirow{4}{*}{\rotatebox{90}{ETTm1-L}}
& 96  & 0.344{\footnotesize $\pm$0.050} & 0.236{\footnotesize $\pm$0.006} & 0.522{\footnotesize $\pm$0.105} & 0.295{\footnotesize $\pm$0.055} & 0.313{\footnotesize $\pm$0.045} & 0.383{\footnotesize $\pm$0.053} & \second{0.232{\footnotesize $\pm$0.010}} & \best{0.215{\footnotesize $\pm$0.001}} \\
& 192 & 0.345{\footnotesize $\pm$0.035} & 0.291{\footnotesize $\pm$0.025} & 0.603{\footnotesize $\pm$0.092} & 0.389{\footnotesize $\pm$0.033} & 0.424{\footnotesize $\pm$0.029} & 0.396{\footnotesize $\pm$0.030} & \second{0.259{\footnotesize $\pm$0.013}} & \best{0.236{\footnotesize $\pm$0.001}} \\
& 336 & 0.462{\footnotesize $\pm$0.043} & 0.322{\footnotesize $\pm$0.033} & 0.601{\footnotesize $\pm$0.028} & 0.429{\footnotesize $\pm$0.021} & 0.481{\footnotesize $\pm$0.019} & 0.486{\footnotesize $\pm$0.032} & \second{0.262{\footnotesize $\pm$0.030}} & \best{0.249{\footnotesize $\pm$0.000}} \\
& 720 & 0.478{\footnotesize $\pm$0.027} & 0.448{\footnotesize $\pm$0.038} & 0.621{\footnotesize $\pm$0.037} & 0.536{\footnotesize $\pm$0.033} & 0.688{\footnotesize $\pm$0.043} & 0.546{\footnotesize $\pm$0.036} & \second{0.294{\footnotesize $\pm$0.026}} & \best{0.276{\footnotesize $\pm$0.001}} \\
\midrule

\multirow{4}{*}{\rotatebox{90}{ETTm2-L}}
& 96  & 0.175{\footnotesize $\pm$0.019} & \second{0.115{\footnotesize $\pm$0.009}} & 0.427{\footnotesize $\pm$0.042} & 0.177{\footnotesize $\pm$0.024} & 0.227{\footnotesize $\pm$0.013} & 0.319{\footnotesize $\pm$0.044} & 0.126{\footnotesize $\pm$0.007} & \best{0.106{\footnotesize $\pm$0.001}} \\
& 192 & 0.255{\footnotesize $\pm$0.029} & \second{0.147{\footnotesize $\pm$0.008}} & 0.424{\footnotesize $\pm$0.061} & 0.411{\footnotesize $\pm$0.026} & 0.253{\footnotesize $\pm$0.037} & 0.326{\footnotesize $\pm$0.025} & 0.148{\footnotesize $\pm$0.009} & \best{0.130{\footnotesize $\pm$0.001}} \\
& 336 & 0.328{\footnotesize $\pm$0.047} & 0.190{\footnotesize $\pm$0.018} & 0.469{\footnotesize $\pm$0.049} & 0.377{\footnotesize $\pm$0.023} & 0.253{\footnotesize $\pm$0.013} & 0.449{\footnotesize $\pm$0.145} & \second{0.164{\footnotesize $\pm$0.010}} & \best{0.147{\footnotesize $\pm$0.004}} \\
& 720 & 0.344{\footnotesize $\pm$0.046} & 0.239{\footnotesize $\pm$0.035} & 0.470{\footnotesize $\pm$0.054} & 0.272{\footnotesize $\pm$0.029} & 0.355{\footnotesize $\pm$0.043} & 0.561{\footnotesize $\pm$0.273} & \second{0.221{\footnotesize $\pm$0.023}} & \best{0.170{\footnotesize $\pm$0.001}} \\
\midrule

\multirow{4}{*}{\rotatebox{90}{ETTh1-L}}
& 96  & 0.395{\footnotesize $\pm$0.052} & 0.437{\footnotesize $\pm$0.018} & 0.455{\footnotesize $\pm$0.046} & 0.293{\footnotesize $\pm$0.034} & 0.333{\footnotesize $\pm$0.045} & 0.379{\footnotesize $\pm$0.030} & \second{0.264{\footnotesize $\pm$0.020}} & \best{0.240{\footnotesize $\pm$0.001}} \\
& 192 & 0.467{\footnotesize $\pm$0.044} & 0.496{\footnotesize $\pm$0.051} & 0.516{\footnotesize $\pm$0.038} & 0.348{\footnotesize $\pm$0.075} & 0.351{\footnotesize $\pm$0.063} & 0.425{\footnotesize $\pm$0.019} & \second{0.290{\footnotesize $\pm$0.016}} & \best{0.265{\footnotesize $\pm$0.001}} \\
& 336 & 0.450{\footnotesize $\pm$0.027} & 0.454{\footnotesize $\pm$0.025} & 0.512{\footnotesize $\pm$0.026} & 0.377{\footnotesize $\pm$0.026} & 0.371{\footnotesize $\pm$0.031} & 0.458{\footnotesize $\pm$0.054} & \second{0.308{\footnotesize $\pm$0.021}} & \best{0.287{\footnotesize $\pm$0.004}} \\
& 720 & 0.516{\footnotesize $\pm$0.027} & 0.528{\footnotesize $\pm$0.012} & 0.523{\footnotesize $\pm$0.027} & 0.393{\footnotesize $\pm$0.043} & 0.363{\footnotesize $\pm$0.053} & 0.502{\footnotesize $\pm$0.039} & \second{0.314{\footnotesize $\pm$0.011}} & \best{0.289{\footnotesize $\pm$0.005}} \\
\midrule

\multirow{4}{*}{\rotatebox{90}{ETTh2-L}}
& 96  & 0.336{\footnotesize $\pm$0.021} & 0.164{\footnotesize $\pm$0.013} & 0.358{\footnotesize $\pm$0.026} & 0.239{\footnotesize $\pm$0.019} & 0.263{\footnotesize $\pm$0.020} & 0.432{\footnotesize $\pm$0.141} & \second{0.162{\footnotesize $\pm$0.009}} & \best{0.134{\footnotesize $\pm$0.001}} \\
& 192 & 0.265{\footnotesize $\pm$0.043} & 0.226{\footnotesize $\pm$0.018} & 0.457{\footnotesize $\pm$0.081} & 0.313{\footnotesize $\pm$0.034} & 0.273{\footnotesize $\pm$0.024} & 0.625{\footnotesize $\pm$0.170} & \second{0.186{\footnotesize $\pm$0.018}} & \best{0.155{\footnotesize $\pm$0.002}} \\
& 336 & 0.350{\footnotesize $\pm$0.031} & 0.274{\footnotesize $\pm$0.022} & 0.481{\footnotesize $\pm$0.078} & 0.376{\footnotesize $\pm$0.024} & 0.265{\footnotesize $\pm$0.042} & 0.793{\footnotesize $\pm$0.319} & \second{0.257{\footnotesize $\pm$0.023}} & \best{0.173{\footnotesize $\pm$0.002}} \\
& 720 & 0.406{\footnotesize $\pm$0.056} & 0.302{\footnotesize $\pm$0.040} & 0.445{\footnotesize $\pm$0.016} & 0.990{\footnotesize $\pm$0.023} & 0.327{\footnotesize $\pm$0.033} & 0.539{\footnotesize $\pm$0.090} & \second{0.280{\footnotesize $\pm$0.014}} & \best{0.177{\footnotesize $\pm$0.002}} \\
\midrule

\multirow{4}{*}{\rotatebox{90}{Elec.-L}}
& 96  & 0.344{\footnotesize $\pm$0.006} & 0.153{\footnotesize $\pm$0.137} & 0.096{\footnotesize $\pm$0.002} & 0.083{\footnotesize $\pm$0.009} & 0.088{\footnotesize $\pm$0.014} & 0.094{\footnotesize $\pm$0.003} & \second{0.073{\footnotesize $\pm$0.002}} & \best{0.066{\footnotesize $\pm$0.000}} \\
& 192 & 0.345{\footnotesize $\pm$0.006} & 0.200{\footnotesize $\pm$0.094} & 0.100{\footnotesize $\pm$0.004} & 0.093{\footnotesize $\pm$0.024} & 0.097{\footnotesize $\pm$0.009} & 0.097{\footnotesize $\pm$0.002} & \second{0.080{\footnotesize $\pm$0.004}} & \best{0.074{\footnotesize $\pm$0.000}} \\
& 336 & 0.462{\footnotesize $\pm$0.054} & -- & 0.102{\footnotesize $\pm$0.007} & 0.095{\footnotesize $\pm$0.001} & -- & 0.099{\footnotesize $\pm$0.001} & \best{0.054{\footnotesize $\pm$0.001}} & \second{0.081{\footnotesize $\pm$0.000}} \\
& 720 & 0.478{\footnotesize $\pm$0.005} & -- & 0.108{\footnotesize $\pm$0.003} & 0.106{\footnotesize $\pm$0.007} & -- & 0.114{\footnotesize $\pm$0.013} & \best{0.057{\footnotesize $\pm$0.005}} & \second{0.099{\footnotesize $\pm$0.000}} \\
\midrule

\multirow{4}{*}{\rotatebox{90}{Exch.-L}}
& 96  & 0.079{\footnotesize $\pm$0.007} & 0.028{\footnotesize $\pm$0.003} & 0.068{\footnotesize $\pm$0.003} & \second{0.026{\footnotesize $\pm$0.010}} & 0.028{\footnotesize $\pm$0.002} & 0.071{\footnotesize $\pm$0.006} & 0.031{\footnotesize $\pm$0.002} & \best{0.020{\footnotesize $\pm$0.001}} \\
& 192 & 0.093{\footnotesize $\pm$0.011} & 0.045{\footnotesize $\pm$0.003} & 0.087{\footnotesize $\pm$0.013} & 0.034{\footnotesize $\pm$0.009} & 0.046{\footnotesize $\pm$0.017} & 0.068{\footnotesize $\pm$0.004} & \second{0.032{\footnotesize $\pm$0.010}} & \best{0.029{\footnotesize $\pm$0.001}} \\
& 336 & 0.081{\footnotesize $\pm$0.010} & 0.060{\footnotesize $\pm$0.004} & 0.074{\footnotesize $\pm$0.009} & 0.058{\footnotesize $\pm$0.023} & \second{0.045{\footnotesize $\pm$0.010}} & 0.072{\footnotesize $\pm$0.002} & 0.048{\footnotesize $\pm$0.004} & \best{0.040{\footnotesize $\pm$0.002}} \\
& 720 & 0.082{\footnotesize $\pm$0.010} & 0.143{\footnotesize $\pm$0.020} & 0.099{\footnotesize $\pm$0.015} & 0.160{\footnotesize $\pm$0.019} & 0.148{\footnotesize $\pm$0.017} & 0.079{\footnotesize $\pm$0.009} & \second{0.069{\footnotesize $\pm$0.005}} & \best{0.068{\footnotesize $\pm$0.002}} \\
\midrule

\multirow{4}{*}{\rotatebox{90}{Traffic-L}}
& 96  & 0.294{\footnotesize $\pm$0.003} & -- & 0.202{\footnotesize $\pm$0.004} & 0.215{\footnotesize $\pm$0.003} & 0.208{\footnotesize $\pm$0.004} & 0.187{\footnotesize $\pm$0.002} & \best{0.086{\footnotesize $\pm$0.001}} & \second{0.172{\footnotesize $\pm$0.001}} \\
& 192 & 0.306{\footnotesize $\pm$0.004} & -- & 0.208{\footnotesize $\pm$0.003} & -- & -- & 0.192{\footnotesize $\pm$0.001} & \best{0.088{\footnotesize $\pm$0.002}} & \second{0.178{\footnotesize $\pm$0.000}} \\
& 336 & 0.317{\footnotesize $\pm$0.006} & -- & 0.213{\footnotesize $\pm$0.003} & -- & -- & 0.201{\footnotesize $\pm$0.004} & \second{0.195{\footnotesize $\pm$0.003}} & \best{0.192{\footnotesize $\pm$0.000}} \\
& 720 & 0.391{\footnotesize $\pm$0.002} & -- & 0.220{\footnotesize $\pm$0.002} & -- & -- & \second{0.211{\footnotesize $\pm$0.004}} & \best{0.200{\footnotesize $\pm$0.001}} & 0.228{\footnotesize $\pm$0.003} \\
\midrule

\multirow{4}{*}{\rotatebox{90}{Weather-L}}
& 96  & 0.104{\footnotesize $\pm$0.020} & \second{0.068{\footnotesize $\pm$0.008}} & 0.130{\footnotesize $\pm$0.017} & 0.139{\footnotesize $\pm$0.008} & 0.105{\footnotesize $\pm$0.011} & 0.116{\footnotesize $\pm$0.013} & 0.080{\footnotesize $\pm$0.007} & \best{0.053{\footnotesize $\pm$0.001}} \\
& 192 & 0.134{\footnotesize $\pm$0.012} & \second{0.068{\footnotesize $\pm$0.006}} & 0.127{\footnotesize $\pm$0.019} & 0.143{\footnotesize $\pm$0.020} & 0.142{\footnotesize $\pm$0.022} & 0.122{\footnotesize $\pm$0.021} & 0.079{\footnotesize $\pm$0.009} & \best{0.058{\footnotesize $\pm$0.001}} \\
& 336 & 0.137{\footnotesize $\pm$0.010} & 0.083{\footnotesize $\pm$0.002} & 0.130{\footnotesize $\pm$0.006} & 0.129{\footnotesize $\pm$0.012} & 0.133{\footnotesize $\pm$0.014} & 0.128{\footnotesize $\pm$0.011} & \second{0.082{\footnotesize $\pm$0.010}} & \best{0.062{\footnotesize $\pm$0.000}} \\
& 720 & 0.152{\footnotesize $\pm$0.003} & 0.087{\footnotesize $\pm$0.003} & 0.113{\footnotesize $\pm$0.011} & 0.122{\footnotesize $\pm$0.006} & 0.113{\footnotesize $\pm$0.004} & 0.110{\footnotesize $\pm$0.011} & \second{0.084{\footnotesize $\pm$0.005}} & \best{0.068{\footnotesize $\pm$0.001}} \\
\midrule

\multirow{4}{*}{\rotatebox{90}{ILI-L}}
& 24  & 0.228{\footnotesize $\pm$0.024} & 0.250{\footnotesize $\pm$0.013} & 0.275{\footnotesize $\pm$0.047} & 0.097{\footnotesize $\pm$0.010} & 0.092{\footnotesize $\pm$0.019} & 0.257{\footnotesize $\pm$0.003} & \second{0.087{\footnotesize $\pm$0.003}} & \best{0.067{\footnotesize $\pm$0.010}} \\
& 36  & 0.235{\footnotesize $\pm$0.010} & 0.285{\footnotesize $\pm$0.010} & 0.272{\footnotesize $\pm$0.057} & 0.117{\footnotesize $\pm$0.017} & 0.115{\footnotesize $\pm$0.011} & 0.281{\footnotesize $\pm$0.004} & \second{0.113{\footnotesize $\pm$0.005}} & \best{0.088{\footnotesize $\pm$0.003}} \\
& 48  & 0.265{\footnotesize $\pm$0.039} & 0.285{\footnotesize $\pm$0.036} & 0.295{\footnotesize $\pm$0.033} & 0.128{\footnotesize $\pm$0.019} & 0.133{\footnotesize $\pm$0.022} & 0.288{\footnotesize $\pm$0.008} & \second{0.124{\footnotesize $\pm$0.010}} & \best{0.108{\footnotesize $\pm$0.014}} \\
& 60  & 0.263{\footnotesize $\pm$0.022} & 0.283{\footnotesize $\pm$0.013} & 0.295{\footnotesize $\pm$0.083} & 0.172{\footnotesize $\pm$0.034} & 0.155{\footnotesize $\pm$0.018} & 0.307{\footnotesize $\pm$0.005} & \second{0.142{\footnotesize $\pm$0.008}} & \best{0.124{\footnotesize $\pm$0.003}} \\

\bottomrule
\end{tabular}
\end{adjustbox}

\endgroup
\end{table}

\begin{table}[!htbp]
\centering
\begingroup
\setlength{\tabcolsep}{2pt}
\renewcommand{\arraystretch}{1.1}
\caption{Full long-term probabilistic forecasting results across all baselines and datasets; protocol as in Table~\ref{tab:ltsf}. Lower NMAE values indicate better predictions. Results are reported as mean\,$\pm$\,standard deviation over 5 independent runs of retraining and evaluation. \best{Red bold}: best; \second{blue underline}: second best. Due to the excessive time and memory consumption, some results are unavailable and denoted as ``--''.}
\label{tab:full:ltsf:nmae}

\begin{adjustbox}{max width=\linewidth, max totalheight=0.92\textheight}
\begin{tabular}{ll cccccc cc c}
\toprule
Dataset & Horizon &
TSDiff & CSDI & TimeGrad &
GRU MAF & Trans MAF & GRU NVP &
$K^2$VAE &
\textbf{KGO (Ours)} \\
\midrule

\multirow{4}{*}{\rotatebox{90}{ETTm1-L}}
& 96  & 0.441{\footnotesize $\pm$0.021} & 0.308{\footnotesize $\pm$0.005} & 0.645{\footnotesize $\pm$0.129} & 0.402{\footnotesize $\pm$0.087} & 0.456{\footnotesize $\pm$0.042} & 0.488{\footnotesize $\pm$0.058} & \second{0.284{\footnotesize $\pm$0.011}} & \best{0.270{\footnotesize $\pm$0.002}} \\
& 192 & 0.441{\footnotesize $\pm$0.019} & 0.377{\footnotesize $\pm$0.026} & 0.748{\footnotesize $\pm$0.084} & 0.476{\footnotesize $\pm$0.046} & 0.553{\footnotesize $\pm$0.012} & 0.514{\footnotesize $\pm$0.042} & \second{0.323{\footnotesize $\pm$0.020}} & \best{0.297{\footnotesize $\pm$0.002}} \\
& 336 & 0.571{\footnotesize $\pm$0.033} & 0.419{\footnotesize $\pm$0.042} & 0.759{\footnotesize $\pm$0.015} & 0.522{\footnotesize $\pm$0.019} & 0.590{\footnotesize $\pm$0.047} & 0.630{\footnotesize $\pm$0.029} & \second{0.330{\footnotesize $\pm$0.014}} & \best{0.316{\footnotesize $\pm$0.000}} \\
& 720 & 0.622{\footnotesize $\pm$0.045} & 0.578{\footnotesize $\pm$0.051} & 0.793{\footnotesize $\pm$0.034} & 0.711{\footnotesize $\pm$0.081} & 0.822{\footnotesize $\pm$0.034} & 0.707{\footnotesize $\pm$0.050} & \second{0.373{\footnotesize $\pm$0.032}} & \best{0.349{\footnotesize $\pm$0.001}} \\
\midrule

\multirow{4}{*}{\rotatebox{90}{ETTm2-L}}
& 96  & 0.224{\footnotesize $\pm$0.033} & 0.146{\footnotesize $\pm$0.012} & 0.525{\footnotesize $\pm$0.047} & 0.212{\footnotesize $\pm$0.082} & 0.279{\footnotesize $\pm$0.031} & 0.413{\footnotesize $\pm$0.059} & \second{0.144{\footnotesize $\pm$0.011}} & \best{0.133{\footnotesize $\pm$0.001}} \\
& 192 & 0.316{\footnotesize $\pm$0.040} & 0.189{\footnotesize $\pm$0.012} & 0.530{\footnotesize $\pm$0.060} & 0.535{\footnotesize $\pm$0.029} & 0.292{\footnotesize $\pm$0.041} & 0.427{\footnotesize $\pm$0.033} & \second{0.170{\footnotesize $\pm$0.009}} & \best{0.162{\footnotesize $\pm$0.001}} \\
& 336 & 0.397{\footnotesize $\pm$0.051} & 0.248{\footnotesize $\pm$0.024} & 0.566{\footnotesize $\pm$0.047} & 0.407{\footnotesize $\pm$0.043} & 0.309{\footnotesize $\pm$0.032} & 0.580{\footnotesize $\pm$0.169} & \second{0.187{\footnotesize $\pm$0.021}} & \best{0.184{\footnotesize $\pm$0.005}} \\
& 720 & 0.416{\footnotesize $\pm$0.065} & 0.306{\footnotesize $\pm$0.040} & 0.561{\footnotesize $\pm$0.044} & 0.355{\footnotesize $\pm$0.048} & 0.475{\footnotesize $\pm$0.029} & 0.749{\footnotesize $\pm$0.385} & \second{0.275{\footnotesize $\pm$0.035}} & \best{0.213{\footnotesize $\pm$0.002}} \\
\midrule

\multirow{4}{*}{\rotatebox{90}{ETTh1-L}}
& 96  & 0.510{\footnotesize $\pm$0.029} & 0.557{\footnotesize $\pm$0.022} & 0.585{\footnotesize $\pm$0.058} & 0.371{\footnotesize $\pm$0.034} & 0.423{\footnotesize $\pm$0.047} & 0.481{\footnotesize $\pm$0.037} & \second{0.336{\footnotesize $\pm$0.041}} & \best{0.309{\footnotesize $\pm$0.002}} \\
& 192 & 0.596{\footnotesize $\pm$0.056} & 0.625{\footnotesize $\pm$0.065} & 0.680{\footnotesize $\pm$0.058} & 0.430{\footnotesize $\pm$0.022} & 0.451{\footnotesize $\pm$0.012} & 0.531{\footnotesize $\pm$0.018} & \second{0.372{\footnotesize $\pm$0.023}} & \best{0.341{\footnotesize $\pm$0.001}} \\
& 336 & 0.581{\footnotesize $\pm$0.035} & 0.574{\footnotesize $\pm$0.026} & 0.666{\footnotesize $\pm$0.047} & 0.462{\footnotesize $\pm$0.049} & 0.481{\footnotesize $\pm$0.041} & 0.580{\footnotesize $\pm$0.064} & \second{0.394{\footnotesize $\pm$0.022}} & \best{0.365{\footnotesize $\pm$0.004}} \\
& 720 & 0.657{\footnotesize $\pm$0.017} & 0.657{\footnotesize $\pm$0.014} & 0.672{\footnotesize $\pm$0.015} & 0.496{\footnotesize $\pm$0.019} & 0.455{\footnotesize $\pm$0.025} & 0.643{\footnotesize $\pm$0.046} & \second{0.396{\footnotesize $\pm$0.012}} & \best{0.375{\footnotesize $\pm$0.006}} \\
\midrule

\multirow{4}{*}{\rotatebox{90}{ETTh2-L}}
& 96  & 0.421{\footnotesize $\pm$0.033} & 0.214{\footnotesize $\pm$0.018} & 0.448{\footnotesize $\pm$0.031} & 0.292{\footnotesize $\pm$0.012} & 0.345{\footnotesize $\pm$0.042} & 0.548{\footnotesize $\pm$0.158} & \second{0.189{\footnotesize $\pm$0.010}} & \best{0.170{\footnotesize $\pm$0.002}} \\
& 192 & 0.339{\footnotesize $\pm$0.033} & 0.294{\footnotesize $\pm$0.027} & 0.575{\footnotesize $\pm$0.089} & 0.376{\footnotesize $\pm$0.112} & 0.343{\footnotesize $\pm$0.044} & 0.766{\footnotesize $\pm$0.223} & \second{0.213{\footnotesize $\pm$0.021}} & \best{0.198{\footnotesize $\pm$0.003}} \\
& 336 & 0.427{\footnotesize $\pm$0.041} & 0.353{\footnotesize $\pm$0.028} & 0.606{\footnotesize $\pm$0.095} & 0.454{\footnotesize $\pm$0.057} & 0.333{\footnotesize $\pm$0.078} & 0.942{\footnotesize $\pm$0.408} & \second{0.263{\footnotesize $\pm$0.039}} & \best{0.220{\footnotesize $\pm$0.003}} \\
& 720 & 0.482{\footnotesize $\pm$0.022} & 0.382{\footnotesize $\pm$0.030} & 0.550{\footnotesize $\pm$0.018} & 1.092{\footnotesize $\pm$0.019} & 0.412{\footnotesize $\pm$0.020} & 0.688{\footnotesize $\pm$0.161} & \second{0.278{\footnotesize $\pm$0.020}} & \best{0.228{\footnotesize $\pm$0.002}} \\
\midrule

\multirow{4}{*}{\rotatebox{90}{Elec.-L}}
& 96  & 0.441{\footnotesize $\pm$0.013} & 0.203{\footnotesize $\pm$0.189} & 0.119{\footnotesize $\pm$0.003} & 0.108{\footnotesize $\pm$0.009} & 0.114{\footnotesize $\pm$0.010} & 0.118{\footnotesize $\pm$0.003} & \second{0.093{\footnotesize $\pm$0.002}} & \best{0.079{\footnotesize $\pm$0.000}} \\
& 192 & 0.441{\footnotesize $\pm$0.005} & 0.264{\footnotesize $\pm$0.129} & 0.124{\footnotesize $\pm$0.005} & 0.120{\footnotesize $\pm$0.033} & 0.131{\footnotesize $\pm$0.008} & 0.121{\footnotesize $\pm$0.003} & \second{0.102{\footnotesize $\pm$0.010}} & \best{0.087{\footnotesize $\pm$0.000}} \\
& 336 & 0.571{\footnotesize $\pm$0.022} & -- & 0.126{\footnotesize $\pm$0.008} & 0.122{\footnotesize $\pm$0.018} & -- & 0.123{\footnotesize $\pm$0.001} & \second{0.107{\footnotesize $\pm$0.002}} & \best{0.096{\footnotesize $\pm$0.000}} \\
& 720 & 0.622{\footnotesize $\pm$0.142} & -- & 0.134{\footnotesize $\pm$0.004} & 0.136{\footnotesize $\pm$0.098} & -- & 0.144{\footnotesize $\pm$0.017} & \best{0.117{\footnotesize $\pm$0.019}} & \second{0.118{\footnotesize $\pm$0.001}} \\
\midrule

\multirow{4}{*}{\rotatebox{90}{Exch.-L}}
& 96  & 0.090{\footnotesize $\pm$0.010} & 0.036{\footnotesize $\pm$0.005} & 0.079{\footnotesize $\pm$0.002} & 0.033{\footnotesize $\pm$0.003} & 0.036{\footnotesize $\pm$0.009} & 0.091{\footnotesize $\pm$0.009} & \second{0.032{\footnotesize $\pm$0.002}} & \best{0.025{\footnotesize $\pm$0.001}} \\
& 192 & 0.106{\footnotesize $\pm$0.010} & 0.058{\footnotesize $\pm$0.005} & 0.100{\footnotesize $\pm$0.019} & 0.044{\footnotesize $\pm$0.004} & 0.058{\footnotesize $\pm$0.007} & 0.087{\footnotesize $\pm$0.005} & \second{0.040{\footnotesize $\pm$0.005}} & \best{0.036{\footnotesize $\pm$0.001}} \\
& 336 & 0.106{\footnotesize $\pm$0.010} & 0.076{\footnotesize $\pm$0.006} & 0.086{\footnotesize $\pm$0.008} & 0.074{\footnotesize $\pm$0.017} & 0.058{\footnotesize $\pm$0.009} & 0.091{\footnotesize $\pm$0.002} & \second{0.054{\footnotesize $\pm$0.001}} & \best{0.050{\footnotesize $\pm$0.002}} \\
& 720 & 0.142{\footnotesize $\pm$0.009} & 0.173{\footnotesize $\pm$0.020} & 0.113{\footnotesize $\pm$0.016} & 0.182{\footnotesize $\pm$0.010} & 0.191{\footnotesize $\pm$0.006} & 0.103{\footnotesize $\pm$0.009} & \second{0.084{\footnotesize $\pm$0.017}} & \best{0.083{\footnotesize $\pm$0.002}} \\
\midrule

\multirow{4}{*}{\rotatebox{90}{Traffic-L}}
& 96  & 0.342{\footnotesize $\pm$0.042} & -- & 0.234{\footnotesize $\pm$0.006} & 0.274{\footnotesize $\pm$0.012} & 0.265{\footnotesize $\pm$0.007} & 0.231{\footnotesize $\pm$0.003} & \second{0.230{\footnotesize $\pm$0.010}} & \best{0.198{\footnotesize $\pm$0.001}} \\
& 192 & 0.354{\footnotesize $\pm$0.012} & -- & 0.239{\footnotesize $\pm$0.004} & -- & -- & 0.236{\footnotesize $\pm$0.002} & \second{0.234{\footnotesize $\pm$0.003}} & \best{0.205{\footnotesize $\pm$0.000}} \\
& 336 & 0.392{\footnotesize $\pm$0.006} & -- & 0.246{\footnotesize $\pm$0.003} & -- & -- & 0.248{\footnotesize $\pm$0.006} & \second{0.242{\footnotesize $\pm$0.007}} & \best{0.222{\footnotesize $\pm$0.000}} \\
& 720 & 0.478{\footnotesize $\pm$0.006} & -- & \second{0.263{\footnotesize $\pm$0.001}} & -- & -- & 0.264{\footnotesize $\pm$0.006} & \best{0.248{\footnotesize $\pm$0.010}} & 0.269{\footnotesize $\pm$0.003} \\
\midrule

\multirow{4}{*}{\rotatebox{90}{Weather-L}}
& 96  & 0.113{\footnotesize $\pm$0.022} & 0.087{\footnotesize $\pm$0.012} & 0.164{\footnotesize $\pm$0.023} & 0.176{\footnotesize $\pm$0.011} & 0.139{\footnotesize $\pm$0.010} & 0.145{\footnotesize $\pm$0.017} & \second{0.086{\footnotesize $\pm$0.011}} & \best{0.065{\footnotesize $\pm$0.001}} \\
& 192 & 0.144{\footnotesize $\pm$0.020} & 0.086{\footnotesize $\pm$0.007} & 0.158{\footnotesize $\pm$0.024} & 0.166{\footnotesize $\pm$0.022} & 0.160{\footnotesize $\pm$0.037} & 0.147{\footnotesize $\pm$0.025} & \second{0.083{\footnotesize $\pm$0.011}} & \best{0.072{\footnotesize $\pm$0.001}} \\
& 336 & 0.138{\footnotesize $\pm$0.033} & 0.098{\footnotesize $\pm$0.002} & 0.162{\footnotesize $\pm$0.006} & 0.168{\footnotesize $\pm$0.014} & 0.170{\footnotesize $\pm$0.027} & 0.160{\footnotesize $\pm$0.012} & \second{0.093{\footnotesize $\pm$0.010}} & \best{0.076{\footnotesize $\pm$0.001}} \\
& 720 & 0.141{\footnotesize $\pm$0.026} & 0.102{\footnotesize $\pm$0.005} & 0.136{\footnotesize $\pm$0.020} & 0.149{\footnotesize $\pm$0.034} & 0.148{\footnotesize $\pm$0.040} & 0.135{\footnotesize $\pm$0.008} & \second{0.099{\footnotesize $\pm$0.009}} & \best{0.083{\footnotesize $\pm$0.001}} \\
\midrule

\multirow{4}{*}{\rotatebox{90}{ILI-L}}
& 24  & 0.242{\footnotesize $\pm$0.086} & 0.263{\footnotesize $\pm$0.012} & 0.296{\footnotesize $\pm$0.044} & 0.124{\footnotesize $\pm$0.019} & 0.118{\footnotesize $\pm$0.033} & 0.283{\footnotesize $\pm$0.001} & \second{0.116{\footnotesize $\pm$0.011}} & \best{0.089{\footnotesize $\pm$0.011}} \\
& 36  & 0.246{\footnotesize $\pm$0.117} & 0.298{\footnotesize $\pm$0.011} & 0.298{\footnotesize $\pm$0.048} & 0.144{\footnotesize $\pm$0.011} & 0.143{\footnotesize $\pm$0.089} & 0.307{\footnotesize $\pm$0.007} & \second{0.142{\footnotesize $\pm$0.008}} & \best{0.109{\footnotesize $\pm$0.004}} \\
& 48  & 0.275{\footnotesize $\pm$0.044} & 0.301{\footnotesize $\pm$0.034} & 0.320{\footnotesize $\pm$0.025} & 0.159{\footnotesize $\pm$0.020} & 0.160{\footnotesize $\pm$0.039} & 0.314{\footnotesize $\pm$0.009} & \second{0.152{\footnotesize $\pm$0.017}} & \best{0.130{\footnotesize $\pm$0.014}} \\
& 60  & 0.272{\footnotesize $\pm$0.020} & 0.299{\footnotesize $\pm$0.013} & 0.325{\footnotesize $\pm$0.068} & 0.216{\footnotesize $\pm$0.014} & 0.183{\footnotesize $\pm$0.019} & 0.333{\footnotesize $\pm$0.005} & \second{0.167{\footnotesize $\pm$0.007}} & \best{0.146{\footnotesize $\pm$0.004}} \\

\bottomrule
\end{tabular}
\end{adjustbox}

\endgroup
\end{table}

\clearpage
\begin{table}[!htbp]
\centering
\begingroup
\setlength{\tabcolsep}{2.2pt}
\renewcommand{\arraystretch}{1.35}
\caption{Full short-term probabilistic forecasting results across all baselines and datasets; protocol as in Table~\ref{tab:stsf}. Lower CRPS values indicate better predictions. Each cell reports the mean $\pm$ standard deviation of the CRPS across 5 independent runs of retraining and evaluation. \best{Red bold}: best; \second{blue underline}: second best.}
\label{tab:full:stsf_CRPS}

\noindent\makebox[\linewidth][c]{%
\begin{adjustbox}{max width=\linewidth, max totalheight=0.92\textheight}
\begin{tabular}{c c c c c c c c c c}
\toprule
Dataset & TSDiff & $D^3$VAE & CSDI & TimeGrad & GRU MAF & Trans MAF & GRU NVP & $K^2$VAE & \textbf{KGO (Ours)} \\
\midrule
ETTm1-S & 0.209{\footnotesize $\pm$0.013} & 0.198{\footnotesize $\pm$0.015} & 0.140{\footnotesize $\pm$0.012} & 0.186{\footnotesize $\pm$0.003} & 0.151{\footnotesize $\pm$0.009} & 0.139{\footnotesize $\pm$0.005} & 0.455{\footnotesize $\pm$0.029} & \second{0.135{\footnotesize $\pm$0.008}} & \best{0.104{\footnotesize $\pm$0.002}} \\
ETTm2-S & 0.124{\footnotesize $\pm$0.008} & 0.303{\footnotesize $\pm$0.024} & 0.144{\footnotesize $\pm$0.018} & 0.148{\footnotesize $\pm$0.004} & 0.146{\footnotesize $\pm$0.011} & 0.180{\footnotesize $\pm$0.010} & 0.276{\footnotesize $\pm$0.014} & \second{0.122{\footnotesize $\pm$0.008}} & \best{0.092{\footnotesize $\pm$0.002}} \\
ETTh1-S & 0.304{\footnotesize $\pm$0.016} & 0.324{\footnotesize $\pm$0.019} & 0.262{\footnotesize $\pm$0.012} & 0.273{\footnotesize $\pm$0.007} & 0.258{\footnotesize $\pm$0.013} & 0.309{\footnotesize $\pm$0.009} & 0.398{\footnotesize $\pm$0.034} & \second{0.256{\footnotesize $\pm$0.008}} & \best{0.228{\footnotesize $\pm$0.001}} \\
ETTh2-S & 0.204{\footnotesize $\pm$0.006} & 0.216{\footnotesize $\pm$0.015} & 0.133{\footnotesize $\pm$0.006} & 0.184{\footnotesize $\pm$0.006} & 0.160{\footnotesize $\pm$0.008} & 0.200{\footnotesize $\pm$0.012} & 0.309{\footnotesize $\pm$0.023} & \second{0.128{\footnotesize $\pm$0.006}} & \best{0.104{\footnotesize $\pm$0.001}} \\
Electricity-S & 0.111{\footnotesize $\pm$0.013} & 0.071{\footnotesize $\pm$0.009} & \best{0.051{\footnotesize $\pm$0.001}} & 0.057{\footnotesize $\pm$0.002} & 0.056{\footnotesize $\pm$0.002} & 0.054{\footnotesize $\pm$0.002} & 0.062{\footnotesize $\pm$0.003} & 0.053{\footnotesize $\pm$0.002} & \second{0.052{\footnotesize $\pm$0.000}} \\
Exchange-S & 0.077{\footnotesize $\pm$0.019} & 0.011{\footnotesize $\pm$0.002} & \second{0.009{\footnotesize $\pm$0.001}} & \second{0.009{\footnotesize $\pm$0.001}} & 0.012{\footnotesize $\pm$0.003} & 0.012{\footnotesize $\pm$0.001} & 0.019{\footnotesize $\pm$0.006} & \second{0.009{\footnotesize $\pm$0.001}} & \best{0.007{\footnotesize $\pm$0.000}} \\
Traffic-S & 0.189{\footnotesize $\pm$0.009} & 0.143{\footnotesize $\pm$0.008} & 0.147{\footnotesize $\pm$0.014} & 0.130{\footnotesize $\pm$0.005} & 0.144{\footnotesize $\pm$0.022} & 0.133{\footnotesize $\pm$0.004} & 0.168{\footnotesize $\pm$0.008} & \second{0.129{\footnotesize $\pm$0.004}} & \best{0.127{\footnotesize $\pm$0.000}} \\
Solar-S & 0.568{\footnotesize $\pm$0.015} & 0.769{\footnotesize $\pm$0.029} & \second{0.392{\footnotesize $\pm$0.006}} & 0.465{\footnotesize $\pm$0.016} & 0.486{\footnotesize $\pm$0.007} & 0.442{\footnotesize $\pm$0.011} & 0.530{\footnotesize $\pm$0.008} & \best{0.367{\footnotesize $\pm$0.005}} & 0.456{\footnotesize $\pm$0.010} \\
\bottomrule
\end{tabular}
\end{adjustbox}%
}

\vspace{4pt}
\raggedright
\endgroup
\end{table}

\begin{table}[!htbp]
\centering
\begingroup
\setlength{\tabcolsep}{2.2pt}
\renewcommand{\arraystretch}{1.35}
\caption{Full short-term probabilistic forecasting results across all baselines and datasets; protocol as in Table~\ref{tab:stsf}. Lower NMAE values indicate better predictions. Results are reported as mean\,$\pm$\,standard deviation over 5 independent runs of retraining and evaluation. \best{Red bold}: best; \second{blue underline}: second best.}
\label{tab:full:stsf_NMAE}

\noindent\makebox[\linewidth][c]{%
\begin{adjustbox}{max width=\linewidth, max totalheight=0.92\textheight}
\begin{tabular}{c cccccccc c}
\toprule
\multicolumn{1}{c}{Dataset} &
TSDiff & $D^3$VAE & CSDI & TimeGrad &
GRU MAF & Trans MAF & GRU NVP &
$K^2$VAE &
\textbf{KGO (Ours)} \\
\midrule
ETTm1-S       & 0.276{\footnotesize $\pm$0.008} & 0.250{\footnotesize $\pm$0.018} & 0.169{\footnotesize $\pm$0.021} & 0.246{\footnotesize $\pm$0.007} & 0.198{\footnotesize $\pm$0.004} & 0.162{\footnotesize $\pm$0.006} & 0.584{\footnotesize $\pm$0.047} & \second{0.152{\footnotesize $\pm$0.007}} & \best{0.133{\footnotesize $\pm$0.003}} \\
ETTm2-S       & 0.162{\footnotesize $\pm$0.008} & 0.378{\footnotesize $\pm$0.031} & 0.181{\footnotesize $\pm$0.024} & 0.189{\footnotesize $\pm$0.006} & 0.193{\footnotesize $\pm$0.008} & 0.224{\footnotesize $\pm$0.009} & 0.349{\footnotesize $\pm$0.028} & \second{0.146{\footnotesize $\pm$0.009}} & \best{0.118{\footnotesize $\pm$0.003}} \\
ETTh1-S       & 0.400{\footnotesize $\pm$0.025} & 0.410{\footnotesize $\pm$0.016} & 0.339{\footnotesize $\pm$0.009} & 0.356{\footnotesize $\pm$0.013} & 0.326{\footnotesize $\pm$0.016} & 0.400{\footnotesize $\pm$0.011} & 0.477{\footnotesize $\pm$0.040} & \second{0.312{\footnotesize $\pm$0.008}} & \best{0.280{\footnotesize $\pm$0.002}} \\
ETTh2-S       & 0.272{\footnotesize $\pm$0.015} & 0.267{\footnotesize $\pm$0.018} & 0.161{\footnotesize $\pm$0.013} & 0.224{\footnotesize $\pm$0.014} & 0.208{\footnotesize $\pm$0.003} & 0.256{\footnotesize $\pm$0.009} & 0.375{\footnotesize $\pm$0.024} & \second{0.140{\footnotesize $\pm$0.007}} & \best{0.130{\footnotesize $\pm$0.001}} \\
Electricity-S & 0.115{\footnotesize $\pm$0.018} & 0.092{\footnotesize $\pm$0.013} & \second{0.066{\footnotesize $\pm$0.001}} & 0.073{\footnotesize $\pm$0.004} & 0.073{\footnotesize $\pm$0.004} & 0.071{\footnotesize $\pm$0.003} & 0.081{\footnotesize $\pm$0.006} & 0.068{\footnotesize $\pm$0.002} & \best{0.064{\footnotesize $\pm$0.001}} \\
Exchange-S    & 0.096{\footnotesize $\pm$0.024} & \second{0.012{\footnotesize $\pm$0.002}} & 0.013{\footnotesize $\pm$0.001} & \second{0.012{\footnotesize $\pm$0.002}} & 0.016{\footnotesize $\pm$0.002} & 0.016{\footnotesize $\pm$0.001} & 0.024{\footnotesize $\pm$0.007} & \best{0.009{\footnotesize $\pm$0.001}} & \best{0.009{\footnotesize $\pm$0.000}} \\
Traffic-S     & 0.206{\footnotesize $\pm$0.011} & 0.178{\footnotesize $\pm$0.013} & 0.175{\footnotesize $\pm$0.013} & \second{0.155{\footnotesize $\pm$0.007}} & 0.182{\footnotesize $\pm$0.029} & 0.160{\footnotesize $\pm$0.006} & 0.209{\footnotesize $\pm$0.013} & 0.157{\footnotesize $\pm$0.007} & \best{0.152{\footnotesize $\pm$0.001}} \\
Solar-S       & 0.635{\footnotesize $\pm$0.012} & 0.998{\footnotesize $\pm$0.049} & \second{0.533{\footnotesize $\pm$0.007}} & 0.609{\footnotesize $\pm$0.015} & 0.603{\footnotesize $\pm$0.009} & 0.577{\footnotesize $\pm$0.014} & 0.670{\footnotesize $\pm$0.011} & \best{0.480{\footnotesize $\pm$0.008}} & 0.569{\footnotesize $\pm$0.012} \\
\bottomrule
\end{tabular}
\end{adjustbox}%
}

\vspace{4pt}
\raggedright
\endgroup
\end{table}

\FloatBarrier

\clearpage
\subsection{Flow Matching}
\label{app:flow_matching}

This subsection details the comparison in
Sec.~\ref{subsec:ablation_flow_matching} between KGO, using single-step
closed-form KoFM transport, and pretrained Sundial-base-128m. We sweep
Sundial's Euler integration budget over $N\in\{1,5,10,20,50,100\}$ to
examine its accuracy--latency trade-off relative to KGO's matrix-exponential
transport. This compares complete forecasting models; their architectures
and training procedures differ in addition to their sampling mechanisms.

\paragraph{Datasets and forecast accuracy.}
We evaluate seven LTSF datasets: ETTh1/2, ETTm1/2, Electricity, Weather,
and Exchange. Traffic and ILI are excluded because they appear in
Sundial's pretraining corpus. Both models receive a context of length~96
and generate 100 forecast trajectories for each prediction horizon
$H\in\{96,192,336,720\}$. CRPS and NMAE are computed on real test windows
using the standard ProbTS Evaluator, with KGO results from seed~$0$. The
latency benchmark is run separately and does not recompute these accuracy
results.

\paragraph{Inference-time measurement.}
We measure latency in FP32 on a single NVIDIA RTX~6000 Ada GPU, using
batch size~1 and 100 forecast trajectories per window. We report median
latency after warm-up, with CUDA synchronization around each timed call.
Latency is reported in seconds per window, excluding model loading and
metric computation.

Table~\ref{tab:full:flow-matching-kofm-vs-sundial} reports CRPS and inference
latency for all 28 dataset--horizon pairs.

\begin{table}[!htbp]
\centering
\begingroup
\setlength{\tabcolsep}{2.2pt}
\renewcommand{\arraystretch}{1.35}
\caption{CRPS and inference latency (s/window) for KGO (KoFM) and Sundial (FM $N$, where $N$ is the number of Euler steps); lower is better. See Sec.~\ref{app:flow_matching} for the evaluation and timing protocol. \best{Red bold}: best; \second{blue underline}: second best.}
\label{tab:full:flow-matching-kofm-vs-sundial}

\begin{adjustbox}{max width=\linewidth, max totalheight=0.92\textheight}
\begin{tabular}{ll cccccccccccccc}
\toprule

\multicolumn{2}{c}{} &
\multicolumn{2}{c}{KoFM} &
\multicolumn{2}{c}{FM 1} &
\multicolumn{2}{c}{FM 5} &
\multicolumn{2}{c}{FM 10} &
\multicolumn{2}{c}{FM 20} &
\multicolumn{2}{c}{FM 50} &
\multicolumn{2}{c}{FM 100} \\
\midrule

Dataset & Horizon &
CRPS & Time &
CRPS & Time &
CRPS & Time &
CRPS & Time &
CRPS & Time &
CRPS & Time &
CRPS & Time \\
\midrule

\multirow{4}{*}{\rotatebox{90}{ETTm1-L}} & 96 & \best{0.2145} & \best{0.0030} & 0.3027 & \second{0.0124} & 0.2677 & 0.0173 & 0.2586 & 0.0229 & 0.2546 & 0.0343 & 0.2524 & 0.0696 & \second{0.2517} & 0.1269 \\
 & 192 & \best{0.2363} & \best{0.0030} & 0.3336 & \second{0.0127} & 0.2954 & 0.0174 & 0.2853 & 0.0235 & 0.2809 & 0.0353 & 0.2784 & 0.0704 & \second{0.2777} & 0.1309 \\
 & 336 & \best{0.2491} & \best{0.0030} & 0.3480 & \second{0.0133} & 0.3084 & 0.0186 & 0.2979 & 0.0245 & 0.2934 & 0.0353 & 0.2909 & 0.0697 & \second{0.2902} & 0.1283 \\
 & 720 & \best{0.2747} & \best{0.0030} & 0.3754 & \second{0.0140} & 0.3360 & 0.0185 & 0.3252 & 0.0248 & 0.3205 & 0.0361 & 0.3180 & 0.0733 & \second{0.3173} & 0.1315 \\
\midrule
\multirow{4}{*}{\rotatebox{90}{ETTm2-L}} & 96 & \best{0.1064} & \best{0.0054} & 0.1511 & \second{0.0109} & 0.1359 & 0.0157 & 0.1316 & 0.0196 & 0.1295 & 0.0335 & 0.1282 & 0.0668 & \second{0.1277} & 0.1167 \\
 & 192 & \best{0.1309} & \best{0.0054} & 0.1741 & \second{0.0112} & 0.1584 & 0.0149 & 0.1538 & 0.0210 & 0.1515 & 0.0315 & 0.1501 & 0.0658 & \second{0.1496} & 0.1111 \\
 & 336 & \best{0.1450} & \best{0.0054} & 0.1920 & \second{0.0112} & 0.1759 & 0.0154 & 0.1708 & 0.0230 & 0.1683 & 0.0342 & 0.1667 & 0.0743 & \second{0.1662} & 0.1119 \\
 & 720 & \best{0.1711} & \best{0.0054} & 0.2145 & \second{0.0118} & 0.1982 & 0.0152 & 0.1930 & 0.0231 & 0.1904 & 0.0315 & 0.1887 & 0.0628 & \second{0.1882} & 0.1174 \\
\midrule
\multirow{4}{*}{\rotatebox{90}{ETTh1-L}} & 96 & \best{0.2405} & \best{0.0031} & 0.3031 & \second{0.0105} & 0.2707 & 0.0148 & 0.2653 & 0.0197 & 0.2625 & 0.0300 & 0.2609 & 0.0599 & \second{0.2603} & 0.1082 \\
 & 192 & \best{0.2649} & \best{0.0031} & 0.3414 & \second{0.0107} & 0.3084 & 0.0145 & 0.3026 & 0.0195 & 0.2995 & 0.0294 & 0.2977 & 0.0608 & \second{0.2970} & 0.1080 \\
 & 336 & \best{0.2856} & \best{0.0031} & 0.3589 & \second{0.0110} & 0.3250 & 0.0149 & 0.3189 & 0.0205 & 0.3157 & 0.0294 & 0.3137 & 0.0607 & \second{0.3130} & 0.1084 \\
 & 720 & \best{0.2841} & \best{0.0031} & 0.3608 & \second{0.0114} & 0.3240 & 0.0151 & 0.3177 & 0.0205 & 0.3146 & 0.0314 & 0.3126 & 0.0610 & \second{0.3119} & 0.1113 \\
\midrule
\multirow{4}{*}{\rotatebox{90}{ETTh2-L}} & 96 & \best{0.1338} & \best{0.0032} & 0.1836 & \second{0.0124} & 0.1652 & 0.0176 & 0.1608 & 0.0231 & 0.1585 & 0.0354 & 0.1571 & 0.0741 & \second{0.1566} & 0.1300 \\
 & 192 & \best{0.1549} & \best{0.0031} & 0.2094 & \second{0.0125} & 0.1912 & 0.0173 & 0.1867 & 0.0232 & 0.1843 & 0.0352 & 0.1827 & 0.0710 & \second{0.1822} & 0.1368 \\
 & 336 & \best{0.1721} & \best{0.0032} & 0.2295 & \second{0.0126} & 0.2115 & 0.0175 & 0.2070 & 0.0236 & 0.2045 & 0.0349 & 0.2029 & 0.0710 & \second{0.2024} & 0.1320 \\
 & 720 & \best{0.1764} & \best{0.0032} & 0.2299 & \second{0.0133} & 0.2113 & 0.0188 & 0.2067 & 0.0245 & 0.2041 & 0.0359 & 0.2025 & 0.0711 & \second{0.2019} & 0.1316 \\
\midrule
\multirow{4}{*}{\rotatebox{90}{Elec.-L}} & 96 & \best{0.0666} & \best{0.0166} & 0.0811 & \second{0.1242} & 0.0708 & 0.2728 & 0.0693 & 0.4767 & 0.0686 & 0.8964 & 0.0682 & 2.2059 & \second{0.0681} & 4.3803 \\
 & 192 & \best{0.0736} & \best{0.0168} & 0.0882 & \second{0.1288} & 0.0775 & 0.2794 & 0.0759 & 0.4887 & 0.0751 & 0.9044 & 0.0747 & 2.2060 & \second{0.0745} & 4.3846 \\
 & 336 & \best{0.0814} & \best{0.0225} & 0.0976 & \second{0.1427} & 0.0861 & 0.2874 & 0.0843 & 0.4912 & 0.0834 & 0.9066 & 0.0829 & 2.2133 & \second{0.0827} & 4.3850 \\
 & 720 & 0.0982 & \best{0.0494} & 0.1150 & \second{0.2614} & 0.1021 & 0.3969 & 0.0998 & 0.5996 & 0.0987 & 0.9124 & \second{0.0980} & 2.2090 & \best{0.0978} & 4.3862 \\
\midrule
\multirow{4}{*}{\rotatebox{90}{Exch.-L}} & 96 & \best{0.0193} & \best{0.0065} & 0.0272 & \second{0.0129} & 0.0256 & 0.0175 & 0.0250 & 0.0225 & 0.0246 & 0.0340 & 0.0244 & 0.0677 & \second{0.0243} & 0.1244 \\
 & 192 & \best{0.0294} & \best{0.0065} & 0.0374 & \second{0.0129} & 0.0357 & 0.0175 & 0.0350 & 0.0233 & 0.0346 & 0.0346 & 0.0343 & 0.0708 & \second{0.0342} & 0.1256 \\
 & 336 & \best{0.0407} & \best{0.0064} & 0.0494 & \second{0.0134} & 0.0477 & 0.0178 & 0.0469 & 0.0238 & 0.0465 & 0.0350 & 0.0462 & 0.0684 & \second{0.0461} & 0.1267 \\
 & 720 & \best{0.0689} & \best{0.0065} & 0.0792 & \second{0.0153} & 0.0776 & 0.0189 & 0.0769 & 0.0245 & 0.0765 & 0.0361 & 0.0762 & 0.0698 & \second{0.0761} & 0.1281 \\
\midrule
\multirow{4}{*}{\rotatebox{90}{Weather-L}} & 96 & \best{0.0527} & \best{0.0067} & 0.0975 & \second{0.0174} & 0.0837 & 0.0258 & 0.0800 & 0.0399 & 0.0785 & 0.0663 & 0.0778 & 0.1531 & \second{0.0776} & 0.3085 \\
 & 192 & \best{0.0583} & \best{0.0068} & 0.1017 & \second{0.0177} & 0.0892 & 0.0266 & 0.0857 & 0.0411 & 0.0842 & 0.0690 & 0.0833 & 0.1589 & \second{0.0831} & 0.3138 \\
 & 336 & \best{0.0620} & \best{0.0068} & 0.1103 & \second{0.0181} & 0.0970 & 0.0289 & 0.0931 & 0.0399 & 0.0913 & 0.0687 & 0.0902 & 0.1601 & \second{0.0899} & 0.3146 \\
 & 720 & \best{0.0682} & \best{0.0070} & 0.1175 & \second{0.0197} & 0.1029 & 0.0280 & 0.0988 & 0.0400 & 0.0970 & 0.0700 & 0.0959 & 0.1593 & \second{0.0956} & 0.3144 \\

\bottomrule
\end{tabular}
\end{adjustbox}
\endgroup
\end{table}

\FloatBarrier

\clearpage
\subsection{Foundation Model Comparison}
\label{app:foundation_models}

We compare KGO against three zero-shot time-series foundation models on the
standard long-term probabilistic benchmark. KGO is trained from scratch per
dataset; the foundation models ($35$M--$128$M parameters) are pre-trained on
much larger corpora and used off the shelf.

\paragraph{Baselines.}
We evaluate three publicly released zero-shot models off the shelf, with no
fine-tuning on the target datasets:
\begin{itemize}
    \item \textbf{Sundial}~(\texttt{thuml/sundial-base-128m}, $128$M params) --
    univariate; the wrapper loops over the variable axis.
    \item \textbf{TiRex}~(\texttt{NX-AI/TiRex}, $35$M params) -- per-series; the
    wrapper flattens the variable axis into the batch and decodes each series
    independently.
    \item \textbf{Chronos-2}~(\texttt{amazon/chronos-2}, $120$M params) --
    natively multivariate via long-format input with covariates.
\end{itemize}
All three are run with their published checkpoints and no fine-tuning.

\paragraph{Protocol and context length.}
Every CRPS / NMAE number in the table, including the foundation-model rows, is computed by the same ProbTS \texttt{Evaluator}.
Context length is fixed to $96$ for every model so the comparison sits inside the
standard long-term forecasting protocol used elsewhere in the paper.
The foundation models natively support much longer contexts ($2$k--$8$k), so the table reflects
their behavior at the benchmark's input length, not their best zero-shot
setting.

\paragraph{Sampling and quantile handling.}
All three models are scored by the same ProbTS evaluator at the same
$19$ quantile levels in $[0.05, 0.95]$, so any gap between them reflects
model behavior rather than evaluation drift. Sundial's $100$ stochastic
samples are passed in directly; TiRex and Chronos-2 emit quantile grids
($9$ and $19$ levels respectively), which we transport to the evaluator's
sample-based interface by piecewise linear inverse-CDF interpolation.
This is lossless for Chronos-2 at every scored level. TiRex emits only
$[0.1, 0.9]$, so the evaluator's $q_{0.05}$ and $q_{0.95}$ queries fall
back to TiRex's nearest emitted quantile; of the other $17$ levels, $9$ lie
on TiRex's grid and $8$ are linearly interpolated between grid levels.
We keep all $19$ in order to make a fair comparison across models.

\paragraph{Datasets, horizons, and seed.}
We evaluate on ETTh1, ETTh2, ETTm1, ETTm2, Exchange, Weather, and (Sundial
only) Electricity at horizons $\{96, 192, 336, 720\}$ with a single seed
($0$); Traffic, ILI, and TiRex/Chronos-2 on Electricity are excluded as
pretraining leakage.

\begin{table}[htbp]
\centering
\begingroup
\setlength{\tabcolsep}{3pt}
\renewcommand{\arraystretch}{1.0}
\caption{Multivariate probabilistic time series forecasting results. \best{Red bold}: best; \second{blue underline}: second best. ``--'' = unavailable or omitted because the foundation model was pretrained on that dataset (data-leakage avoidance).}
\label{tab:probabilistic_results}

\begin{adjustbox}{max width=\linewidth, max totalheight=0.88\textheight}
\begin{tabular}{ll cccccccc}
\toprule

\multicolumn{2}{c}{} &
\multicolumn{2}{c}{KGO} &
\multicolumn{2}{c}{Sundial} &
\multicolumn{2}{c}{TiRex} &
\multicolumn{2}{c}{Chronos-2} \\
\midrule

Dataset & Horizon &
CRPS & NMAE &
CRPS & NMAE &
CRPS & NMAE &
CRPS & NMAE \\
\midrule
\multirow{4}{*}{\rotatebox{90}{ETTm1-L}}
& 96 & \best{0.2145} & \best{0.2690} & \second{0.2532} & \second{0.3080} & 0.2876 & 0.3467 & 0.2726 & 0.3449 \\
& 192 & \best{0.2363} & \best{0.2964} & \second{0.2788} & \second{0.3368} & 0.2998 & 0.3630 & 0.2876 & 0.3641 \\
& 336 & \best{0.2491} & \best{0.3158} & \second{0.2913} & \second{0.3503} & 0.3115 & 0.3786 & 0.2925 & 0.3684 \\
& 720 & \best{0.2747} & \best{0.3469} & 0.3183 & \second{0.3799} & 0.3531 & 0.4375 & \second{0.3113} & 0.3873 \\
\midrule
\multirow{4}{*}{\rotatebox{90}{ETTm2-L}}
& 96 & \best{0.1064} & \best{0.1343} & 0.1284 & 0.1535 & 0.1264 & 0.1540 & \second{0.1147} & \second{0.1453} \\
& 192 & \best{0.1309} & \best{0.1638} & 0.1500 & 0.1769 & 0.1440 & 0.1784 & \second{0.1357} & \second{0.1690} \\
& 336 & \best{0.1450} & \best{0.1824} & 0.1667 & 0.1949 & 0.1569 & 0.1956 & \second{0.1558} & \second{0.1924} \\
& 720 & \best{0.1711} & \best{0.2142} & 0.1887 & 0.2174 & \second{0.1784} & 0.2215 & 0.1805 & \second{0.2172} \\
\midrule
\multirow{4}{*}{\rotatebox{90}{ETTh1-L}}
& 96 & \best{0.2405} & \best{0.3077} & 0.2609 & \second{0.3087} & 0.2477 & 0.3185 & \second{0.2421} & 0.3149 \\
& 192 & \best{0.2649} & \best{0.3396} & 0.2977 & \second{0.3470} & 0.2803 & 0.3609 & \second{0.2720} & 0.3546 \\
& 336 & \best{0.2856} & \best{0.3629} & 0.3137 & \second{0.3647} & 0.3041 & 0.3891 & \second{0.2891} & 0.3750 \\
& 720 & \best{0.2841} & \second{0.3694} & 0.3126 & \best{0.3667} & 0.3302 & 0.4331 & \second{0.2938} & 0.3788 \\
\midrule
\multirow{4}{*}{\rotatebox{90}{ETTh2-L}}
& 96 & \best{0.1338} & \best{0.1703} & 0.1571 & 0.1862 & 0.1427 & 0.1838 & \second{0.1411} & \second{0.1824} \\
& 192 & \best{0.1549} & \best{0.1968} & 0.1826 & 0.2126 & 0.1638 & 0.2096 & \second{0.1621} & \second{0.2093} \\
& 336 & \best{0.1721} & \best{0.2179} & 0.2027 & 0.2325 & 0.1819 & \second{0.2316} & \second{0.1815} & 0.2320 \\
& 720 & \best{0.1764} & \best{0.2265} & 0.2025 & \second{0.2326} & \second{0.1864} & 0.2383 & 0.1891 & 0.2397 \\
\midrule
\multirow{4}{*}{\rotatebox{90}{Elec.-L}}
& 96 & \best{0.0666} & \best{0.0783} & \second{0.0681} & \second{0.0831} & -- & -- & -- & -- \\
& 192 & \best{0.0736} & \best{0.0864} & \second{0.0745} & \second{0.0905} & -- & -- & -- & -- \\
& 336 & \best{0.0814} & \best{0.0961} & \second{0.0828} & \second{0.1001} & -- & -- & -- & -- \\
& 720 & \second{0.0982} & \best{0.1168} & \best{0.0978} & \second{0.1177} & -- & -- & -- & -- \\
\midrule
\multirow{4}{*}{\rotatebox{90}{Exch.-L}}
& 96 & \second{0.0193} & \second{0.0242} & 0.0243 & 0.0274 & 0.0195 & 0.0247 & \best{0.0184} & \best{0.0236} \\
& 192 & 0.0294 & 0.0367 & 0.0344 & 0.0377 & \second{0.0282} & \second{0.0348} & \best{0.0268} & \best{0.0338} \\
& 336 & 0.0407 & 0.0498 & 0.0462 & 0.0497 & \second{0.0392} & \second{0.0469} & \best{0.0372} & \best{0.0456} \\
& 720 & 0.0689 & 0.0821 & 0.0763 & 0.0794 & \second{0.0684} & \second{0.0761} & \best{0.0640} & \best{0.0751} \\
\midrule
\multirow{4}{*}{\rotatebox{90}{Weather-L}}
& 96 & \best{0.0527} & \best{0.0655} & \second{0.0783} & \second{0.0913} & 0.1176 & 0.1457 & 0.0977 & 0.1226 \\
& 192 & \best{0.0583} & \best{0.0725} & \second{0.0834} & \second{0.0973} & 0.1149 & 0.1386 & 0.0962 & 0.1170 \\
& 336 & \best{0.0620} & \best{0.0769} & 0.0902 & \second{0.1059} & 0.1040 & 0.1235 & \second{0.0895} & 0.1071 \\
& 720 & \best{0.0682} & \best{0.0839} & \second{0.0960} & 0.1127 & 0.1003 & 0.1168 & 0.0972 & \second{0.1125} \\
\bottomrule
\end{tabular}
\end{adjustbox}

\endgroup
\end{table}

\FloatBarrier

\clearpage
\subsection{Efficiency}
\label{app:ablation:efficiency}

\paragraph{Measurement protocol.}
We benchmark all methods on a single NVIDIA RTX~6000 Ada GPU in FP32
in eager mode (without compilation), using batch size~1 and generating
100 forecast trajectories per input
window. All latencies are measured after warm-up, with CUDA
synchronization around each timed forward pass; we report the mean
latency for baselines and the median for KGO. Peak GPU memory is the
maximum allocated memory, reported in GB. Each baseline uses its native
sampler; KGO reuses sample-independent conditioning computations and
generates trajectories in chunks of five on Electricity and Traffic to
limit peak memory.
\begin{table}[!htbp]
\centering
\begingroup
\setlength{\tabcolsep}{2pt}
\renewcommand{\arraystretch}{1.1}
\caption{Inference latency (s/window) and peak GPU memory (GB) across nine datasets; lower is better. See Sec.~\ref{app:ablation:efficiency} for the measurement protocol. \best{Red bold}: best; \second{blue underline}: second best.}
\label{tab:full:efficiency}

\begin{adjustbox}{max width=\linewidth, max totalheight=0.92\textheight}
\begin{tabular}{ll cc cc cc cc cc cc cc cc}
\toprule
\multicolumn{2}{c}{} &
\multicolumn{2}{c}{TSDiff} & \multicolumn{2}{c}{CSDI} & \multicolumn{2}{c}{TimeGrad} & \multicolumn{2}{c}{GRU MAF} & \multicolumn{2}{c}{Trans MAF} & \multicolumn{2}{c}{GRU NVP} & \multicolumn{2}{c}{$K^2$VAE} & \multicolumn{2}{c}{\textbf{KGO (Ours)}} \\
\midrule

Dataset & Horizon &
Time & Mem & Time & Mem & Time & Mem & Time & Mem & Time & Mem & Time & Mem & Time & Mem & \textbf{Time} & \textbf{Mem} \\
\midrule

\multirow{4}{*}{\rotatebox{90}{ETTm1-L}}& 96& 0.454& 0.092& 24.969& 0.018& 31.406& \second{0.015}& 0.685& 0.022& 0.734& 0.079& 0.222& \best{0.013}& \second{0.018}& \second{0.015}& \best{0.0030}& 0.019 \\
& 192& 0.454& 0.132& 24.920& 0.022& 63.597& \second{0.016}& 1.265& 0.022& 1.553& 0.080& 0.456& \best{0.014}& \second{0.022}& 0.017& \best{0.0030}& 0.028 \\
& 336& 0.451& 0.198& 25.193& 0.027& 111.064& \second{0.018}& 2.266& 0.024& 2.616& 0.082& 0.795& \best{0.016}& \second{0.027}& 0.021& \best{0.0030}& 0.039 \\
& 720& 0.484& 0.357& 24.062& 0.040& 254.576& \best{0.021}& 4.950& \second{0.029}& 5.804& 0.087& 1.776& \best{0.021}& \second{0.041}& 0.031& \best{0.0030}& 0.070 \\
\midrule
\multirow{4}{*}{\rotatebox{90}{ETTm2-L}}& 96& 0.462& 0.092& 25.103& 0.018& 34.132& \best{0.012}& 0.684& \second{0.014}& 0.731& 0.040& 0.153& \second{0.014}& \second{0.018}& 0.015& \best{0.0054}& 0.019 \\
& 192& 0.451& 0.132& 24.991& 0.022& 67.229& \best{0.013}& 1.403& \second{0.015}& 1.586& 0.041& 0.300& \second{0.015}& \second{0.022}& 0.017& \best{0.0054}& 0.028 \\
& 336& 0.484& 0.198& 25.976& 0.027& 119.362& \best{0.015}& 2.228& \second{0.017}& 2.828& 0.043& 0.495& \second{0.017}& \second{0.027}& 0.021& \best{0.0054}& 0.038 \\
& 720& 0.485& 0.357& 24.905& 0.040& 249.600& \best{0.020}& 5.065& \second{0.022}& 5.920& 0.048& 1.152& \second{0.022}& \second{0.044}& 0.031& \best{0.0054}& 0.069 \\
\midrule
\multirow{4}{*}{\rotatebox{90}{ETTh1-L}}& 96& 0.482& 0.092& 25.048& 0.018& 33.937& \second{0.015}& 0.635& \second{0.015}& 0.735& 0.041& 0.225& \best{0.013}& \second{0.018}& \second{0.015}& \best{0.0031}& 0.019 \\
& 192& 0.453& 0.132& 24.012& 0.022& 66.838& 0.017& 1.384& \second{0.016}& 1.600& 0.042& 0.481& \best{0.015}& \second{0.022}& 0.017& \best{0.0031}& 0.028 \\
& 336& 0.486& 0.198& 24.674& 0.027& 118.148& \second{0.018}& 2.363& \second{0.018}& 2.819& 0.043& 0.793& \best{0.016}& \second{0.027}& 0.021& \best{0.0031}& 0.039 \\
& 720& 0.483& 0.357& 25.638& 0.040& 256.945& \best{0.023}& 5.227& \best{0.023}& 5.835& 0.047& 1.741& \best{0.023}& \second{0.043}& \second{0.031}& \best{0.0031}& 0.070 \\
\midrule
\multirow{4}{*}{\rotatebox{90}{ETTh2-L}}& 96& 0.453& 0.092& 25.116& 0.018& 34.045& \best{0.013}& 0.692& 0.015& 0.728& 0.041& 0.160& \second{0.014}& \second{0.018}& 0.015& \best{0.0032}& 0.022 \\
& 192& 0.452& 0.132& 25.605& 0.022& 67.710& \best{0.014}& 1.278& 0.016& 1.465& 0.042& 0.321& \second{0.015}& \second{0.022}& 0.017& \best{0.0031}& 0.031 \\
& 336& 0.451& 0.198& 25.886& 0.027& 118.779& \best{0.016}& 2.429& 0.018& 2.723& 0.043& 0.565& \second{0.017}& \second{0.027}& 0.021& \best{0.0032}& 0.042 \\
& 720& 0.485& 0.357& 25.893& 0.040& 256.475& \best{0.022}& 5.154& \second{0.023}& 6.004& 0.047& 1.224& \second{0.023}& \second{0.041}& 0.031& \best{0.0032}& 0.073 \\
\midrule
\multirow{4}{*}{\rotatebox{90}{Elec.-L}}& 96& 0.489& 0.238& 96.645& 0.296& 33.504& 0.132& 25.746& 0.134& 26.389& 0.165& 0.125& 0.132& \second{0.019}& \best{0.067}& \best{0.0166}& \second{0.117} \\
& 192& 0.537& 0.352& 165.237& 0.435& 66.456& 0.182& 53.662& 0.184& 54.411& 0.215& 0.251& 0.182& \second{0.022}& \best{0.099}& \best{0.0168}& \second{0.176} \\
& 336& 0.643& 0.513& 286.608& 0.647& 119.140& \second{0.258}& 94.409& 0.261& 94.014& 0.292& 0.477& 0.259& \second{0.029}& \best{0.145}& \best{0.0225}& 0.271 \\
& 720& 0.986& 0.960& 712.445& 1.212& 248.970& \second{0.455}& 202.201& 0.458& 196.062& 0.491& 0.942& 0.456& \best{0.040}& \best{0.268}& \second{0.0494}& 0.523 \\
\midrule
\multirow{4}{*}{\rotatebox{90}{Exch.-L}}& 96& 0.443& 0.092& 24.805& 0.019& 33.105& \best{0.013}& 0.776& \second{0.014}& 0.805& 0.040& 0.134& 0.015& \second{0.018}& 0.015& \best{0.0065}& 0.020 \\
& 192& 0.445& 0.133& 25.398& 0.023& 66.895& \best{0.014}& 1.506& \second{0.015}& 1.721& 0.041& 0.269& 0.016& \second{0.021}& 0.017& \best{0.0065}& 0.025 \\
& 336& 0.474& 0.199& 25.458& 0.030& 116.148& \best{0.017}& 2.702& \second{0.018}& 3.015& 0.043& 0.473& \second{0.018}& \second{0.027}& 0.021& \best{0.0064}& 0.034 \\
& 720& 0.480& 0.358& 25.551& 0.043& 250.412& \best{0.021}& 5.723& \second{0.022}& 6.258& 0.048& 1.087& 0.023& \second{0.041}& 0.031& \best{0.0065}& 0.056 \\
\midrule
\multirow{4}{*}{\rotatebox{90}{Traffic-L}}& 96& 0.653& 0.619& 386.904& 0.772& 33.289& 0.336& 67.177& 0.336& 67.180& 0.342& 0.273& 0.343& \best{0.019}& \best{0.159}& \second{0.0225}& \second{0.223} \\
& 192& 0.824& 0.910& 650.234& 1.150& 66.347& 0.469& 134.161& 0.470& 133.921& 0.476& 0.551& 0.477& \best{0.022}& \best{0.241}& \second{0.0229}& \second{0.264} \\
& 336& 1.093& 1.359& 1057.751& 1.717& 114.615& 0.670& 234.524& 0.670& 234.948& 0.677& 1.055& 0.677& \best{0.027}& \best{0.361}& \second{0.0329}& \second{0.429} \\
& 720& 1.852& 2.558& 2371.636& 3.236& 250.309& 1.204& 507.890& 1.203& 508.772& 1.212& 2.300& 1.211& \best{0.042}& \best{0.688}& \second{0.0731}& \second{0.852} \\
\midrule
\multirow{4}{*}{\rotatebox{90}{Weather-L}}& 96& 0.479& 0.097& 25.588& 0.032& 33.777& \best{0.016}& 1.741& 0.025& 1.960& 0.081& 0.226& 0.018& \second{0.018}& \second{0.017}& \best{0.0067}& 0.037 \\
& 192& 0.446& 0.139& 24.682& 0.041& 67.780& \best{0.020}& 3.710& 0.027& 3.935& 0.083& 0.456& \second{0.022}& \second{0.023}& \best{0.020}& \best{0.0068}& 0.052 \\
& 336& 0.446& 0.208& 24.927& 0.056& 117.750& \best{0.025}& 6.603& 0.031& 6.934& 0.086& 0.829& \second{0.028}& \second{0.027}& \best{0.025}& \best{0.0068}& 0.072 \\
& 720& 0.488& 0.376& 36.440& 0.096& 254.377& \second{0.039}& 13.993& 0.046& 13.877& 0.096& 1.713& 0.041& \second{0.041}& \best{0.037}& \best{0.0070}& 0.126 \\
\midrule
\multirow{4}{*}{\rotatebox{90}{ILI-L}}& 24& 0.447& 0.036& 25.964& \second{0.013}& 7.779& \best{0.011}& 0.156& 0.014& 0.180& 0.024& 0.050& 0.014& \second{0.018}& 0.017& \best{0.0043}& 0.016 \\
& 36& 0.441& 0.041& 24.946& \second{0.014}& 12.765& \best{0.011}& 0.234& \second{0.014}& 0.276& 0.024& 0.076& \second{0.014}& \second{0.020}& 0.018& \best{0.0043}& 0.018 \\
& 48& 0.467& 0.047& 25.527& \second{0.014}& 16.660& \best{0.011}& 0.314& \second{0.014}& 0.362& 0.024& 0.100& \second{0.014}& \second{0.022}& 0.018& \best{0.0043}& 0.020 \\
& 60& 0.443& 0.051& 24.585& 0.015& 21.114& \best{0.012}& 0.431& \second{0.014}& 0.470& 0.024& 0.126& \second{0.014}& \second{0.023}& 0.020& \best{0.0043}& 0.022 \\

\bottomrule
\end{tabular}
\end{adjustbox}

\endgroup
\end{table}

\FloatBarrier

\clearpage

\section{Ablation Studies}
\label{app:ablation_studies}
\subsection{Patching (KoPE)}
\label{app:patching_ablation}

In this ablation we examine the KoPE component. To check its importance, we compare this component
against alternative techniques such as using the last patch representation and pooling the full context.
To isolate the representation alone, we keep all other components the same.
We evaluate on a grid of $36$ dataset--horizon settings using seed $0$,
reporting CRPS and NMAE with the standard ProbTS evaluator.
\begin{table}[!htbp]
\centering
\caption{Patch representation ablation. Multivariate probabilistic forecasting results. \best{Red bold}: best; \second{blue underline}: second best.}
\label{tab:full:context}

\begin{adjustbox}{width=\linewidth, max totalheight=0.65\textheight}
\begin{tabular}{ll cccccc}
\toprule

\multicolumn{2}{c}{} &
\multicolumn{2}{c}{KoPE} &
\multicolumn{2}{c}{Last Patch} &
\multicolumn{2}{c}{Context Pooling} \\
\midrule

Dataset & Horizon &
CRPS & NMAE &
CRPS & NMAE &
CRPS & NMAE \\
\midrule

\multirow{4}{*}{\rotatebox{90}{ETTm1-L}}
& 96  & \best{0.2145} & \best{0.2690} & 0.3127 & \second{0.3925} & \second{0.3119} & 0.3938 \\
& 192 & \best{0.2363} & \best{0.2964} & 0.3263 & 0.4107 & \second{0.3255} & \second{0.4093} \\
& 336 & \best{0.2491} & \best{0.3158} & 0.3371 & 0.4236 & \second{0.3351} & \second{0.4173} \\
& 720 & \best{0.2747} & \best{0.3469} & 0.3563 & 0.4412 & \second{0.3558} & \second{0.4400} \\
\midrule

\multirow{4}{*}{\rotatebox{90}{ETTm2-L}}
& 96  & \best{0.1064} & \best{0.1343} & \second{0.1242} & \second{0.1602} & 0.1250 & 0.1611 \\
& 192 & \best{0.1309} & \best{0.1638} & 0.1454 & 0.1849 & \second{0.1438} & \second{0.1836} \\
& 336 & \best{0.1450} & \best{0.1824} & 0.1565 & 0.1990 & \second{0.1554} & \second{0.1972} \\
& 720 & \best{0.1711} & \best{0.2142} & 0.1783 & \second{0.2234} & \second{0.1780} & 0.2239 \\
\midrule

\multirow{4}{*}{\rotatebox{90}{ETTh1-L}}
& 96  & \best{0.2405} & \best{0.3077} & \second{0.2413} & \second{0.3105} & 0.2442 & 0.3167 \\
& 192 & \best{0.2649} & \second{0.3396} & \second{0.2652} & \best{0.3389} & 0.2712 & 0.3478 \\
& 336 & \second{0.2856} & \second{0.3629} & \best{0.2842} & \best{0.3595} & 0.2932 & 0.3706 \\
& 720 & \best{0.2841} & \best{0.3694} & 0.2992 & 0.3887 & \second{0.2959} & \second{0.3884} \\
\midrule

\multirow{4}{*}{\rotatebox{90}{ETTh2-L}}
& 96  & \best{0.1338} & \best{0.1703} & \second{0.1369} & \second{0.1743} & 0.1376 & 0.1752 \\
& 192 & 0.1549 & 0.1968 & \second{0.1538} & \second{0.1947} & \best{0.1529} & \best{0.1932} \\
& 336 & \best{0.1721} & \best{0.2179} & \second{0.1724} & \second{0.2184} & 0.1748 & 0.2240 \\
& 720 & \second{0.1764} & \second{0.2265} & \best{0.1740} & \best{0.2228} & 0.1771 & 0.2280 \\
\midrule

\multirow{4}{*}{\rotatebox{90}{Elec.-L}}
& 96  & \best{0.0666} & \best{0.0783} & 0.0687 & 0.0817 & \second{0.0681} & \second{0.0816} \\
& 192 & \best{0.0736} & \best{0.0864} & 0.0764 & 0.0899 & \second{0.0758} & \second{0.0895} \\
& 336 & \best{0.0814} & \best{0.0961} & 0.0840 & 0.0992 & \second{0.0830} & \second{0.0991} \\
& 720 & \best{0.0982} & 0.1168 & \second{0.0989} & \best{0.1165} & 0.1013 & \second{0.1167} \\
\midrule

\multirow{4}{*}{\rotatebox{90}{Exch.-L}}
& 96  & \second{0.0193} & 0.0242 & \best{0.0191} & \second{0.0238} & \best{0.0191} & \best{0.0237} \\
& 192 & 0.0294 & 0.0367 & \best{0.0286} & \best{0.0350} & \second{0.0292} & \second{0.0359} \\
& 336 & \second{0.0407} & \second{0.0498} & \best{0.0393} & \best{0.0482} & 0.0410 & \second{0.0498} \\
& 720 & 0.0689 & \best{0.0821} & \best{0.0680} & \second{0.0822} & \second{0.0683} & \best{0.0821} \\
\midrule

\multirow{4}{*}{\rotatebox{90}{Traffic-L}}
& 96  & \best{0.1732} & \best{0.1988} & 0.2062 & 0.2464 & \second{0.2050} & \second{0.2438} \\
& 192 & \best{0.1777} & \best{0.2048} & 0.2283 & 0.2797 & \second{0.2268} & \second{0.2759} \\
& 336 & \best{0.1921} & \best{0.2217} & 0.2350 & 0.2860 & \second{0.2318} & \second{0.2811} \\
& 720 & \best{0.2274} & \best{0.2671} & 0.2490 & 0.3025 & \second{0.2461} & \second{0.3003} \\
\midrule

\multirow{4}{*}{\rotatebox{90}{Weather-L}}
& 96  & \best{0.0527} & \best{0.0655} & \second{0.0679} & \second{0.0828} & 0.0680 & 0.0834 \\
& 192 & \best{0.0583} & \best{0.0725} & \second{0.0791} & \second{0.0966} & 0.0795 & 0.0975 \\
& 336 & \best{0.0620} & \best{0.0769} & \second{0.0830} & 0.1010 & 0.0831 & \second{0.1008} \\
& 720 & \best{0.0682} & \best{0.0839} & 0.0881 & 0.1073 & \second{0.0878} & \second{0.1054} \\
\midrule

\multirow{4}{*}{\rotatebox{90}{ILI-L}}
& 24  & \best{0.0531} & \best{0.0750} & 0.1252 & 0.1609 & \second{0.1222} & \second{0.1600} \\
& 36  & \best{0.0821} & \best{0.1024} & 0.1492 & 0.1933 & \second{0.1490} & \second{0.1925} \\
& 48  & \best{0.0991} & \best{0.1191} & 0.1453 & 0.2012 & \second{0.1422} & \second{0.1978} \\
& 60  & \best{0.1206} & \best{0.1423} & 0.1435 & 0.1919 & \second{0.1336} & \second{0.1790} \\

\bottomrule
\end{tabular}
\end{adjustbox}

\end{table}

\clearpage
\subsection{Sequence Encoder}
\label{app:sequence_encoder}
In this subsection, we investigate the Sequence Encoder that produces the patch representation.
This representation is propagated through KoPE; beyond memory and time efficiency,
it is important to model the temporal dynamics well. We therefore compare three cornerstone
sequential encoders: Transformer, Linear RNN, and sLSTM. To isolate the representation alone,
we keep all other components the same.
We evaluate on a grid of $36$ dataset--horizon settings using seed $0$,
reporting CRPS and NMAE with the standard ProbTS evaluator.
\begin{table}[!htbp]
\centering
\caption{Sequence encoder ablation. KGO with different sequential encoders (sLSTM is the KGO baseline). \best{Red bold}: best; \second{blue underline}: second best.}
\label{tab:full:seq-encoder}

\begin{adjustbox}{width=\linewidth, max totalheight=0.65\textheight}
\begin{tabular}{ll cccccc}
\toprule

\multicolumn{2}{c}{} &
\multicolumn{2}{c}{sLSTM} &
\multicolumn{2}{c}{Transformer} &
\multicolumn{2}{c}{Linear RNN} \\
\midrule

Dataset & Horizon &
CRPS & NMAE &
CRPS & NMAE &
CRPS & NMAE \\
\midrule

\multirow{4}{*}{\rotatebox{90}{ETTm1-L}}
& 96  & \best{0.214} & \best{0.269} & 0.237 & 0.297 & \second{0.220} & \second{0.275} \\
& 192 & \best{0.236} & \best{0.296} & 0.265 & 0.333 & \second{0.238} & \second{0.299} \\
& 336 & \best{0.249} & \best{0.316} & 0.276 & 0.349 & \second{0.253} & \second{0.319} \\
& 720 & \best{0.275} & \best{0.347} & 0.301 & 0.381 & \second{0.278} & \second{0.349} \\
\midrule
\multirow{4}{*}{\rotatebox{90}{ETTm2-L}}
& 96  & \second{0.106} & \second{0.134} & \second{0.106} & \second{0.134} & \best{0.105} & \best{0.132} \\
& 192 & 0.131 & 0.164 & \second{0.129} & \second{0.161} & \best{0.127} & \best{0.158} \\
& 336 & 0.145 & 0.182 & \second{0.144} & \second{0.181} & \best{0.142} & \best{0.177} \\
& 720 & 0.171 & 0.214 & \second{0.168} & \second{0.210} & \best{0.166} & \best{0.207} \\
\midrule
\multirow{4}{*}{\rotatebox{90}{ETTh1-L}}
& 96  & \second{0.240} & \second{0.308} & \best{0.238} & \best{0.307} & 0.241 & 0.310 \\
& 192 & \best{0.265} & \best{0.340} & \second{0.267} & \second{0.341} & \second{0.267} & \second{0.341} \\
& 336 & \second{0.286} & \second{0.363} & 0.288 & 0.366 & \best{0.281} & \best{0.358} \\
& 720 & \best{0.284} & \best{0.369} & 0.297 & 0.382 & \second{0.287} & \second{0.372} \\
\midrule
\multirow{4}{*}{\rotatebox{90}{ETTh2-L}}
& 96  & \best{0.134} & \best{0.170} & \best{0.134} & \second{0.171} & \second{0.135} & 0.172 \\
& 192 & \second{0.155} & \second{0.197} & \best{0.154} & \best{0.196} & \second{0.155} & 0.198 \\
& 336 & \second{0.172} & \second{0.218} & \best{0.171} & \best{0.217} & 0.174 & 0.221 \\
& 720 & \second{0.176} & \second{0.226} & \best{0.175} & \best{0.225} & 0.177 & 0.227 \\
\midrule
\multirow{4}{*}{\rotatebox{90}{Elec.-L}}
& 96  & \best{0.067} & \best{0.078} & \best{0.067} & \second{0.079} & \best{0.067} & 0.080 \\
& 192 & \best{0.074} & \best{0.086} & \best{0.074} & \second{0.087} & \second{0.075} & 0.088 \\
& 336 & \best{0.081} & \best{0.096} & \second{0.082} & \second{0.097} & \second{0.082} & \second{0.097} \\
& 720 & \second{0.098} & \second{0.117} & \best{0.097} & \best{0.116} & 0.099 & 0.118 \\
\midrule
\multirow{4}{*}{\rotatebox{90}{Exch.-L}}
& 96  & \best{0.019} & \best{0.024} & \best{0.019} & \best{0.024} & \second{0.020} & \best{0.024} \\
& 192 & \second{0.029} & 0.037 & \best{0.027} & \best{0.034} & \second{0.029} & \second{0.036} \\
& 336 & 0.041 & \second{0.050} & \second{0.040} & \second{0.050} & \best{0.039} & \best{0.047} \\
& 720 & 0.069 & 0.082 & \second{0.065} & \best{0.078} & \best{0.064} & \second{0.079} \\
\midrule
\multirow{4}{*}{\rotatebox{90}{Traffic-L}}
& 96  & \best{0.173} & \second{0.199} & \best{0.173} & \best{0.197} & \second{0.178} & 0.205 \\
& 192 & \best{0.178} & \second{0.205} & \best{0.178} & \best{0.204} & \second{0.182} & 0.210 \\
& 336 & \best{0.192} & \best{0.222} & 0.200 & 0.231 & \second{0.199} & \second{0.230} \\
& 720 & \best{0.227} & \best{0.267} & 0.313 & 0.366 & \second{0.262} & \second{0.307} \\
\midrule
\multirow{4}{*}{\rotatebox{90}{Weather-L}}
& 96  & \best{0.053} & \best{0.066} & \second{0.054} & \second{0.068} & \best{0.053} & \best{0.066} \\
& 192 & \best{0.058} & \best{0.073} & \second{0.060} & \second{0.075} & \best{0.058} & \best{0.073} \\
& 336 & \best{0.062} & \best{0.077} & \second{0.063} & \second{0.078} & \best{0.062} & \second{0.078} \\
& 720 & \best{0.068} & \best{0.084} & \second{0.069} & \second{0.085} & \second{0.069} & \second{0.085} \\
\midrule
\multirow{4}{*}{\rotatebox{90}{ILI-L}}
& 24  & \best{0.053} & \best{0.075} & \second{0.072} & \second{0.093} & 0.099 & 0.124 \\
& 36  & \second{0.082} & \second{0.102} & \best{0.075} & \best{0.098} & 0.100 & 0.125 \\
& 48  & \best{0.099} & \best{0.119} & 0.118 & 0.143 & \second{0.117} & \second{0.139} \\
& 60  & \best{0.121} & \best{0.142} & \second{0.129} & \second{0.151} & 0.132 & 0.153 \\
\bottomrule
\end{tabular}
\end{adjustbox}

\end{table}

\clearpage
\subsection{Decoder \& Gating}
\label{app:decoder_gating}
In this ablation we examine the coupling between the deterministic and probabilistic decoders.
The Adaptive Uncertainty Gate (AUG) modulates their contributions per (variate, horizon)
based on the predictability of the signal, allocating uncertainty dynamically rather than with a fixed split.
To check the importance of this learned gating, we compare AUG against three alternatives:
removing the deterministic stream entirely, conditioning KoFM on the deterministic prediction,
and a fixed residual split. To isolate the coupling rule alone, we keep all other components the same.
We evaluate on a grid of $36$ dataset--horizon settings using seed $0$,
reporting CRPS and NMAE with the standard ProbTS evaluator.
\begin{table}[!htbp]
\centering
\caption{Decoder and gating ablation. KGO is compared with a stochastic-only decoder (Prob Only), KoFM conditioned on the deterministic prediction (Condition), and fixed residual addition (Residual). Lower CRPS/NMAE is better. \best{Red bold}: best; \second{blue underline}: second best.}
\label{tab:full:det-decoder}

\begin{adjustbox}{width=\linewidth, max totalheight=0.65\textheight}
\begin{tabular}{ll cccccccc}
\toprule

\multicolumn{2}{c}{} &
\multicolumn{2}{c}{KGO} &
\multicolumn{2}{c}{Prob Only} &
\multicolumn{2}{c}{Condition} &
\multicolumn{2}{c}{Residual} \\
\midrule

Dataset & Horizon &
CRPS & NMAE &
CRPS & NMAE &
CRPS & NMAE &
CRPS & NMAE \\
\midrule

\multirow{4}{*}{\rotatebox{90}{ETTm1-L}}
& 96  & \second{0.2145} & 0.2690 & 0.2150 & 0.2695 & \best{0.2129} & \second{0.2675} & 0.2158 & \best{0.2642} \\
& 192 & 0.2363 & 0.2964 & \best{0.2342} & 0.2955 & \second{0.2349} & \second{0.2945} & 0.2382 & \best{0.2922} \\
& 336 & \best{0.2491} & \best{0.3158} & \best{0.2491} & \second{0.3159} & \second{0.2511} & 0.3170 & 0.2568 & 0.3166 \\
& 720 & \best{0.2747} & \best{0.3469} & 0.2766 & 0.3490 & \second{0.2765} & 0.3493 & 0.2831 & \second{0.3481} \\
\midrule

\multirow{4}{*}{\rotatebox{90}{ETTm2-L}}
& 96  & 0.1064 & 0.1343 & \best{0.1047} & \second{0.1321} & 0.1060 & 0.1333 & \second{0.1058} & \best{0.1320} \\
& 192 & 0.1309 & 0.1638 & 0.1306 & 0.1637 & \best{0.1292} & \second{0.1615} & \second{0.1298} & \best{0.1610} \\
& 336 & 0.1450 & 0.1824 & \second{0.1449} & \second{0.1823} & 0.1520 & 0.1902 & \best{0.1440} & \best{0.1789} \\
& 720 & 0.1711 & 0.2142 & \best{0.1679} & \second{0.2092} & 0.1691 & 0.2106 & \second{0.1685} & \best{0.2082} \\
\midrule

\multirow{4}{*}{\rotatebox{90}{ETTh1-L}}
& 96  & \best{0.2405} & \second{0.3077} & \second{0.2434} & 0.3112 & 0.2460 & 0.3139 & 0.2454 & \best{0.3069} \\
& 192 & \best{0.2649} & \best{0.3396} & \second{0.2654} & \second{0.3410} & 0.2682 & 0.3414 & 0.2784 & 0.3447 \\
& 336 & 0.2856 & 0.3629 & \best{0.2821} & \second{0.3596} & 0.2857 & 0.3657 & \second{0.2849} & \best{0.3572} \\
& 720 & \best{0.2841} & \best{0.3694} & 0.2910 & 0.3781 & \second{0.2870} & \second{0.3744} & 0.2917 & \best{0.3694} \\
\midrule

\multirow{4}{*}{\rotatebox{90}{ETTh2-L}}
& 96  & \second{0.1338} & \best{0.1703} & 0.1345 & 0.1716 & 0.1373 & 0.1746 & \best{0.1334} & \second{0.1713} \\
& 192 & \second{0.1549} & 0.1968 & 0.1552 & 0.1970 & \best{0.1539} & \second{0.1961} & 0.1563 & \best{0.1940} \\
& 336 & \best{0.1721} & \best{0.2179} & 0.1765 & 0.2250 & \second{0.1741} & 0.2210 & 0.1745 & \second{0.2190} \\
& 720 & \best{0.1764} & \best{0.2265} & 0.1804 & \second{0.2326} & \second{0.1802} & 0.2344 & 0.1819 & 0.2343 \\
\midrule

\multirow{4}{*}{\rotatebox{90}{Elec.-L}}
& 96  & 0.0666 & \best{0.0783} & \second{0.0664} & 0.0786 & 0.0670 & 0.0787 & \best{0.0658} & \second{0.0784} \\
& 192 & \second{0.0736} & \second{0.0864} & 0.0739 & 0.0865 & 0.0741 & 0.0869 & \best{0.0733} & \best{0.0862} \\
& 336 & \second{0.0814} & \best{0.0961} & 0.0816 & 0.0969 & 0.0825 & 0.0968 & \best{0.0806} & \second{0.0962} \\
& 720 & \second{0.0982} & \second{0.1168} & 0.0998 & 0.1192 & 0.1025 & 0.1221 & \best{0.0963} & \best{0.1153} \\
\midrule

\multirow{4}{*}{\rotatebox{90}{Exch.-L}}
& 96  & \second{0.0193} & \second{0.0242} & 0.0194 & \best{0.0241} & 0.0201 & 0.0250 & \best{0.0190} & \second{0.0242} \\
& 192 & 0.0294 & 0.0367 & 0.0290 & \second{0.0352} & \second{0.0285} & 0.0360 & \best{0.0271} & \best{0.0344} \\
& 336 & 0.0407 & \second{0.0498} & \second{0.0406} & 0.0514 & \best{0.0378} & \best{0.0471} & 0.0407 & 0.0515 \\
& 720 & 0.0689 & 0.0821 & \second{0.0658} & \second{0.0797} & 0.0676 & 0.0824 & \best{0.0617} & \best{0.0778} \\
\midrule

\multirow{4}{*}{\rotatebox{90}{Traffic-L}}
& 96  & 0.1732 & 0.1988 & 0.1721 & 0.1979 & \best{0.1717} & \second{0.1970} & \second{0.1719} & \best{0.1957} \\
& 192 & \second{0.1777} & 0.2048 & 0.1782 & 0.2044 & 0.1790 & \second{0.2042} & \best{0.1768} & \best{0.2006} \\
& 336 & \second{0.1921} & \second{0.2217} & 0.1934 & 0.2225 & 0.1964 & 0.2235 & \best{0.1898} & \best{0.2160} \\
& 720 & \second{0.2274} & \second{0.2671} & 0.2302 & 0.2694 & 0.2617 & 0.3039 & \best{0.2202} & \best{0.2515} \\
\midrule

\multirow{4}{*}{\rotatebox{90}{Weather-L}}
& 96  & \second{0.0527} & 0.0655 & 0.0533 & 0.0661 & \best{0.0518} & \best{0.0641} & 0.0531 & \second{0.0643} \\
& 192 & 0.0583 & 0.0725 & \second{0.0582} & \second{0.0724} & \best{0.0579} & \best{0.0720} & 0.0592 & \best{0.0720} \\
& 336 & 0.0620 & 0.0769 & \second{0.0617} & \second{0.0764} & \best{0.0606} & \best{0.0749} & 0.0623 & \best{0.0749} \\
& 720 & 0.0682 & 0.0839 & \second{0.0677} & \second{0.0835} & \best{0.0640} & \best{0.0793} & 0.0702 & 0.0844 \\
\midrule

\multirow{4}{*}{\rotatebox{90}{ILI-L}}
& 24  & \best{0.0531} & \best{0.0750} & 0.0768 & 0.1026 & 0.0942 & 0.1191 & \second{0.0680} & \second{0.0964} \\
& 36  & \second{0.0821} & \second{0.1024} & 0.0959 & 0.1233 & 0.0932 & 0.1236 & \best{0.0742} & \best{0.0973} \\
& 48  & \second{0.0991} & \best{0.1191} & 0.1146 & 0.1410 & 0.1032 & 0.1293 & \best{0.0958} & \second{0.1215} \\
& 60  & 0.1206 & 0.1423 & 0.1227 & 0.1479 & \second{0.1161} & \second{0.1412} & \best{0.1122} & \best{0.1408} \\

\bottomrule
\end{tabular}
\end{adjustbox}

\end{table}

\clearpage
\subsection{Generative Decoder}
\label{app:generative_decoder}

In this ablation we examine the generative head that produces the probabilistic
forecasts. KoFM models the predictive distribution via a closed-form Koopman
push-forward in a learned latent space, replacing the iterative sampling of
standard generative paradigms. To check its importance, we compare KoFM against
two alternative stochastic heads: a conditional VAE and a conditional normalizing
flow (RealNVP). To isolate the generative paradigm alone, each head is used on its own,
without AUG or the deterministic decoder, and all three consume the same KoPE latent
representation $\tilde{\mathbf{z}}_{N+i}$ as their conditioning input; the encoder, KoPE,
and training budget are kept identical. KoFM is trained with the KGO objective, while the
VAE and NF heads are trained only with the loss native to their paradigm (ELBO for the VAE,
exact log-likelihood for the NF), since enforcing a shared loss would not be well-defined
across these stochastic decoders.
We evaluate on a grid of $36$ dataset--horizon settings using seed $0$,
reporting CRPS and NMAE with the standard ProbTS evaluator.
\begin{table}[!ht]
\centering
\caption{Generative decoder ablation. Multivariate probabilistic forecasting results. \best{Red bold}: best; \second{blue underline}: second best.}
\label{tab:generative_decoder_ablation}
\begin{adjustbox}{width=\linewidth, max totalheight=0.65\textheight}
\begin{tabular}{ll cccccc}
\toprule
\multicolumn{2}{c}{} &
\multicolumn{2}{c}{KoFM} &
\multicolumn{2}{c}{VAE} &
\multicolumn{2}{c}{NF} \\
\midrule
Dataset & Horizon &
CRPS & NMAE &
CRPS & NMAE &
CRPS & NMAE \\
\midrule

\multirow{4}{*}{\rotatebox{90}{ETTm1-L}}
& 96  & \best{0.2150} & \best{0.2695} & \second{0.2205} & 0.2818 & 0.2206 & \second{0.2752} \\
& 192 & \best{0.2342} & \best{0.2955} & 0.2412 & 0.3081 & \second{0.2370} & \second{0.2997} \\
& 336 & \best{0.2491} & \best{0.3159} & \second{0.2632} & \second{0.3324} & 0.2689 & 0.3329 \\
& 720 & \best{0.2766} & \best{0.3490} & \second{0.3008} & \second{0.3798} & 0.3223 & 0.3979 \\
\midrule

\multirow{4}{*}{\rotatebox{90}{ETTm2-L}}
& 96  & \best{0.1047} & \best{0.1321} & 0.1098 & 0.1377 & \second{0.1077} & \second{0.1357} \\
& 192 & \second{0.1306} & \second{0.1637} & \best{0.1292} & \best{0.1634} & \second{0.1306} & 0.1642 \\
& 336 & \best{0.1449} & \best{0.1823} & 0.1500 & 0.1870 & \second{0.1491} & \second{0.1864} \\
& 720 & \best{0.1679} & \best{0.2092} & \second{0.1724} & \second{0.2135} & 0.1792 & 0.2245 \\
\midrule

\multirow{4}{*}{\rotatebox{90}{ETTh1-L}}
& 96  & \second{0.2434} & \best{0.3112} & 0.2445 & 0.3139 & \best{0.2395} & \second{0.3124} \\
& 192 & \second{0.2654} & \second{0.3410} & 0.2712 & 0.3463 & \best{0.2647} & \best{0.3392} \\
& 336 & \best{0.2821} & \best{0.3596} & 0.2890 & \second{0.3670} & \second{0.2880} & 0.3706 \\
& 720 & \best{0.2910} & \best{0.3781} & 0.3170 & 0.3950 & \second{0.2954} & \second{0.3833} \\
\midrule

\multirow{4}{*}{\rotatebox{90}{ETTh2-L}}
& 96  & \best{0.1345} & \best{0.1716} & \second{0.1390} & \second{0.1761} & 0.1431 & 0.1766 \\
& 192 & \best{0.1552} & \second{0.1970} & \second{0.1555} & \best{0.1967} & 0.1572 & 0.2007 \\
& 336 & \best{0.1765} & \best{0.2250} & \second{0.1789} & 0.2259 & 0.1835 & \second{0.2254} \\
& 720 & \best{0.1804} & \best{0.2326} & 0.1993 & \second{0.2370} & \second{0.1857} & 0.2378 \\
\midrule

\multirow{4}{*}{\rotatebox{90}{Elec.-L}}
& 96  & \second{0.0664} & \best{0.0786} & 0.0678 & 0.0870 & \best{0.0656} & \second{0.0835} \\
& 192 & \best{0.0739} & \best{0.0865} & 0.0751 & 0.0965 & \second{0.0747} & \second{0.0951} \\
& 336 & \best{0.0816} & \best{0.0969} & 0.0854 & 0.1104 & \second{0.0841} & \second{0.1076} \\
& 720 & \best{0.0998} & \best{0.1192} & 0.1074 & 0.1398 & \second{0.1037} & \second{0.1332} \\
\midrule

\multirow{4}{*}{\rotatebox{90}{Exch.-L}}
& 96  & \second{0.0194} & \second{0.0241} & \best{0.0191} & \best{0.0240} & \second{0.0194} & 0.0246 \\
& 192 & \second{0.0290} & \best{0.0352} & 0.0291 & 0.0359 & \best{0.0276} & \second{0.0354} \\
& 336 & \best{0.0406} & \best{0.0514} & \second{0.0422} & \second{0.0535} & 0.0518 & 0.0592 \\
& 720 & \best{0.0658} & \best{0.0797} & \second{0.0719} & 0.0891 & 0.0755 & \second{0.0856} \\
\midrule

\multirow{4}{*}{\rotatebox{90}{Traffic-L}}
& 96  & \best{0.1721} & \best{0.1979} & \second{0.1966} & \second{0.2479} & 0.2024 & 0.2528 \\
& 192 & \best{0.1782} & \best{0.2044} & 0.2078 & 0.2628 & \second{0.2065} & \second{0.2579} \\
& 336 & \best{0.1934} & \best{0.2225} & \second{0.2361} & \second{0.3017} & 0.2699 & 0.3413 \\
& 720 & \best{0.2302} & \best{0.2694} & 0.3364 & 0.4361 & \second{0.3325} & \second{0.4181} \\
\midrule

\multirow{4}{*}{\rotatebox{90}{Weather-L}}
& 96  & \best{0.0533} & \best{0.0661} & \second{0.0558} & \second{0.0680} & 0.0601 & 0.0723 \\
& 192 & \best{0.0582} & \best{0.0724} & \second{0.0628} & \second{0.0754} & 0.0991 & 0.1158 \\
& 336 & \best{0.0617} & \best{0.0764} & \second{0.0712} & \second{0.0845} & 0.0787 & 0.0970 \\
& 720 & \best{0.0677} & \best{0.0835} & \second{0.0787} & \second{0.0974} & 0.0993 & 0.1121 \\
\midrule

\multirow{4}{*}{\rotatebox{90}{ILI-L}}
& 24  & \second{0.0768} & \second{0.1026} & \best{0.0735} & \best{0.0985} & 0.0956 & 0.1333 \\
& 36  & \second{0.0959} & \second{0.1233} & 0.0964 & 0.1289 & \best{0.0699} & \best{0.0992} \\
& 48  & \second{0.1146} & \second{0.1410} & \best{0.1075} & \best{0.1327} & 0.1207 & 0.1568 \\
& 60  & \second{0.1227} & \best{0.1479} & \best{0.1221} & \second{0.1499} & 0.1256 & 0.1632 \\
\bottomrule
\end{tabular}
\end{adjustbox}
\end{table}

\clearpage
\subsection{Loss Function}
\label{app:loss_function}

The \(L_1\) accuracy term and the CRPS quantile loss
\(\rho_q(u) = u(q - \mathbb{I}[u<0])\) over levels \(q \in Q\) are by
now standard ingredients of forecasting objectives: the
\(L_1\) term penalizes deviations of forecast samples from the observed target, while the
quantile (pinball) term directly supervises the empirical quantiles
\(\hat{X}_q\) across probability levels, helping learn the location
and spread of the predictive distribution. In this ablation we isolate the
contribution of the \emph{Energy Score} repulsion term
\(\mathbb{E}\,\lVert \hat{X} - \hat{X}'\rVert\), where \(\hat{X}\) and
\(\hat{X}'\) are independent samples from the model (in practice
\(\hat{X}'\) is obtained by shuffling the samples in \(\hat{X}\)).
This term is meaningful at training time because KoFM produces
per-sample trajectories: each instance is replicated along a sample
dimension and each replica draws its own flow time and noise, so the
pairwise repulsion expectation \(\mathbb{E}\lVert \hat{X} - \hat{X}'\rVert\)
is a well-defined training signal that discourages sample collapse
to a single point prediction. We measure the impact of this term by
comparing the full loss against the loss with the
\(\mathbb{E}\lVert \hat{X} - \hat{X}'\rVert\) term removed, reporting
CRPS and NMAE per (dataset, horizon). Across eight datasets (excluding ILI) and four horizons per dataset, using seed $0$,
including the repulsion term reduces mean CRPS by $5.5\%$, with the
largest reduction of $10.4\%$ on ETTh1 at horizon $720$. The effect is consistent across horizons, with mean CRPS reductions
of $5.3\%$--$5.7\%$ at each horizon. Removing the term slightly improves NMAE, indicating a modest
trade-off between probabilistic and point-forecasting performance in this ablation. Overall, this cost is small
relative to the CRPS gain.
\begin{table}[!htbp]
\centering
\small
\caption{Loss-function ablation. KGO trained with the Repulsion term removed (Full is the KGO baseline). \best{Red bold}: best; \second{blue underline}: second best.}
\label{tab:loss_function_ablation}

\begin{adjustbox}{width=\linewidth, max totalheight=0.55\textheight}
\begin{tabular}{ll cccc}
\toprule

\multicolumn{2}{c}{} &
\multicolumn{2}{c}{Full} &
\multicolumn{2}{c}{w/o Repulsion} \\
\midrule

Dataset & Horizon &
CRPS & NMAE &
CRPS & NMAE \\
\midrule

\multirow{4}{*}{\rotatebox{90}{ETTm1-L}}
& 96  & \best{0.214} & \second{0.269} & \second{0.226} & \best{0.262} \\
& 192 & \best{0.236} & \second{0.296} & \second{0.254} & \best{0.293} \\
& 336 & \best{0.249} & \second{0.316} & \second{0.267} & \best{0.312} \\
& 720 & \best{0.275} & \second{0.347} & \second{0.293} & \best{0.343} \\
\midrule
\multirow{4}{*}{\rotatebox{90}{ETTm2-L}}
& 96  & \best{0.106} & \second{0.134} & \second{0.110} & \best{0.131} \\
& 192 & \best{0.131} & \second{0.164} & \second{0.134} & \best{0.158} \\
& 336 & \best{0.145} & \second{0.182} & \second{0.152} & \best{0.181} \\
& 720 & \best{0.171} & \second{0.214} & \second{0.174} & \best{0.207} \\
\midrule
\multirow{4}{*}{\rotatebox{90}{ETTh1-L}}
& 96  & \best{0.240} & \best{0.308} & \second{0.258} & \best{0.308} \\
& 192 & \best{0.265} & \best{0.340} & \second{0.286} & \second{0.346} \\
& 336 & \best{0.286} & \best{0.363} & \second{0.308} & \second{0.368} \\
& 720 & \best{0.284} & \best{0.369} & \second{0.317} & \second{0.380} \\
\midrule
\multirow{4}{*}{\rotatebox{90}{ETTh2-L}}
& 96  & \best{0.134} & \best{0.170} & \second{0.139} & \second{0.173} \\
& 192 & \best{0.155} & \second{0.197} & \second{0.162} & \best{0.196} \\
& 336 & \best{0.172} & \best{0.218} & \second{0.182} & \best{0.218} \\
& 720 & \best{0.176} & \best{0.226} & \second{0.184} & \second{0.228} \\
\midrule
\multirow{4}{*}{\rotatebox{90}{Elec.-L}}
& 96  & \best{0.067} & \best{0.078} & \second{0.071} & \best{0.078} \\
& 192 & \best{0.074} & \best{0.086} & \second{0.078} & \best{0.086} \\
& 336 & \best{0.081} & \second{0.096} & \second{0.087} & \best{0.095} \\
& 720 & \best{0.098} & \second{0.117} & \second{0.104} & \best{0.115} \\
\midrule
\multirow{4}{*}{\rotatebox{90}{Exch.-L}}
& 96  & \best{0.019} & \best{0.024} & \second{0.021} & \second{0.025} \\
& 192 & \best{0.029} & \second{0.037} & \second{0.030} & \best{0.035} \\
& 336 & \best{0.041} & \second{0.050} & \second{0.043} & \best{0.049} \\
& 720 & \best{0.069} & \second{0.082} & \second{0.070} & \best{0.079} \\
\midrule
\multirow{4}{*}{\rotatebox{90}{Traffic-L}}
& 96  & \best{0.173} & \second{0.199} & \second{0.180} & \best{0.196} \\
& 192 & \best{0.178} & \second{0.205} & \second{0.186} & \best{0.202} \\
& 336 & \best{0.192} & \second{0.222} & \second{0.200} & \best{0.219} \\
& 720 & \best{0.227} & \second{0.267} & \second{0.235} & \best{0.262} \\
\midrule
\multirow{4}{*}{\rotatebox{90}{Weather-L}}
& 96  & \best{0.053} & \second{0.066} & \second{0.057} & \best{0.065} \\
& 192 & \best{0.058} & \second{0.073} & \second{0.062} & \best{0.072} \\
& 336 & \best{0.062} & \second{0.077} & \second{0.065} & \best{0.075} \\
& 720 & \best{0.068} & \second{0.084} & \second{0.071} & \best{0.082} \\
\bottomrule
\end{tabular}
\end{adjustbox}

\end{table}

\clearpage
\subsection{Ablation Summary}
\label{app:ablation_summary}

\begin{figure}[!ht]
\centering
\includegraphics[width=\linewidth,height=0.92\textheight,keepaspectratio]{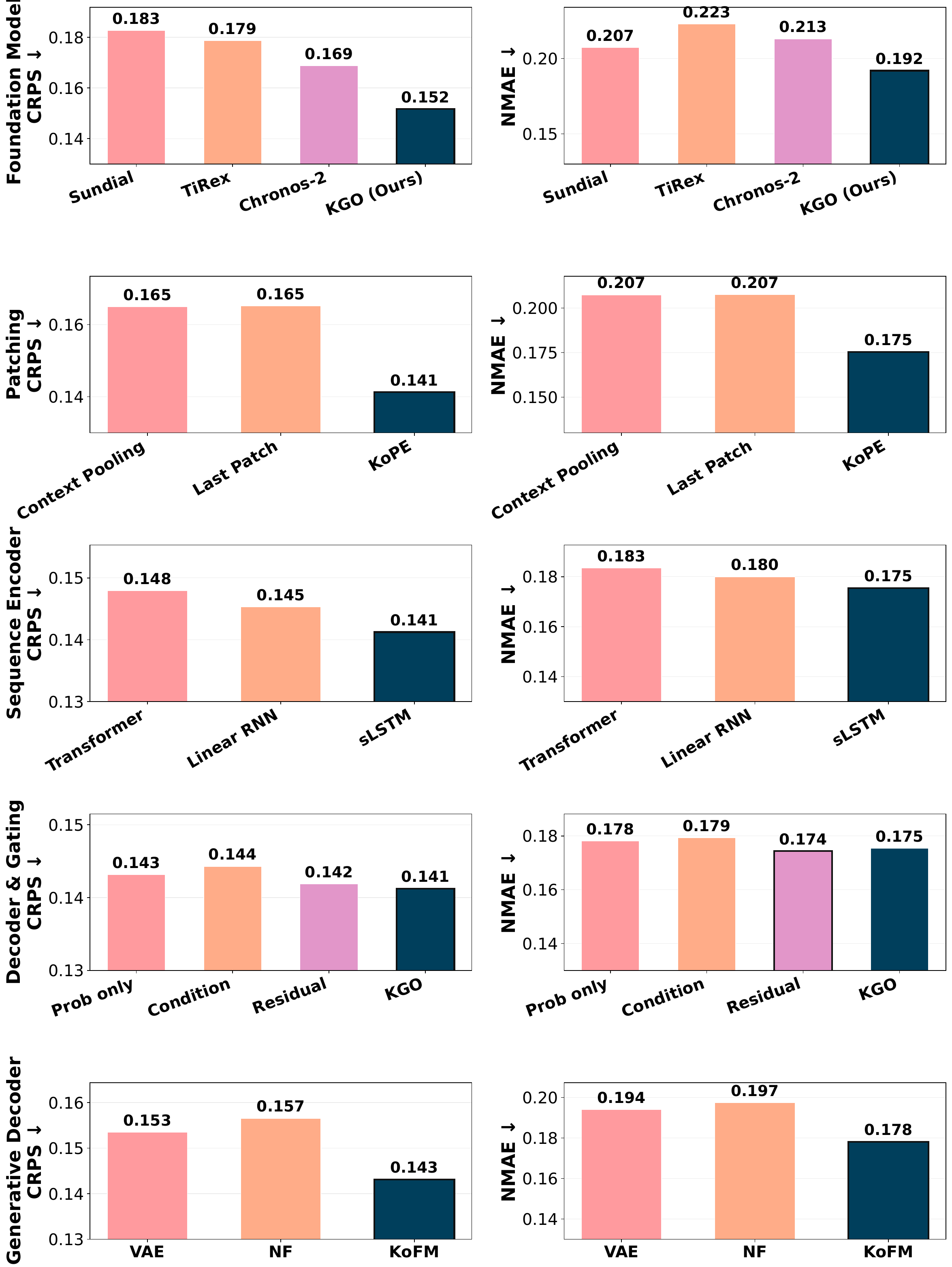}
\caption{Combined ablation summary across all five categories. Rows correspond to ablations (Foundation Models, Patching, Sequence Encoder, Decoder \& Gating, Generative Decoder); columns show CRPS (left) and NMAE (right). Each subplot averages over all 36 (dataset, horizon) cells with seed 0; the Foundation Models row averages over the 24 cells where all four models have results (ILI, Traffic, and Electricity excluded). The best value in each subplot is highlighted with a darker edge.}
\label{fig:ablation-summary-combined}
\end{figure}

\FloatBarrier

\clearpage

\section{Interpretability \& Analysis}
\label{app:analysis}
\subsection{Sampling Convergence}
\label{app:sampling}

In this subsection, we examine the number of Monte Carlo samples required for the
metrics to converge. To quantify this, we take the checkpoints of the trained models
and run inference on the ProbTS benchmark while varying the number of samples.
We sweep $S \in \{1, 5, 10, 20, 50, 100\}$ on the four ETT datasets at horizon
$H=96$ (seed $0$) and report CRPS and NMAE.

Both curves drop sharply from $S=1$ to $S\approx 10$ and then plateau: by $S=20$,
CRPS is within $0.005$ of its $S=100$ value, and NMAE is also within $0.005$.
On these datasets, using $20$ samples retains near-$100$-sample accuracy while
generating five times fewer trajectories, offering a practical option under memory constraints.

\begin{figure}[htbp]
\centering
\includegraphics[width=0.9\linewidth]{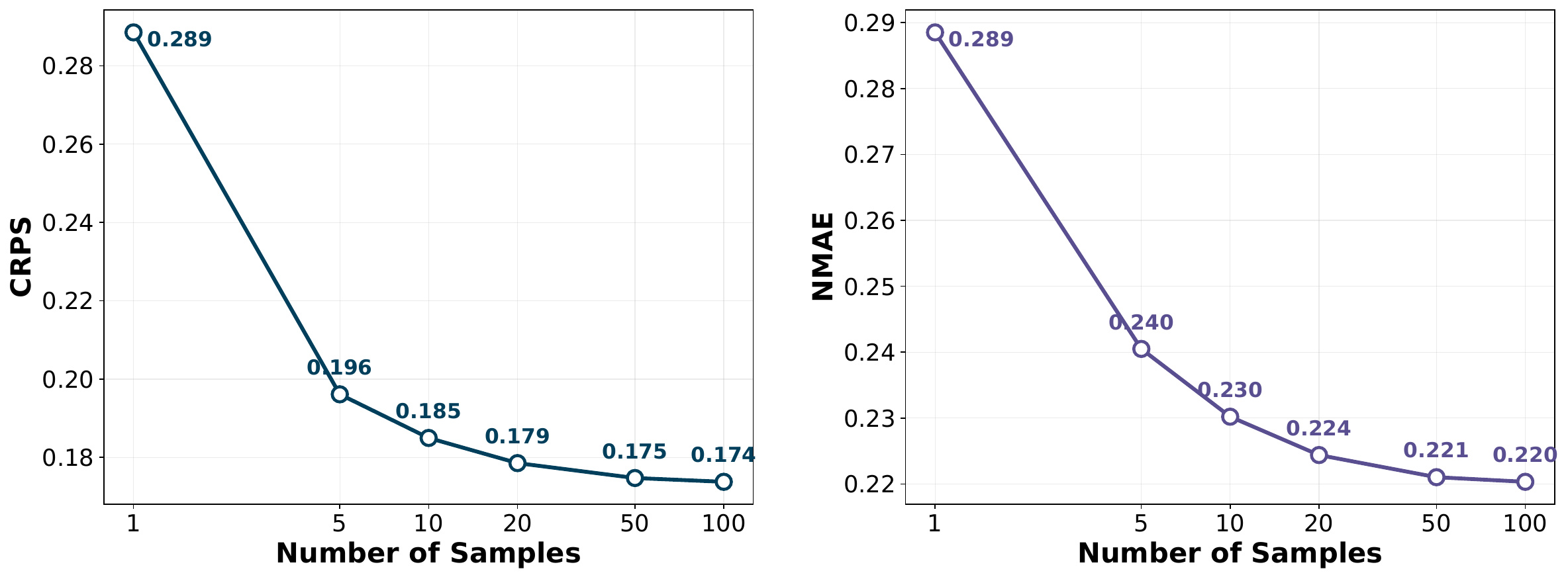}
\caption{Effect of sample count $S$ on CRPS (left) and NMAE (right), averaged over
the four ETT datasets at $H=96$.}
\label{fig:sampling_convergence}
\end{figure}

\FloatBarrier

\clearpage
\subsection{Spectral and Stability Behavior}
\label{app:eigenvalue_spectrum}

\subsubsection{Spectral Analysis}
\label{app:spectral_analysis}
\begin{figure}[!ht]
\centering
\includegraphics[width=\linewidth,height=0.82\textheight,keepaspectratio]{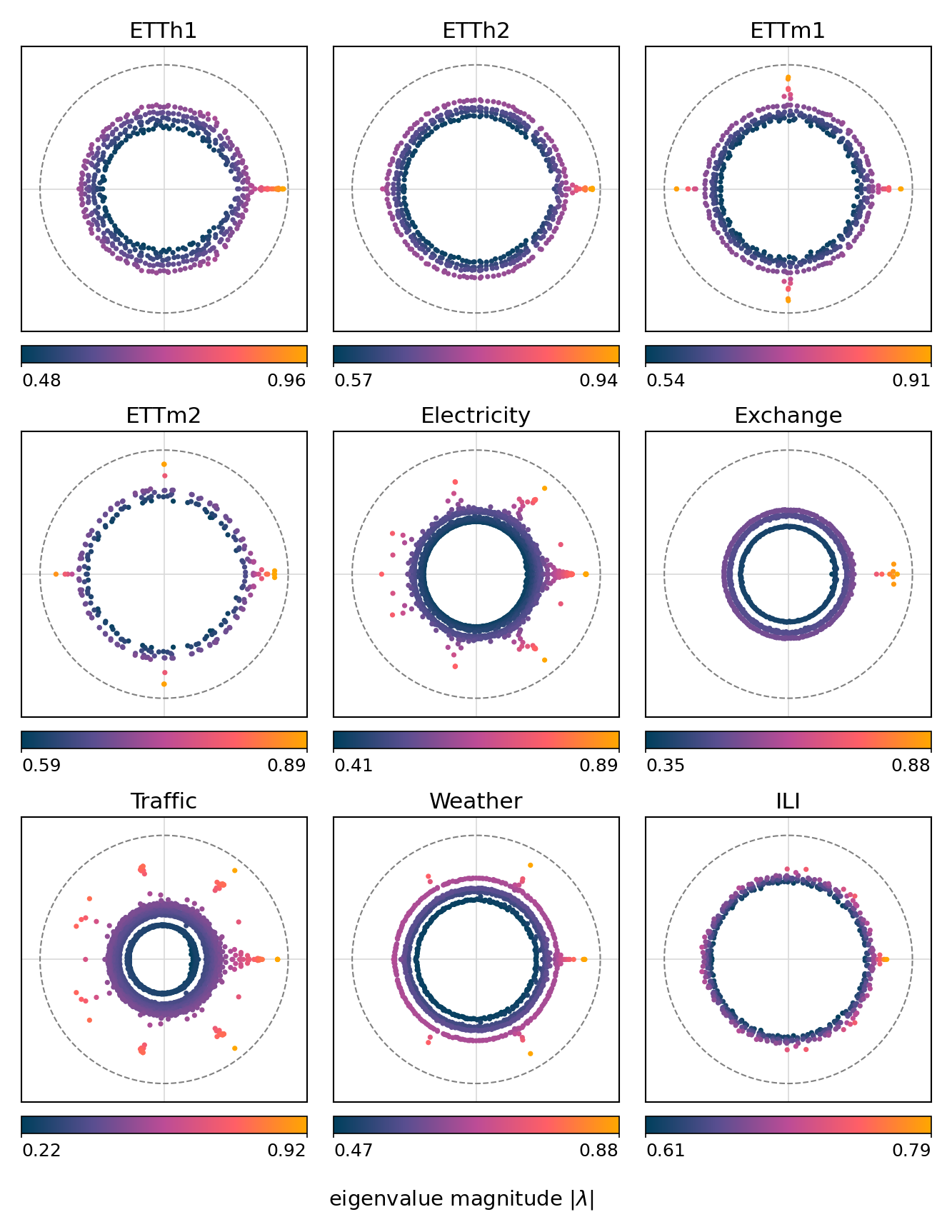}
\caption{Eigenvalues of the learned KoPE operator $\mathbf{K}$ on the nine long-term datasets, pooled over the four horizons (seed~0); the dashed circle is the unit circle. Color encodes $|\lambda|$ on each panel's own range, given by the bar below the panel. The isolated high-magnitude modes outside the bulk include the seasonal modes discussed in App.~\ref{app:spectral_analysis}.}
\label{fig:app:eigenvalue-spectrum}
\end{figure}

We examine the eigenvalue spectrum of the learned KoPE operator $K$ across nine datasets and four forecast horizons per dataset. Complementing the spectrum visualization in Fig.~\ref{fig:app:eigenvalue-spectrum}, Table~\ref{tab:app:kope-spectrum} reports the mean eigenvalue magnitude and the spectral radius $\rho(K)=\max_j|\lambda_j|$. Each statistic is averaged over the evaluated horizons.

\begin{table}[!htbp]
\centering
\begingroup
\setlength{\tabcolsep}{8pt}
\renewcommand{\arraystretch}{1.1}
\caption{Spectral statistics of the learned KoPE operator, averaged over horizons $\{96,192,336,720\}$, or $\{24,36,48,60\}$ for ILI, using seed~0.}
\label{tab:app:kope-spectrum}

\begin{adjustbox}{max width=\linewidth}
\begin{tabular}{l cc}
\toprule
Dataset & Mean $|\lambda|$ & $\rho(K)$ \\
\midrule
ETTh1       & 0.607 & 0.87 \\
ETTh2       & 0.659 & 0.87 \\
ETTm1       & 0.622 & 0.79 \\
ETTm2       & 0.667 & 0.84 \\
Electricity & 0.490 & 0.78 \\
Traffic     & 0.392 & 0.80 \\
Exchange    & 0.465 & 0.83 \\
Weather     & 0.570 & 0.74 \\
ILI         & 0.655 & 0.77 \\
\bottomrule
\end{tabular}
\end{adjustbox}

\endgroup
\end{table}

All eigenvalues lie inside the unit circle: the dataset-level averages of eigenvalue magnitude range from $0.392$ to $0.667$, and $\rho(K)<1$ in all 36 dataset--horizon configurations, ranging from $0.59$ to $0.96$. The horizon-averaged spectral radius ranges from $0.74$ to $0.87$. The bound $\rho(K)\leq\lVert K\rVert_2\leq 1$ follows from the spectral-norm normalization of $K$ (Sec.~\ref{subsec:kope}); in all evaluated configurations the learned operators are strictly contracting in spectrum and therefore asymptotically stable.

The largest-magnitude eigenvalues (the top $2\%$, pooled over horizons) are not scattered at random: their phase angles match the seasonal periods of the data. A mode with phase $\theta$ repeats every $360^\circ/\theta$ patches, so a season of $T$ time steps appears at $\theta=360^\circ P/T$. With $P=24$, the daily cycle of the 15-minute ETTm datasets ($T=96$) appears at $90.1^\circ$ (expected $90^\circ$), with its half-day harmonic at $180.0^\circ$ on ETTm1; the daily cycle of the 10-minute Weather dataset ($T=144$) at $59.9^\circ$ (expected $60^\circ$), with a second harmonic at $120.3^\circ$; and the weekly cycle of the hourly Electricity and Traffic datasets ($T=168$) at $51.5^\circ$ and $51.6^\circ$ (expected $51.4^\circ$), with second harmonics at $102.9^\circ$ and $102.8^\circ$. On the hourly ETTh datasets a patch spans exactly one day, so the daily cycle folds onto the positive real axis, where all of their largest eigenvalues lie; Exchange, which has no calendar seasonality, shows no consistent angle. These modes appear as the isolated high-magnitude points outside the bulk in Fig.~\ref{fig:app:eigenvalue-spectrum}. KoPE thus recovers the dominant daily and weekly cycles of each dataset without being given them.

\subsubsection{Stability}
\label{app:stability}

KoPE is initialized orthogonally, with LayerNorm applied before the operator rollout and to each resulting conditioning state. In addition, $K$ is normalized by its spectral norm whenever this norm exceeds $1$ (Sec.~\ref{subsec:kope}), so $\lVert K^{i}\mathbf{z}\rVert_2\leq\lVert\mathbf{z}\rVert_2$ for every power and the rollout cannot expand. KoPE also uses a finite rollout determined by the prediction horizon and patch size: $M=\lceil H/P\rceil$, giving $30$ conditioning states for $H=720$ and $P=24$.

\FloatBarrier

\clearpage
\subsection{Uncertainty Gate}
\label{app:uncertainty_gate}
Predictability in time series varies across both forecast horizons and variates: longer-range forecasts generally involve greater uncertainty, while different underlying processes and measurement conditions introduce different levels of noise. AUG accommodates this heterogeneity by learning the relative contributions of deterministic and stochastic predictions through the forecasting objective. Its weights describe the contributions of the two branches and should not be interpreted as direct estimates of aleatoric uncertainty.

\begin{figure}[!ht]
\centering
\includegraphics[width=\linewidth,height=0.58\textheight,keepaspectratio]{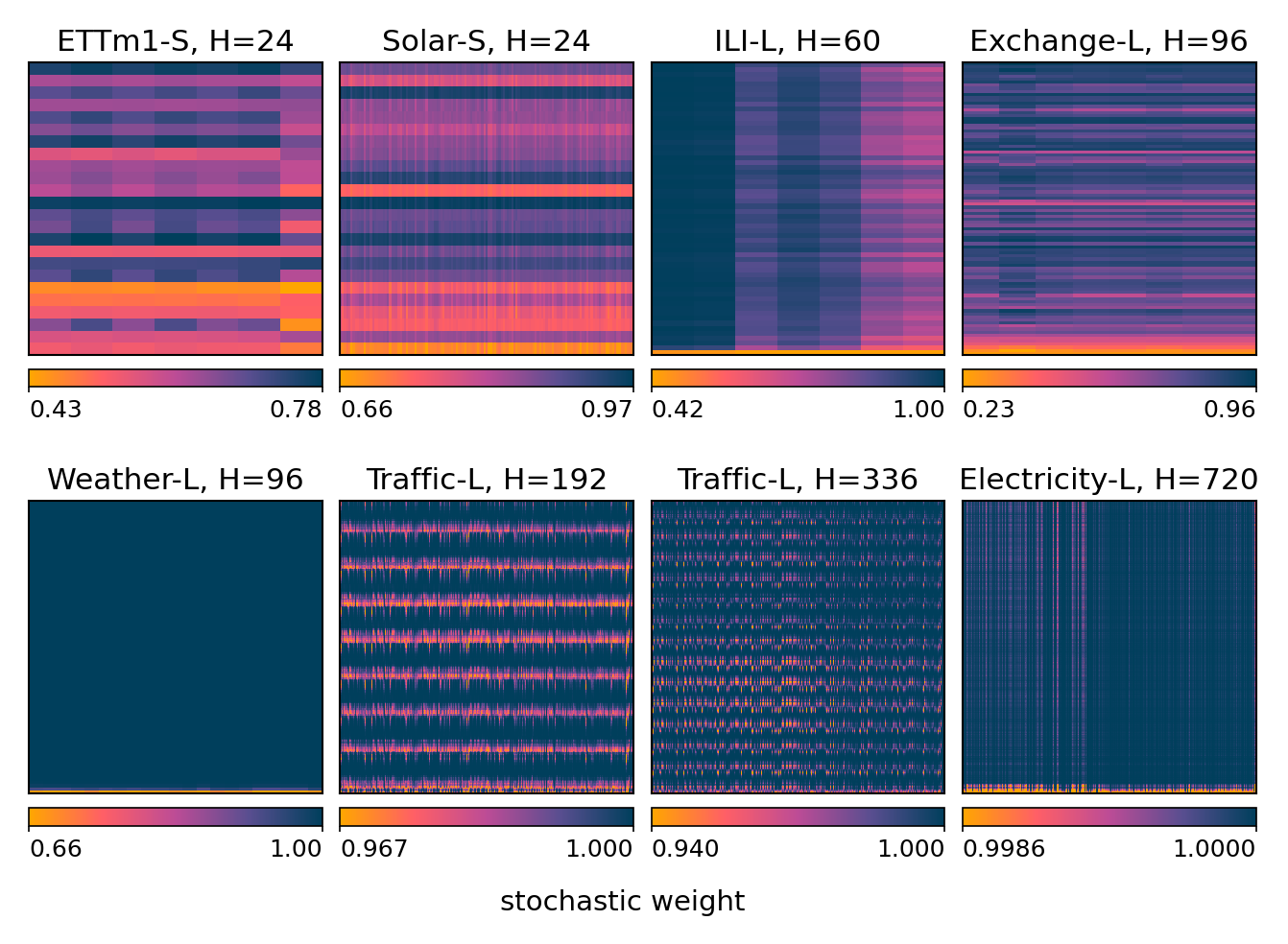}
\caption{Gate weights averaged over the test windows (seed~0) for eight dataset--horizon pairs, named in the panel titles. Within each panel, the horizontal axis indexes variates and the vertical axis indexes forecast steps, increasing from bottom to top. Color encodes the stochastic weight on the panel's own range, from its lowest value (amber; the 1st percentile for Traffic-L and Electricity-L, whose few lowest cells would otherwise wash out the rest) to its highest (navy), given by the bar below the panel.}
\label{fig:app:gate-heatmaps}
\end{figure}

Figure~\ref{fig:app:gate-heatmaps} shows the stochastic weight across forecast steps (vertical axis) and variates (horizontal axis) for eight representative dataset--horizon pairs. Three structures stand out. First, except on Traffic, the first forecast steps are the most deterministic and the stochastic weight rises with lead time (on Exchange-L from $0.27$ to around $0.8$): the immediate future is well predicted from the recent context, while aleatoric uncertainty accumulates further ahead. Second, the short-term panels keep a substantial deterministic share (mean weight $0.64$ on ETTm1-S), whereas on most long-term datasets the gate is nearly fully stochastic beyond the first steps. Third, the gate distinguishes variates: on ILI-L the noisy influenza-like-illness rates are almost fully stochastic ($0.98$), while the slowly changing number of reporting providers relies more on the deterministic branch ($0.78$). Together, these views show that AUG places more weight on the stochastic branch where aleatoric uncertainty is expected to be larger.

\FloatBarrier

\clearpage
\subsection{Qualitative Forecasts}
\label{app:qualitative_forecasts}
\begin{figure}[!ht]
\centering
\includegraphics[width=\linewidth,height=0.82\textheight,keepaspectratio]{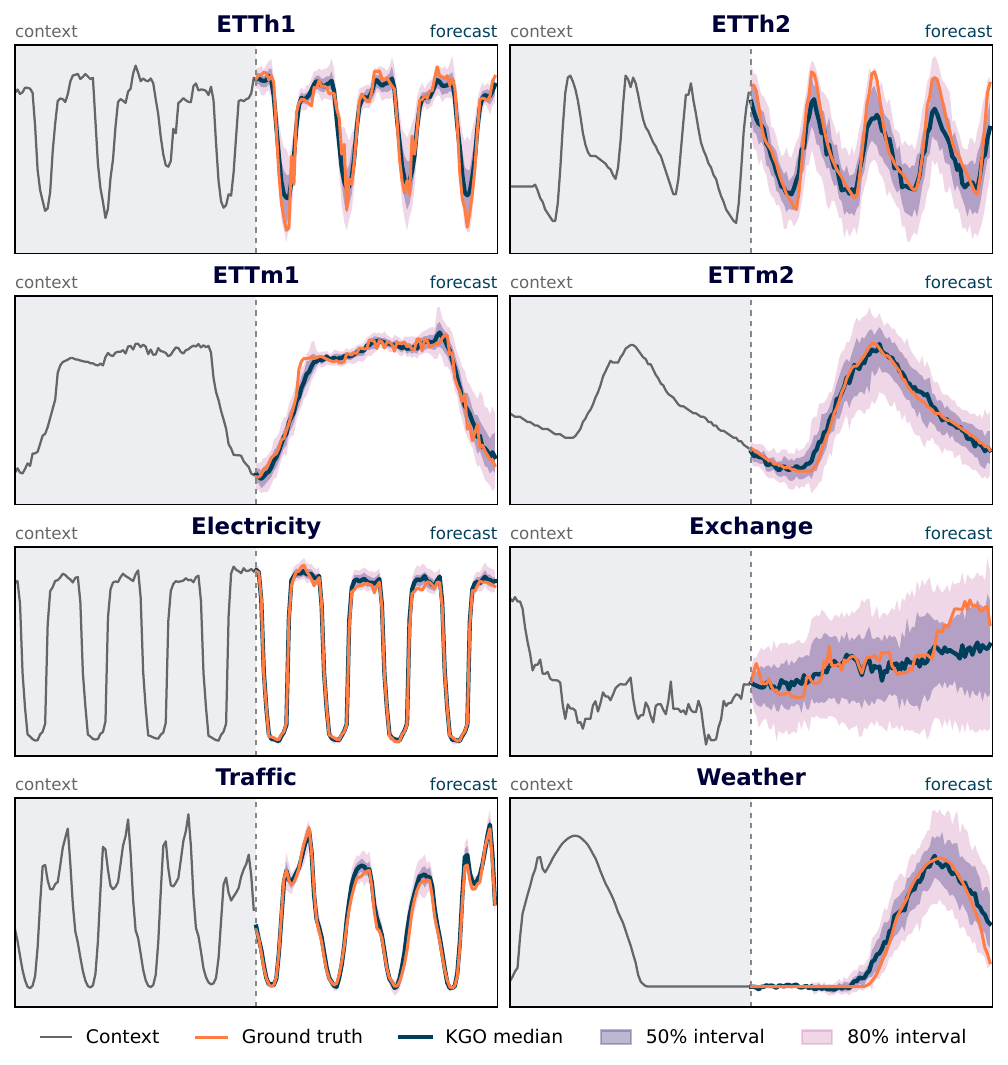}
\caption{Qualitative forecasts at horizon $H{=}96$ (predictive interval view): context (gray), ground truth (orange), and KGO median (navy) with $50\%$ (purple) and $80\%$ (light magenta) predictive intervals.}
\label{fig:forecast_grid_h96}
\end{figure}

\begin{figure}[!ht]
\centering
\includegraphics[width=\linewidth,height=0.82\textheight,keepaspectratio]{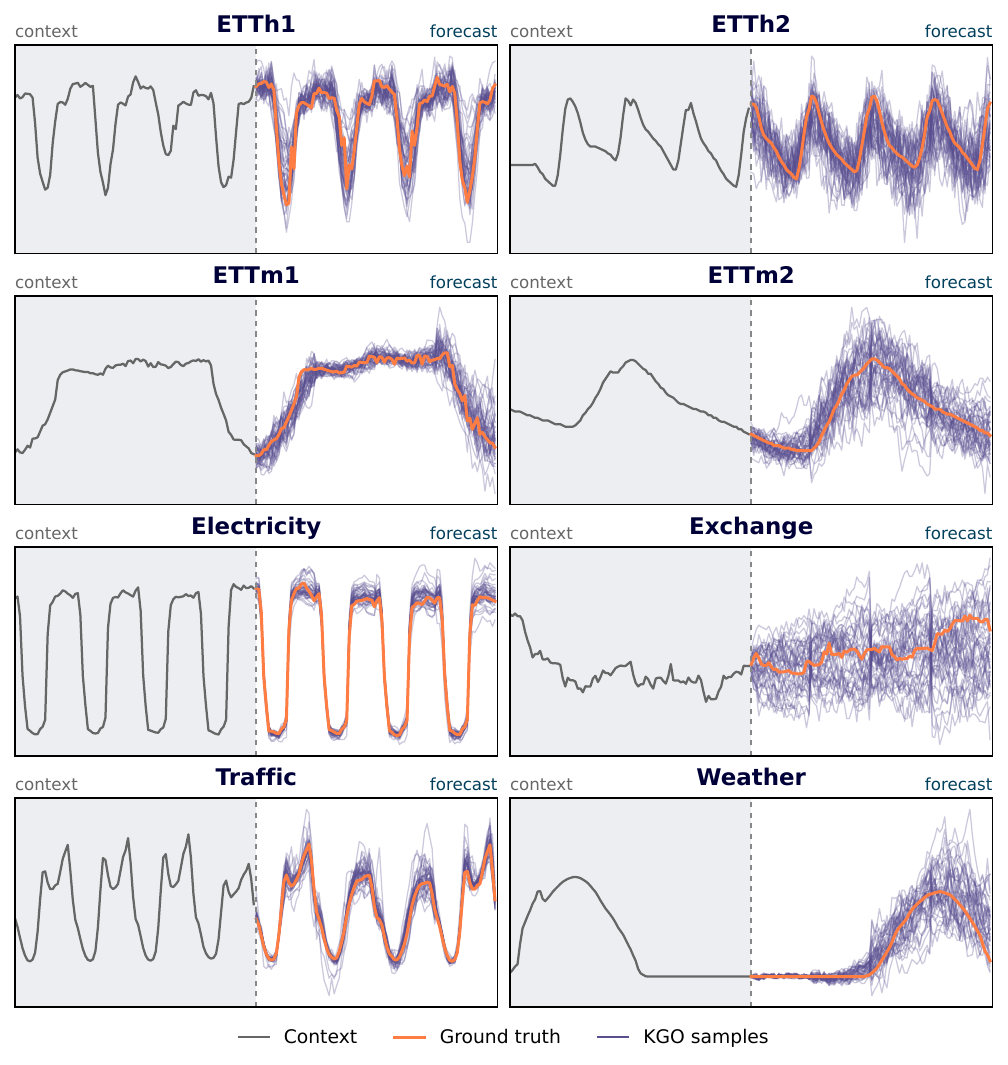}
\caption{Qualitative forecasts at horizon $H{=}96$ (generative view): same windows as Figure~\ref{fig:forecast_grid_h96}, but rendered as $40$ individual forecast trajectories sampled from KGO (purple) overlaid with the ground truth (orange). The density of trajectories visualizes the predictive distribution directly.}
\label{fig:forecast_generative_h96}
\end{figure}

\end{document}